\documentclass{article}

\usepackage{microtype}
\usepackage{graphicx}
\usepackage{booktabs} %

\usepackage{hyperref}

\usepackage[accepted]{icml2024}

\usepackage{amsmath}
\usepackage{amssymb}
\usepackage{mathtools}
\usepackage{amsthm}

\usepackage[capitalize,noabbrev]{cleveref}

\theoremstyle{plain}
\newtheorem{theorem}{Theorem}[section]

\newtheorem{lemma}[theorem]{Lemma}

\theoremstyle{definition}

\theoremstyle{remark}

\usepackage{caption}
\usepackage{subcaption}
\usepackage{amsmath}
\usepackage{paralist}

\usepackage[textsize=tiny]{todonotes}
\newcommand{\norm}[1]{\left\lVert#1\right\rVert}

\icmltitlerunning{Trust the Model Where It Trusts Itself - Model-Based Actor-Critic with Uncertainty-Aware Rollout Adaption}

\begin{document}

\twocolumn[
\icmltitle{Trust the Model Where It Trusts Itself\\ Model-Based Actor-Critic with Uncertainty-Aware Rollout Adaption}

\icmlsetsymbol{equal}{*}

\begin{icmlauthorlist}
\icmlauthor{Bernd Frauenknecht}{equal,dsme}
\icmlauthor{Artur Eisele}{equal,dsme}
\icmlauthor{Devdutt Subhasish}{dsme}
\icmlauthor{Friedrich Solowjow}{dsme}
\icmlauthor{Sebastian Trimpe}{dsme}
\end{icmlauthorlist}

\icmlaffiliation{dsme}{Institute for Data Science in Mechanical Engineering (DSME), RWTH Aachen University, 52068 Aachen, Germany}

\icmlcorrespondingauthor{Bernd Frauenknecht}{bernd.frauenknecht@dsme.rwth-aachen.de}

\icmlkeywords{Model-based Reinforcement Learning, Uncertainty Quantification}

\vskip 0.3in
]

\printAffiliationsAndNotice{\icmlEqualContribution} %

\begin{abstract}
Dyna-style model-based reinforcement learning (MBRL) combines model-free agents with predictive transition models through model-based rollouts.
This combination raises a critical question: ``\textit{When to trust your model?}''; i.e., which rollout length results in the model providing useful data?
\citet{Janner2019Dec} address this question by gradually increasing rollout lengths throughout the training.
While theoretically tempting, uniform model accuracy is a fallacy that collapses at the latest when extrapolating. 
Instead, we propose asking the question ``\textit{Where to trust your model?}''.
Using inherent model uncertainty to consider local accuracy, we obtain the Model-Based Actor-Critic with Uncertainty-Aware Rollout Adaption (MACURA) algorithm. 
We propose an easy-to-tune rollout mechanism and demonstrate substantial improvements in data efficiency and performance compared to state-of-the-art deep MBRL methods on the MuJoCo benchmark.

\end{abstract}

\section{Introduction}

Deep reinforcement learning (RL) has shown unprecedented results in challenging domains such as gameplay \cite{Mnih2015Feb, openai2019dota} and nonlinear control \cite{openai2019rubiks, P.R.2022Feb}. For engineering problems, however, the data inefficiency of model-free state-of-the-art approaches \cite{Schulman2017, Haarnoja2018Jul} remains a substantial challenge \cite{Kostrikov2023Jul, Frauenknecht2023}, prompting the need for more efficient methods \cite{Janner2019Dec, Chen2021}. 
One such method, model-based reinforcement learning (MBRL), 
reduces the necessary degree of environment interaction by inferring information from a learned environment model. 
Unfortunately, model errors can lead to faulty conclusions that severely impact the agent's performance.
It is therefore critical to ensure the accuracy of these models. 

Model-based policy optimization (MBPO) \cite{Janner2019Dec} represents the current state-of-the-art in Dyna-style MBRL \cite{Sutton1991Jul}, combining a Soft Actor-Critic (SAC) agent \cite{Haarnoja2018Jul, Haarnoja2018Dec} with a Probabilistic Ensemble (PE) model \cite{Lakshminarayanan2017Dec}.
\citet{Janner2019Dec} address two distinct learning problems:
using interaction between the agent and the environment to train the model and, simultaneously, employing model-based rollouts to train the agent. 
The agent queries the model in short rollouts branched off from states that were observed during environment interaction.
The length of these rollouts is gradually increased throughout training, balancing model usage against the risk of model exploitation.

\citet{Janner2019Dec} answer the question \textit{``When to trust your model?''} with time-based arguments.
Essentially, uniform model accuracy is postulated after sufficiently long training which motivates to use the model for rollouts of predetermined length.
However, both the training time and the rollout lengths are notoriously difficult to preschedule. Furthermore, the assumption of uniform model improvement is problematic for complex systems.
 
 Instead, we consider model accuracy as a local property, putting the question \textit{``Where to trust your model?''} at the heart of our approach. 
 The inherent uncertainty of PE models allows for adaptive rollout lengths:
 wherever the model is uncertain, rollouts are terminated quickly, while longer rollouts can be generated where it is certain.

In this paper, we analyze the learning process of Dyna-style MBRL and present Model-Based Actor-Critic with Uncertainty-Aware Rollout Adaption (MACURA), an algorithm with an easy-to-tune mechanism for model-based rollout length scheduling. In particular, we present the following technical contributions:
\begin{compactitem}
    \item We show monotonic improvement by restricting model-based rollouts to a subset $\mathcal{E} \subseteq \mathcal{S}$ of the state space $\mathcal{S}$;
    \item We construct the subset $\mathcal{E}$ based on a novel and easy-to-compute model uncertainty measure; and
    \newpage
    \item We outperform state-of-the-art Dyna-style MBRL methods with regard to data efficiency and asymptotic performance on the MuJoCo benchmark.
\end{compactitem}

\section{Background}
In the following, we introduce the fundamental concepts of MBRL, and  Dyna-style architectures in particular.
\subsection{Reinforcement Learning}
We assume the environment to be represented by a discounted Markov decision process (MDP) defined by $\mathcal{M} = (\mathcal{S}, \mathcal{A}, r, p, \gamma, \rho_0 )$, with $\mathcal{S}$ the state and $\mathcal{A}$ the action space, while a dynamics function $p(s^{\prime} \mid s, a)$ describes transitions between states $s \in \mathcal{S}$ and actions $a \in \mathcal{A}$. Reward $r \in \mathbb{R}$ is generated from a reward function $r(s, a)$ and is discounted by $\gamma \in (0, 1)$. The MDP is initialized from an initial state distribution $\rho_0$. The RL agent aims to find an optimal policy $\pi^*$ that maximizes the expected discounted sum of rewards, henceforth referred to as expected return $\eta$. Thus,
\begin{equation}
  \pi^*=\underset{\pi}{\operatorname{argmax~}} \eta[\pi]=\underset{\pi}{\operatorname{argmax~}} E_\pi\left[\sum_{t=0}^{\infty} \gamma^t r\left(s_t, a_t\right)\right],  
  \label{eq:RL_objective}
\end{equation}
with $s_0 \sim \rho_0$, $a_t \sim \pi ( \cdot \mid s_t)$, and $s_{t+1} \sim p ( \cdot \mid s_t, a_t)$.
The Q-function represents $\eta [\pi]$ conditioned on specific state-action pairs and is given by
\begin{equation}
Q^{\pi}(s_t, a_t) = E_\pi\left[\sum_{k=t}^{\infty} \gamma^{k-t} r\left(s_k, a_k\right) \bigg| s_t, a_t \right].
\end{equation}
\subsection{Probabilistic Ensemble Models}

In MBRL, we learn a dynamics model $\tilde{p}(s^{\prime} \mid s, a)$ to approximate the unknown environment dynamics $p(s^{\prime} \mid s, a)$.
In the following, we consider the particularly effective PE approach \cite{Lakshminarayanan2017Dec, Chua2018} as the model class for dynamics learning. A PE consists of $E$ probabilistic neural networks (PNN) with parameters $\theta_e$, $e \in \{1, \dots, E \}$, which are trained on bootstrapped datasets via a negative $log$-likelihood loss to approximate the distribution over the next state with a Gaussian distribution
\begin{equation}
\tilde{p}_{\theta_e}(s^{\prime} \mid s, a) = \mathcal{N}(\mu_{\theta_e}(s,a) , \Sigma_{\theta_e}(s,a)).
\end{equation}
The PE architecture allows a distinction between aleatoric uncertainty due to the process noise of the environment and epistemic uncertainty due to the parametric uncertainty of the model. While aleatoric uncertainty corresponds to high individual variance estimates $\Sigma_{\theta_e}(s, a)$, epistemic uncertainty is measured via model disagreement. \citet{Lakshminarayanan2017Dec} define epistemic uncertainty as the pairwise Kullback-Leibler (KL) divergence $D_{\mathrm{KL}}$ between the individual PNN predictions $p_{\theta_e}$ and the Gaussian mixture distribution of the ensemble prediction $\tilde{p}_{\mathrm{PE}}$, given by
\begin{equation}
\label{eq:KL_epistemic}
       u_{\mathrm{KL}} = \sum_{e=1}^{E} D_{\mathrm{KL}} (\tilde{p}_{\theta_e}(s^{\prime} \mid s, a) \| \tilde{p}_{\mathrm{PE}}(s^{\prime} \mid s, a)),
\end{equation}
with
\begin{equation}
       \tilde{p}_{\mathrm{PE}}(s^{\prime} \mid s, a) := \frac{1}{E} \sum_{e=1}^{E} \tilde{p}_{\theta_e}(s^{\prime} \mid s, a).
\end{equation}
For the following analysis, we assume $r(s, a)$ is known.

\subsection{Dyna-Style Model-Based Reinforcement Learning}
\label{subsec: dyna-style_mbrl}
\begin{figure}
    \centering
    \includegraphics[width=\columnwidth]{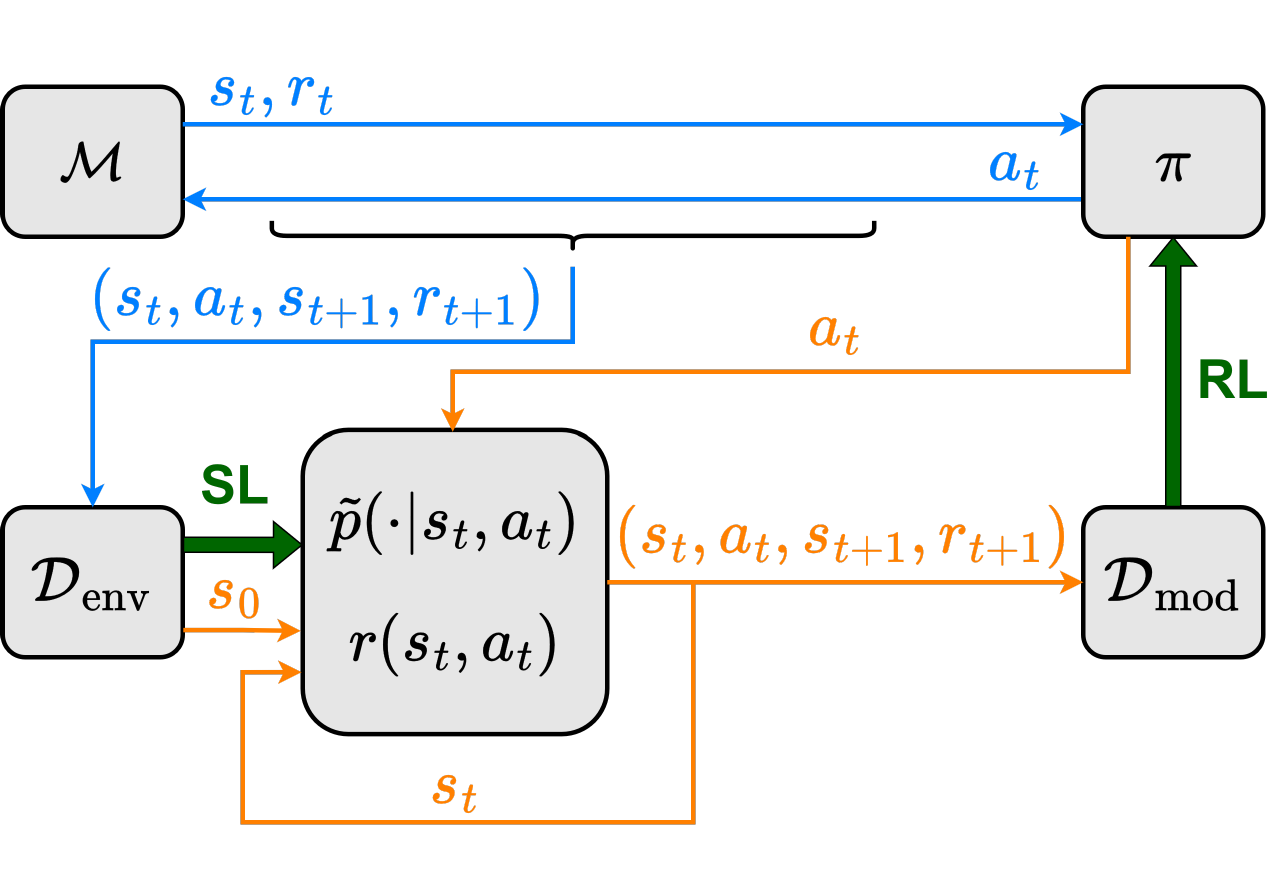}
    \caption{Dyna-style MBRL. \textit{An agent with policy $\pi$ interacts with the environment $\mathcal{M}$. This data is stored in $\mathcal{D}_{\mathrm{env}}$ and used to train a dynamics model $\tilde{p}$ via supervised learning (SL). Model-based rollouts under $\pi$ are performed from start states $s_0$ in $\mathcal{D}_{\mathrm{env}}$ and stored in $\mathcal{D}_{\mathrm{mod}}$. The policy is trained on $\mathcal{D}_{\mathrm{mod}}$ via reinforcement learning (RL).}}
    \label{fig:dyna_mbrl}
    \vspace{-5mm}
\end{figure}
We focus on Dyna-style MBRL \cite{Sutton1991Jul} and MBPO \cite{Janner2019Dec} in particular as it represents the current state-of-the-art approach. A schematic of the architecture is depicted in Figure \ref{fig:dyna_mbrl}. In MBPO, a model-free SAC agent \cite{Haarnoja2018Jul, Haarnoja2018Dec} with policy $\pi$ interacts with the environment. The corresponding agent-environment interaction data are stored in a replay buffer $\mathcal{D}_{\mathrm{env}}$ and are used to train a dynamics model $\tilde{p}$. This model then generates experience data in branched model-based rollouts that are stored in a replay buffer $\mathcal{D}_{\mathrm{mod}}$ and used to train the model-free agent.  Branched model-based rollouts are thus essential to Dyna-style MBRL as the accuracy of the generated experience data determines the performance of the agent. The mechanism is illustrated in Algorithm \ref{alg:rollout_basic}.

Branched model-based rollouts start at random $s_0 \sim \mathcal{U}(\mathcal{D}_{\mathrm{env}})$, with $\mathcal{U}(\cdot)$ the uniform distribution, and stop at a maximum rollout length $T_{\mathrm{max}} \in \mathbb{N}$. During the rollout, the propagating PNN model within the PE is randomly sampled for each time step. Actions and states are sampled from the respective Gaussians of the policy and the PNN model.

The environment buffer $\mathcal{D}_{\mathrm{env}}$ therefore serves two purposes for model-based branched rollouts. 
It acts as the training data set for $\tilde{p}$, determining where the model is accurate, and it 
 induces the set of start states for model-based rollouts, influencing the data distribution in $\mathcal{D}_{\mathrm{mod}}$. 

A key advantage of branched model-based rollouts is that they reduce the quantity of environment data necessary for learning. This advantage stems from two learning mechanisms. First, the number of update steps per observed transition is limited due to instabilities in value function learning \cite{Chen2021}. The model allows a multitude of transitions to be generated, mitigating this problem. Second, the data distribution in $\mathcal{D}_{\mathrm{mod}}$ differs from that of $\mathcal{D}_{\mathrm{env}}$ and can be more informative to the agent. Generalization capabilities of $\tilde{p}$ allow for a richer set of transitions to be collected from the model. Further, model-based rollouts are conducted under the current policy $\pi$ of the agent. Therefore, model-based rollouts shift the off-policy distribution in $\mathcal{D}_{\mathrm{env}}$ more towards an on-policy distribution. This generally puts a stronger emphasis on the effects of the current policy and interesting areas of $\mathcal{S}$, accelerating policy improvement.
The longer branched model-based rollouts are, the more the data distributions of $\mathcal{D}_{\mathrm{env}}$ and $\mathcal{D}_{\mathrm{mod}}$ may differ. 
Typically, $T_{\mathrm{max}}$ is gradually increased throughout training.

\begin{algorithm}[tb]
\caption{``Vanilla'' Branched Model-based Rollouts }\label{alg:rollout_basic}
\begin{algorithmic}
    \STATE Given $\mathcal{D}_{\mathrm{env}}$, $\mathcal{D}_{\mathrm{mod}}$, $\tilde{p}_{\theta_{1, \dots, E}}$, $\pi$, and $T_{\mathrm{max}}$
    \STATE $s_0 \sim \mathcal{U} ( \mathcal{D}_{\mathrm{env}} )$
    \FOR {$t =0, \dots, T_{\mathrm{max}}-1$}
    \STATE $e_t \sim \mathcal{U} (1, \dots, E)$
    \STATE $a_t \sim \pi(\cdot \mid s_t)$
    \STATE $s_{t+1} \sim \tilde{p}_{\theta_{e_t}}(\cdot \mid s_t, a_t)$
    \STATE $r_{t+1} = r(s_t, a_t)$
    \STATE $\mathcal{D}_{\mathrm{mod}} \gets \mathcal{D}_{\mathrm{mod}} \cup \{ (s_t, a_t, r_{t+1}, s_{t+1} )\}$
    \ENDFOR
\end{algorithmic}
\end{algorithm}
Algorithm \ref{alg:standard_deep_dyna} in Appendix \ref{appendix:algos} provides a detailed description of MBPO.
We build on the eminent MBPO method, modifying its core mechanism of branched model-based rollouts. We develop a method to estimate \emph{where} to trust the model and build a new adaptive rollout scheme around it. This scheme may likewise benefit a multitude of derivatives of MBPO such as \cite{Zhang2020Jun, Shen2020Dec, Lai2020Nov, Lai2021Nov, Morgan2021May, Frohlich2021, Luis2023Feb, Luis2023Dec, Wang2021Dec}.

\section{Where to Trust your Model?}
\label{sec:branched_rollouts}

Our key idea is to define a subset $\mathcal{E}_{\Tilde{p}, k} \subseteq \mathcal{S}$ on which the model $\Tilde{p}$ is sufficiently accurate at time $k$. We refer to this subset as $\mathcal{E}$ for brevity and put it at the heart of our approach.

The main technical  \textbf{problem formulation} then becomes:
 How can we construct the set $\mathcal{E}$ that ensures a desired accuracy? In particular, we need to:
\begin{compactenum}[i)]
    \item quantify the notion of \emph{sufficiently accurate};
    \item estimate $\mathcal{E}$ from a given model $\Tilde{p}$; and
    \item expand $\mathcal{E}$ quickly via informative data in $\mathcal{D}_{\mathrm{env}}$.
\end{compactenum}

Once we have a suitable $\mathcal{E}$, we obtain a straightforward yet precise mechanism for branched model-based rollouts. As long as the rollout stays within $\mathcal{E}$, the benefits of long rollouts discussed in Section \ref{subsec: dyna-style_mbrl} outweigh the risk of model exploitation, while the opposite is the case once it leaves $\mathcal{E}$.

\begin{figure}
    \centering
    \includegraphics[width=\columnwidth]{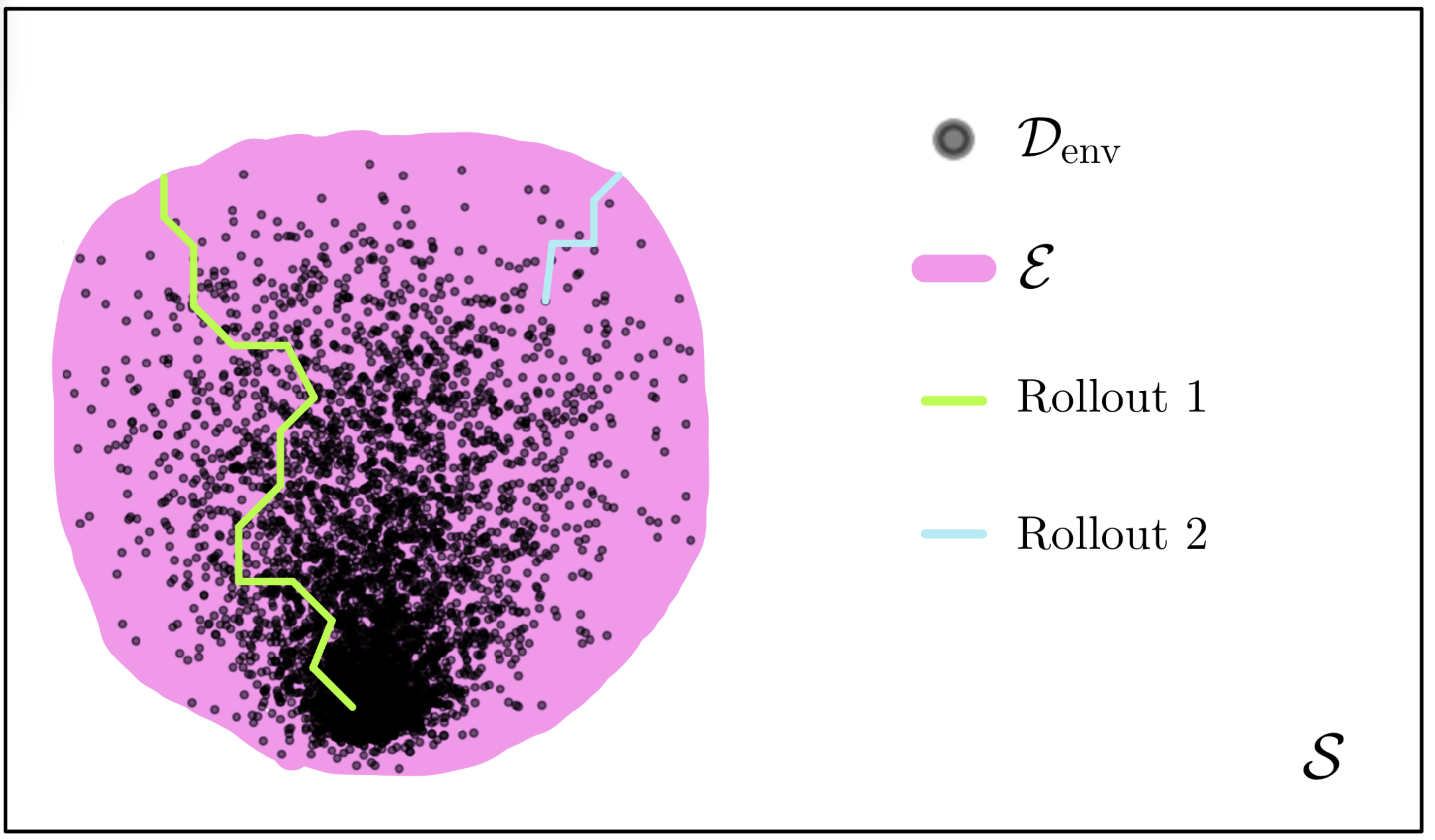}
    \caption{Where to trust your Model? \textit{ $\mathcal{D}_{\mathrm{env}}$ induces a set of sufficient model accuracy $\mathcal{E} \subseteq \mathcal{S}$. A notion of $\mathcal{E}$ allows to reason whether rollouts are in a region of sufficient model accuracy. We use this resoning to schedule rollout length. }}
    \label{fig:subset}
    \vspace{-5mm}
\end{figure}

Such a distinction is vital as model-based rollouts are performed from random start states of $\mathcal{D}_{\mathrm{env}}$, as depicted in Figure \ref{fig:subset}. Depending on the rollout policy, these start states can either lead to rollouts staying in $\mathcal{E}$ for a long time without the necessity to terminate quickly, or leaving it early,  where a careful rollout length adaption is crucial. A fixed rollout length based on training time \cite{Janner2019Dec} cannot account for local differences in model accuracy.

\section{Monotonic Improvement under\\ Dynamics Misalignment on $\mathcal{E}$}
\label{sec: improvement bound}

Following the insights set out in Section \ref{sec:branched_rollouts}, we provide a formal justification of our approach.
We analyze the effect of dynamics mismatch on the expected return, which will be instrumental for formally defining $\mathcal{E} \subseteq \mathcal{S}$ in Section \ref{sec:uncertainty estimate} that will be used to schedule rollout length.

\subsection{Formulation of Monotonic Improvement}
In the context of branched model-based rollouts, we define two MDPs, $\hat{\mathcal{M}}$ and $\tilde{\mathcal{M}}$, on the same $\mathcal{S}$, $\mathcal{A}$, $r(s,a)$, $\gamma$ and start state distribution $\rho_{\mathrm{BR}}$. Here, $\hat{\mathcal{M}}$ represents the MDP for branched rollouts under the environment dynamics $p(s^{\prime} \mid s, a)$, whereas $\tilde{\mathcal{M}}$ is the MDP for branched model-based rollouts following model dynamics $\tilde{p}(s^{\prime} \mid s, a)$.
Both, $\hat{\mathcal{M}}$ and $\tilde{\mathcal{M}}$, have identical start states $s_0^{\hat{\mathcal{M}}} = s_0^{\tilde{\mathcal{M}}} \in \mathcal{E}$.

 The MDPs are coupled through the following stopping times (which are random variables and thus measurable functions $T : \Omega \rightarrow \mathbb{N}$ as indicated by the argument $T(\omega)$):
\begin{equation}
T(\omega) := \min\{T^{\hat{\mathcal{M}}}, T^\mathcal{\tilde M}\} - 1 ,
\label{eq:stopping_time_both}
\end{equation}
\begin{equation}
 T^{\hat{\mathcal{M}}}(\omega) = \min \{ t \in \mathbb{N} \mid s^{\hat{\mathcal{M}}}_{t} \in \mathcal{E}^\mathrm{C}  \} ,
 \label{eq:stopping_time_real_dyn}
 \end{equation}
 \begin{equation}
T^\mathcal{\tilde{M}}(\omega) = \min \{ t \in \mathbb{N} \mid s^{\mathcal{\tilde M}}_{t}  \in \mathcal{E}^\mathrm{C} \},
\label{eq:stopping_time_model}
\end{equation}
with $\mathcal{E}^\mathrm{C}$ the complement of $\mathcal{E}$. The stopping times \eqref{eq:stopping_time_real_dyn} and \eqref{eq:stopping_time_model} denote the first time a rollout under $\hat{\mathcal{M}}$ and $\tilde{\mathcal{M}}$ leaves $\mathcal{E}$, respectively.
Thus, restricting the rollout length to $T(\omega)$ enforces that branched rollouts remain in $\mathcal{E}$.

We show a monotonic improvement similar to \cite{Luo2018Jul, Janner2019Dec, Pan2020Dec}
\begin{equation}
    \eta[\pi] \ge \tilde{\eta}[\pi]-C,
    \label{eq:bound_basic}
\end{equation}
where $\eta[\pi]$ corresponds to the expected return of the policy $\pi$ in $\hat{\mathcal{M}}$, while $\tilde{\eta}[\pi]$ denotes the expected return of $\pi$ under $\tilde{\mathcal{M}}$. As long as the agent improves by more than $C$ in $\tilde{\mathcal{M}}$, we can guarantee improvement in $\hat{\mathcal{M}}$.

\begin{theorem}
\label{theo:return bound}
Suppose the expected return following policy $\pi$ under $\hat{\mathcal{M}}$ is denoted by $\eta[\pi]$ and $\tilde{\eta}[\pi]$ describes the expected return following $\pi$ under $\tilde{\mathcal{M}}$, then we can define a lower bound for $\eta[\pi]$ on $\mathcal{E} \subseteq \mathcal{S}$ of the form 
\begin{equation}
    \eta[\pi] \ge \tilde{\eta}[\pi]-2 r_{\max } \sum_{t=0}^{T(\omega)} \gamma^t \sum_{\tau=0}^t \Delta p_{\mathcal{E}} [\pi],
    \label{eq:theorem_result}
\end{equation}
with
\begin{equation}
    \Delta p_{\mathcal{E}} [\pi]\!:=\!\sup_{s \in \mathcal{E}, a \sim \pi}\!D_{ \mathrm{TV} } \left(p\left(s^{\prime} \mid s, a\right) \| \tilde{p}\left(s^{\prime} \mid s, a\right)\right).
    \label{eq:worst_dyn_mis}
\end{equation}
\begin{proof}
See Appendix \ref{appendix_proofs}, Theorem \ref{theo: return_bound_appendix}.
\end{proof}
\end{theorem}

We define 
\begin{equation}
C :=  2r_{\max } \sum_{t=0}^{T(\omega)} \gamma^t \sum_{\tau=0}^t \Delta p_{\mathcal{E}} [\pi],
\label{eq: C_err_imp_bound}
\end{equation}
which intuitively represents the accumulated worst-case dynamics misalignment. 
The choice of $\mathcal{E}$ thus influences the bound $C$ since the supremum in \eqref{eq:worst_dyn_mis} is taken over $\mathcal{E}$ and $T(\omega)$ directly depends on $\mathcal{E}$.

\subsection{Interpretation of the Result}
In contrast to improvement bounds in previous work \cite{Luo2018Jul, Janner2019Dec, Pan2020Dec}, Theorem \ref{theo:return bound} closely resembles branched model-based rollouts and allows a practical mechanism to be inferred directly.

Rollouts under $\tilde{\mathcal{M}}$ represent the data generated in the practical MBRL algorithm, while $\hat{\mathcal{M}}$ captures the true environment behavior.
Both share $\rho_{\mathrm{BR}} = \mathcal{U}(\mathcal{D}_{\mathrm{env}})$, the start state distribution of model-based rollouts introduced in Algorithm \ref{alg:rollout_basic}, where we assume all states in $\mathcal{D}_{\mathrm{env}}$ to be within $\mathcal{E}$ as depicted in Figure \ref{fig:subset}.
Theorem \ref{theo:return bound} thus bounds the difference in expected return between using the model MDP $\tilde{\mathcal{M}}$ for generating branched model rollouts as compared to performing these rollouts under the environment MDP $\hat{\mathcal{M}}$. Thus, Theorem \ref{theo:return bound} specifically considers model exploitation in branched model-based rollouts that are the predominant data-generating process for training the agent.

 By construction, both processes start from identical start states and deviate from each other based on dynamic misalignment upper bounded by $\Delta p_{\mathcal{E}} [\pi]$ until $T(\omega)$ is reached.
 Unfortunately, the result of Theorem \ref{theo:return bound} is not directly amenable for algorithmic use, as the process $\hat{\mathcal{M}}$ is unknown in practice. Specifically, we are unable to detect when $\hat{\mathcal{M}}$ leaves $\mathcal{E}$ as indicated by $T^{\hat{\mathcal{M}}}(\omega)$ in \eqref{eq:stopping_time_real_dyn} and need an approximation to obtain a practical algorithm.  Assuming trajectories under $\hat{\mathcal{M}}$ and $\tilde{\mathcal{M}}$ stay sufficiently close to each other and replacing $T(\omega)$ with $T^\mathcal{\tilde{M}}(\omega) -1$, we obtain an effective approximation that works well in practice.

\section{Constructing $\mathcal{E}$ from Model Uncertainty}
\label{sec:uncertainty estimate}

In the following, we define 
 $\mathcal{E}$ such that it represents parts of the state space with high model accuracy. There, we can trust our model and learn efficiently. 
\begin{figure*}[tb]
     \centering
     \begin{subfigure}[b]{0.5\columnwidth}
         \centering
         \includegraphics[width=\textwidth]{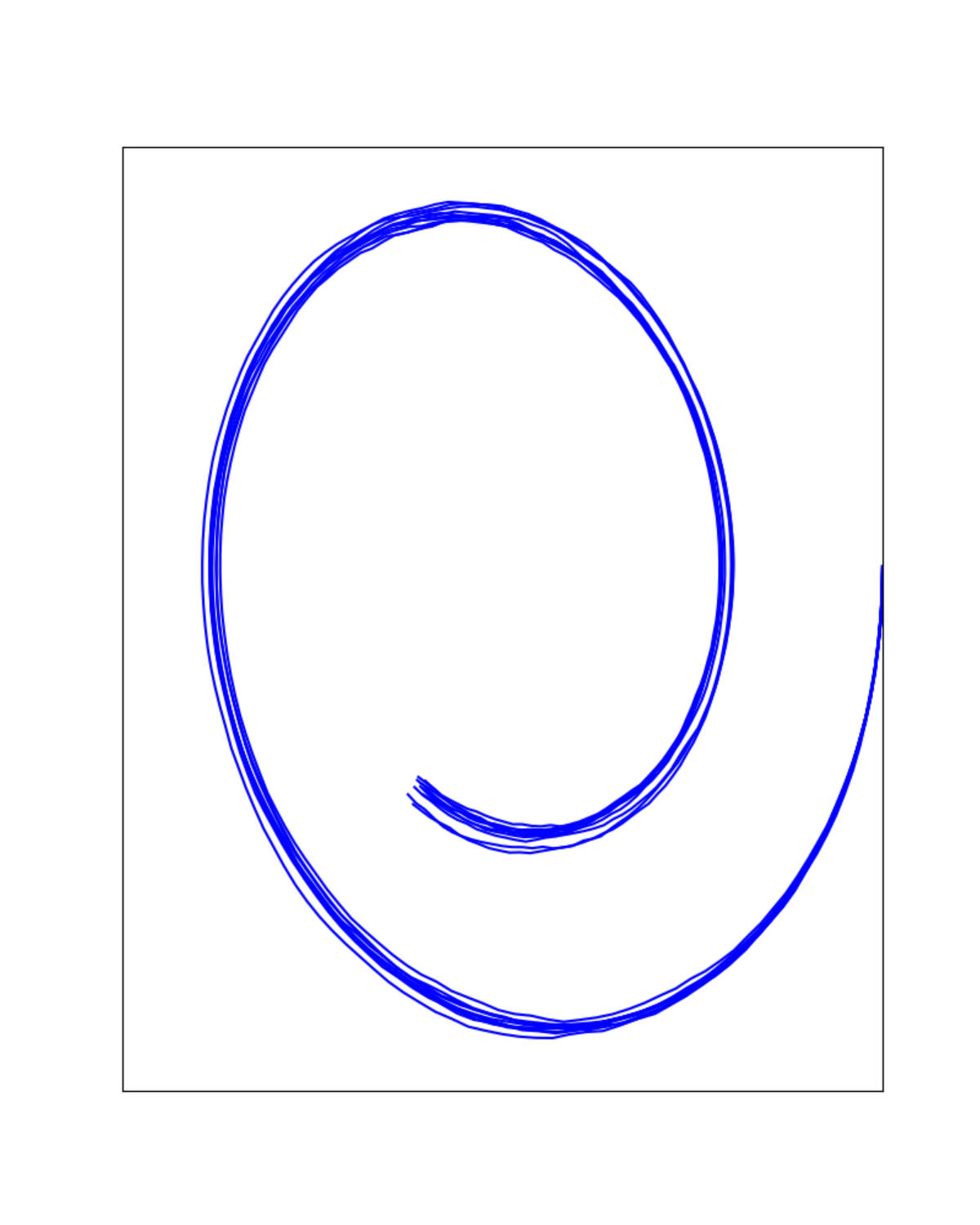}
         \vspace{-2\baselineskip}
         \caption{$\mathcal{D}_{\mathrm{env}}$}
         \label{fig:unc_data}
     \end{subfigure}
     \hfill
     \begin{subfigure}[b]{0.5\columnwidth}
         \centering
         \includegraphics[width=\textwidth]{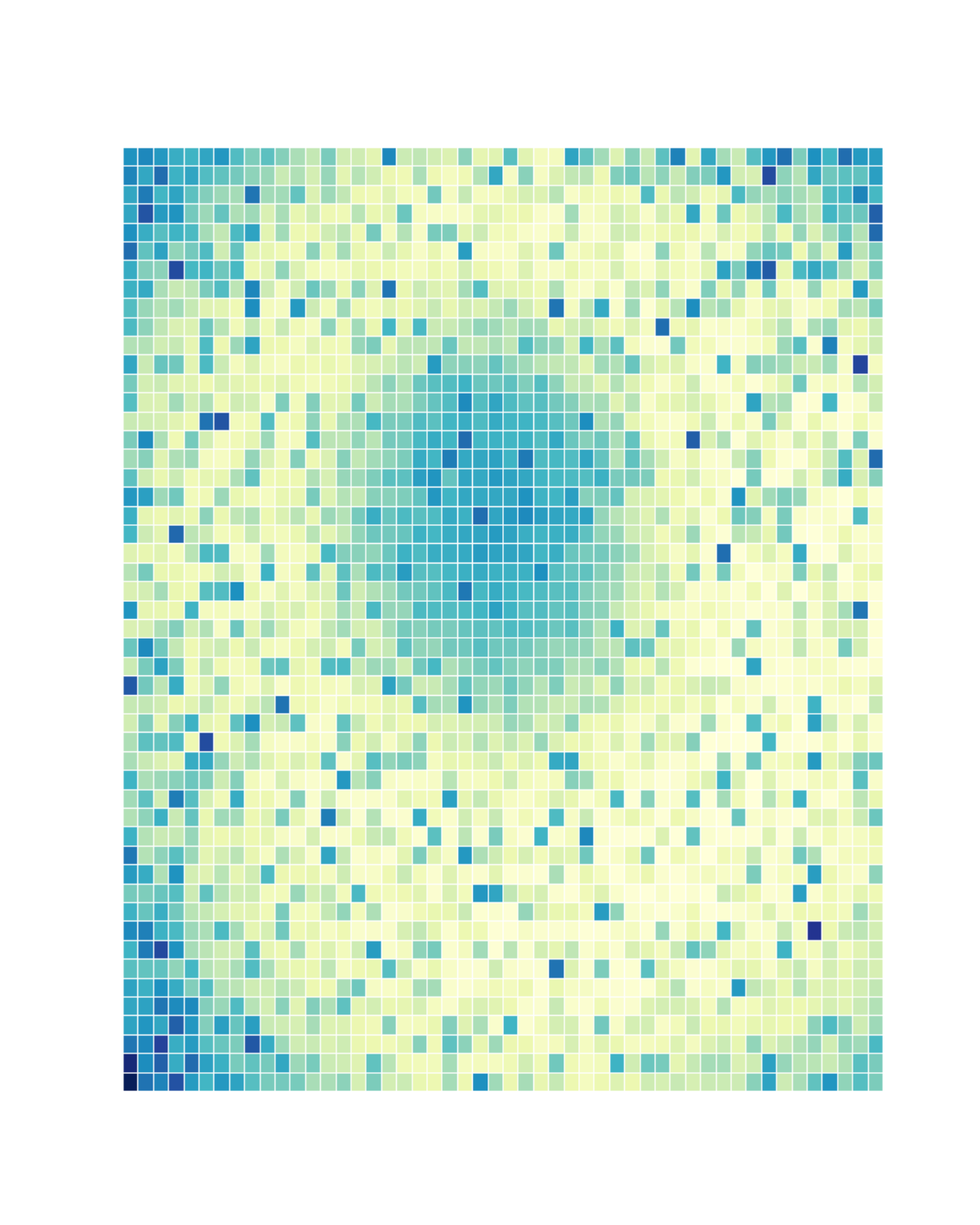}
         \vspace{-2\baselineskip}
         \caption{$D_{\mathrm{TV}}(p(s^{\prime}|s, a) \| \tilde{p}(s^{\prime}|s, a))$}
         \label{fig:unc_tv}
     \end{subfigure}
     \hfill
     \begin{subfigure}[b]{0.5\columnwidth}
         \centering
         \includegraphics[width=\textwidth]{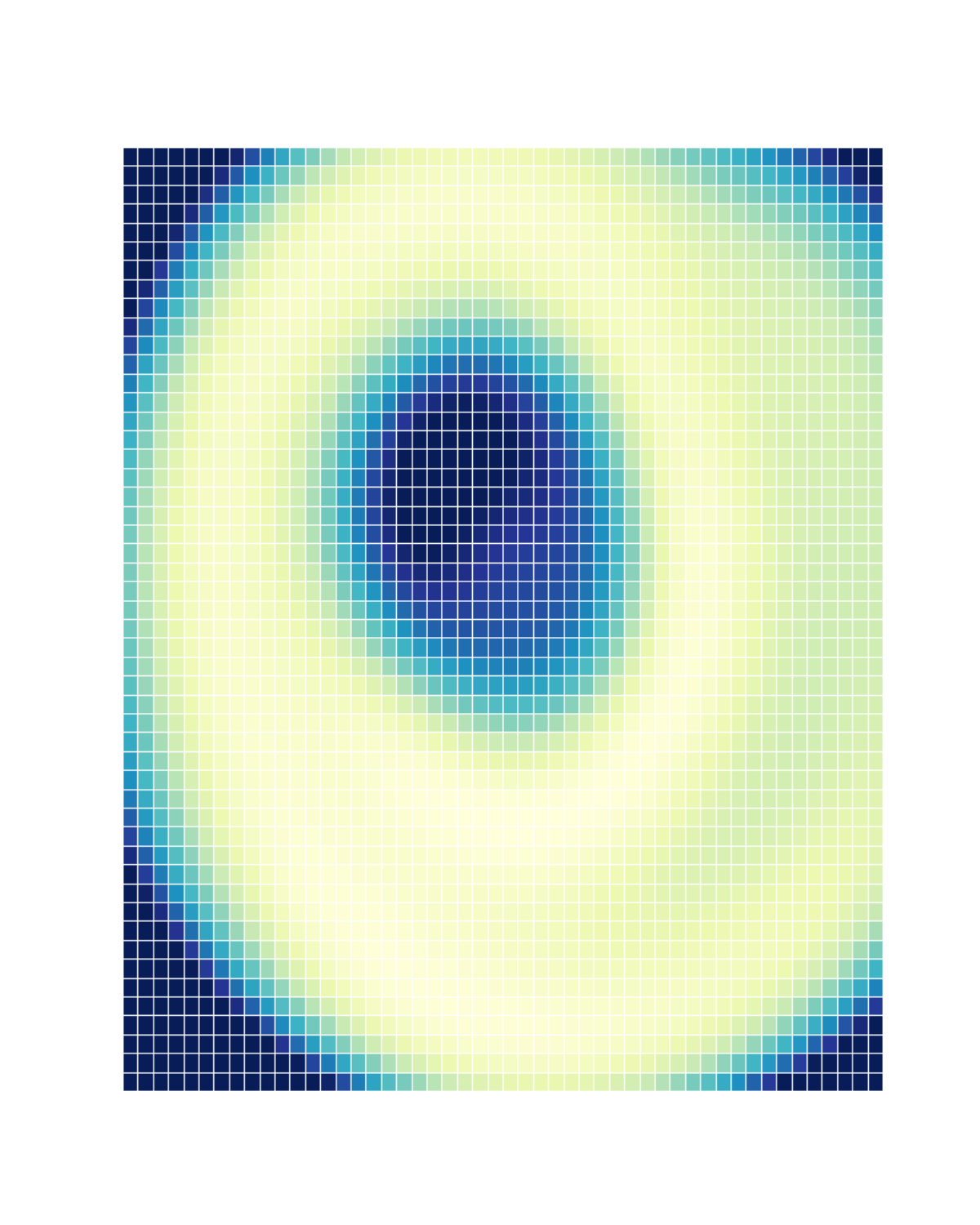}
         \vspace{-2\baselineskip}
         \caption{$u_{\mathrm{GJS}}$}
         \label{fig:unc_ours}
     \end{subfigure}
     \hfill
     \begin{subfigure}[b]{0.5\columnwidth}
         \centering
         \includegraphics[width=\textwidth]{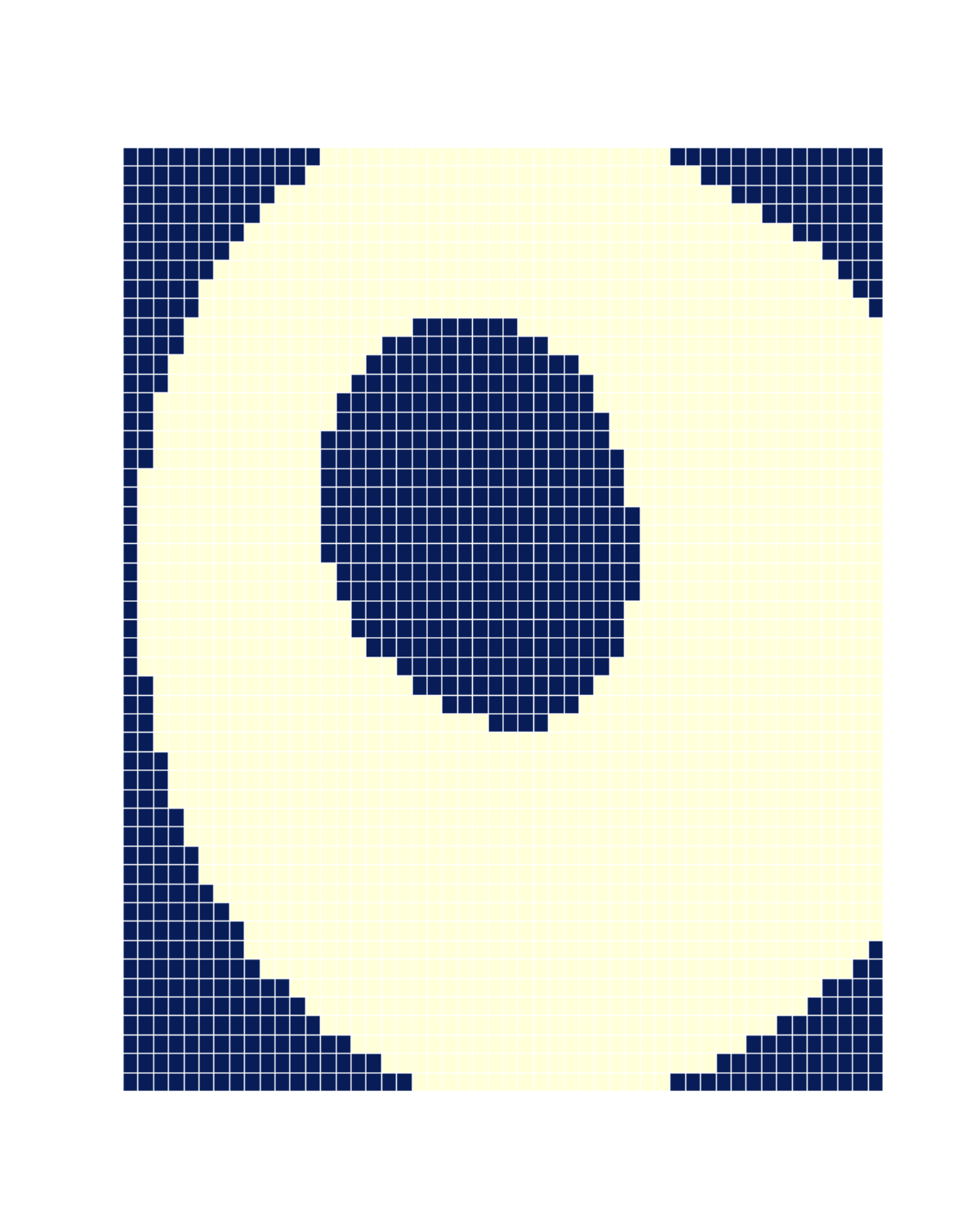}
         \vspace{-2\baselineskip}
         \caption{$\mathcal{E}$ with $u_{\mathrm{GJS}} < \kappa $}
         \label{fig:E_ours}
     \end{subfigure}
        \caption{Constructing $\mathcal{E}$ on a toy example. \textit{ (a) Data to train the PE model. (b) Dynamics misalignment. (c) Proposed measure for model uncertainty \eqref{eq:u_gjs}. (d) Set of sufficient model accuracy to perform branched model-based rollouts \eqref{eq:subset_gjs} }.}
        \label{fig:uncertainty_measures}
        \vspace{-5mm}
\end{figure*}

\subsection{Defining $\mathcal{E}$ in Practice}
Following Theorem \ref{theo:return bound}, ideally we would define 
\begin{equation}
\begin{aligned}
\mathcal{E}^* := \{ s \in \mathcal{S} \mid & D_{ \mathrm{TV} } \left(p\left(s^{\prime} \mid s, a\right) \| \tilde{p}\left(s^{\prime} \mid s, a\right)\right) \leq \kappa, \\
&a \sim \pi(\cdot \mid s) \}
\label{eq: subset_id}
\end{aligned}
\end{equation}
such that dynamics misalignment is upper-bounded by a threshold $\kappa$. This definition of $\mathcal{E}$ yields $\Delta p_{\mathcal{E}} [\pi] \leq \kappa$ in \eqref{eq:worst_dyn_mis} and thus allows $C$ in \eqref{eq: C_err_imp_bound} to be influenced by choosing $\kappa$. 
Estimating the ideal set $\mathcal{E}^*$ is intractable. 
Instead, we leverage these insights and consider a slightly different reformulation that yields a computationally efficient, numerically well-behaved, and meaningful definition of $\mathcal{E}$.

Assuming that the individual PNNs of the PE model have sufficient representational capacity to model $p(s^{\prime} | s, a)$ accurately makes epistemic uncertainty $u_{\mathrm{PE}}$ an expressive quantity for dynamics misalignment.
One option to determine epistemic uncertainty would be using the formulation in \eqref{eq:KL_epistemic} and setting $u_{\mathrm{PE}} = u_{\mathrm{KL}}$.
If all ensemble members agree sufficiently well on a subset of $\mathcal{S} \times \mathcal{A}$, the PE model approximates the environment dynamics to a sufficient degree of accuracy.
Following this reasoning, we construct 
\begin{equation}
\mathcal{E}_{\mathrm{PE}} := \{ s \in \mathcal{S} \mid u_{\mathrm{PE}}(s, a) < \kappa , a \sim \pi(\cdot \mid s) \}
\label{eq: subset}
\end{equation}
such that some uncertainty $u_{\mathrm{PE}}$ under the rollout policy $\pi$ is upper bounded by a threshold $\kappa$.

\subsection{Efficient Measure for Model Uncertainty}
In an algorithmic implementation, efficient computation of $u_{\mathrm{PE}}$ is vital as uncertainty is queried for every model-based transition. The original formulation in \eqref{eq:KL_epistemic} has no closed-form solution, rendering it unsuitable for this purpose.
Instead, we propose an estimate based on the geometric Jensen-Shannon (GJS) divergence \cite{Nielsen2019May}
\begin{equation}
        u_\mathrm{GJS}(s,a) = \frac{2}{E(E-1)} \sum_{e=1}^E \sum_{f=1}^{e-1} D_{\mathrm{GJS}} \left( \mathcal{N}_e \| \mathcal{N}_f \right),
\label{eq:u_gjs}
\end{equation}
with
\begin{equation}
        \mathcal{N}_i =: \mathcal{N}(\mu_{\theta_i}(s,a), \Sigma_{\theta_i}(s,a)).
\end{equation}

The Jensen-Shannon divergence is a symmetrized version of the KL divergence but has no closed-form solution for Gaussian distributions. While losing some of the properties of the Jensen-Shannon divergence, the GJS divergence
\begin{equation}
    D_{\mathrm{GJS}} \left( \mathcal{N}_e \| \mathcal{N}_f \right) = \frac{1}{2} D_{\mathrm{KL}} \left( \mathcal{N}_e \| \mathcal{N}_{ef} \right) + \frac{1}{2} D_{\mathrm{KL}} \left( \mathcal{N}_f \| \mathcal{N}_{ef} \right)
\end{equation}
with
\begin{equation}
    \Sigma_{ef} = \left( \frac{1}{2} \left(\Sigma_{\theta_e}(s,a)\right)^{-1} + \frac{1}{2} \left(\Sigma_{\theta_f}(s,a)\right)^{-1} \right)^{-1}
\end{equation}
and
\begin{equation}
\begin{aligned}
    \mu_{ef} = \Sigma_{ef} & \left(  \frac{1}{2} \left(\Sigma_{\theta_e}(s,a) \right)^{-1} \mu_{\theta_e}(s,a) \right.\\
    & \left. + \frac{1}{2} \left(\Sigma_{\theta_f}(s,a) \right)^{-1} \mu_{\theta_f}(s,a)\right)
\end{aligned}
\end{equation}
recovers a closed-form solution for Gaussian distributions.
This allows us to compute model uncertainty in a closed form replacing the comparison to the Gaussian mixture distribution in \eqref{eq:KL_epistemic} with a pairwise comparison between PNN predictions. Exploiting the symmetry of the GJS divergence allows to reduce the number of pairwise comparisons.

Thus, $u_\mathrm{GJS}$ yields an efficient-to-compute uncertainty measure, leading to a practical definition of
\begin{equation}
\mathcal{E} := \{ s \in \mathcal{S} \mid u_{\mathrm{GJS}}(s, a) < \kappa , a \sim \pi(\cdot \mid s) \}
\label{eq:subset_gjs}
\end{equation}
that is directly applicable to algorithmic use.

\subsection{Illustrative Example}
\label{sec:u_gjs_illustration}
 As an illustrative example, we use a pendulum with known dynamics and a two-dimensional state. Figure \ref{fig:uncertainty_measures} visualizes the construction of $\mathcal{E}$ according to \eqref{eq:subset_gjs} for this system. An in-depth description is provided in Appendix \ref{appendix:toy_example}.

We create a $\mathcal{D}_{\mathrm{env}}$ with a characteristic spiral form by performing rollouts with a feedback controller $\pi_{\mathrm{FL}}$ as depicted in Figure \ref{fig:unc_data}, and use the data to train a PE dynamics model. In the following, dynamic misalignment and uncertainty of this model under $\pi_{\mathrm{FL}}$ are analyzed over $\mathcal{S}$ using heat maps.

Dynamics misalignment as used in \eqref{eq: subset_id} is depicted in Figure \ref{fig:unc_tv}, where a clear trend can be observed: dynamic misalignment is low, indicated in yellow, close to data in $\mathcal{D}_{\mathrm{env}}$ but grows further away, where blue corresponds to high misalignment. For approximations of the total variation distance we use the common upper bound \eqref{eq:upper_bound_TV} in Appendix \ref{sec:toy_example_ugjs_test}.

Figure \ref{fig:unc_ours} shows the proposed model uncertainty $u_{\mathrm{GJS}}$ used in \eqref{eq:subset_gjs}. We observe a smooth behavior of the uncertainty estimate and rediscover the data distribution in $\mathcal{D}_{\mathrm{env}}$ for the lowest uncertainty values indicated in bright yellow. Most importantly, areas of low uncertainty coincide with areas of low dynamics misalignment in Figure \ref{fig:unc_tv}.

As depicted in Figure \ref{fig:E_ours}, choosing a suitable threshold $\kappa$ allows $\mathcal{E}$ to be defined such that a considerable portion of $\mathcal{S}$ can be explored in model-based rollouts while avoiding areas of high dynamics misalignment. Here $\mathcal{E}$ is visualized in yellow, while blue regions indicate $\mathcal{E}^{\mathrm{C}}$.

\section{MACURA: Model-Based Actor-Critic with Uncertainty-Aware Rollout Adaption}
\label{sec:macura_algorithm}
Combining the insights into how to choose a set $\mathcal{E}$ to enforce monotonic improvement, discussed in Section \ref{sec: improvement bound}, with those into how to construct $\mathcal{E}$ using model uncertainty, detailed in Section \ref{sec:uncertainty estimate}, we present an uncertainty-aware adaption scheme for model-based rollouts. We further discuss expanding $\mathcal{E}$ by efficiently exploring the environment.
These building blocks lead to the Model-Based Actor-Critic with Uncertainty-Aware Rollout Adaption (MACURA) algorithm.

\subsection{Uncertainty-Based Rollout Adaption}
\label{subsec:rollout_scheme}

The set $\mathcal{E}$ depends on the uncertainty threshold $\kappa$ \eqref{eq:subset_gjs}. Thus, an appropriate choice of $\kappa$ is critical in algorithmic design.
 We propose an adaptive mechanism for determining $\kappa$ that transfers to different applications and stages of training.

In the algorithm, $M$ branched model-based rollouts are performed in parallel. As these start from states in $\mathcal{D}_{\mathrm{env}}$, we evaluate the model uncertainty after the first prediction step to update $\kappa$ proportionally to the current model uncertainty.

We therefore define the base uncertainty $\hat{u}_{\mathrm{GJS}, k}$ to be the $\zeta$ quantile of the $M$ uncertainty measures after the first prediction step at the $k^{\mathrm{th}}$ round of model-based rollouts. We use this heuristic as an upper bound on what can be considered certain.
Further, we introduce $\xi \in \mathbb{R}^+$ as a tunable scaling factor, which allows $\mathcal{E}$ to be increased or shrunk by scaling $\hat{u}_{\mathrm{GJS}}$ up or down. To stabilize $\kappa$ over iterations, we define it to be the average scaled base uncertainty:
\begin{equation}
   \kappa = \frac{\xi}{K} \sum_{k=1}^K \hat{u}_{\mathrm{GJS}, k},
   \label{eq:kappa_update}
\end{equation}
with $K$ the rounds of rollouts performed thus far.

Theorem \ref{theo:return bound} indicates that the difference in expected return accumulates over the rollout length, even with bounded dynamics misalignment. 
Since we only provide a pointwise bound for each transition step, we still need to enforce a maximum rollout length
$T_{\mathrm{max}}$, larger than the typical length of adapted rollouts, to avoid extensive error accumulation.

During the experimental evaluation in Section \ref{sec:experiments}, we see that predefined values of $T_{\mathrm{max}} = 10$ and $\zeta = 95\%$ perform well across different applications, requiring no environment-specific tuning. Thus, $\xi$ is a single, interpretable hyperparameter defining the rollout scheme.
This makes the proposed uncertainty-aware rollout adaption mechanism formulated in Algorithm \ref{alg:rollout} considerably easier to tune than those in existing work \cite{Janner2019Dec, Pan2020Dec}.

\begin{algorithm}[tb]
\caption{Uncertainty-Aware Adapted Branched Model-Based Rollouts }\label{alg:rollout}
\begin{algorithmic}
    \STATE Given $\mathcal{D}_{\mathrm{env}}$, $\mathcal{D}_{\mathrm{mod}}$, $\tilde{p}_{\theta_{1, \dots, E}}$, $\pi$, $\zeta$, $\xi$, and $T_{\mathrm{max}}$
    \STATE $s_0 \sim \mathcal{U} (\mathcal{D}_{\mathrm{env}})$
    \FOR {$t =0, \dots, T_{\mathrm{max}}-1$}
    \STATE $e_t \sim \mathcal{U}(1, \dots, E)$
    \STATE $a_t \sim \pi(\cdot \mid s_t)$
    \STATE $s_{t+1} \sim \tilde{p}_{\theta_{e_t}}(\cdot \mid s_t, a_t)$
    \STATE $r_{t+1} = r(s_t, a_t)$
    \STATE $u_\mathrm{GJS}(s_t, a_t)$ according to (\ref{eq:u_gjs})
    \IF{$t=0$}
    \STATE update $\kappa$ according to \eqref{eq:kappa_update}
    \ENDIF
    \IF{$u_\mathrm{GJS}(s_t, a_t) < \kappa$}
    \STATE $\mathcal{D}_{\mathrm{mod}} \gets \mathcal{D}_{\mathrm{mod}} \cup \{ (s_t, a_t, r_{t+1}, s_{t+1} )\}$
    \ELSE
    \STATE \textbf{break}
    \ENDIF
    \ENDFOR
\end{algorithmic}
\end{algorithm}
Due to its variable rollout length, the proposed model-based rollout scheme produces varying amounts of data to train the model-free agent.
As the update-to-data ratio plays a crucial role in the stability of model-free RL \cite{Chen2021}, we adapt the number of update steps 
\begin{equation}
    G = \Bigl\lfloor G_{\mathrm{max}} \frac{|\mathcal{D}_{\mathrm{mod}}|}{|\mathcal{D}_{\mathrm{mod}}|_{\mathrm{max}}} \Bigr\rceil
    \label{eq:G_adaption}
\end{equation}
to the SAC agent according to the amount of data in the model buffer $|\mathcal{D}_{\mathrm{mod}}|$ compared to its capacity $|\mathcal{D}_{\mathrm{mod}}|_{\mathrm{max}}$. We use this ratio to scale the maximum number of update steps $G_{\mathrm{max}}$ and round the result to the nearest integer. An empirical analysis of how the update-to-data ratio interacts with varying rollout lengths is provided in Appendix \ref{sec:app_exp_rollout}.

\subsection{Expanding $\mathcal{E}$ through Environment Exploration}
Through the notion of $\mathcal{E}$, MACURA has a reliable estimate of where to trust the model $\tilde{p}$ that is trained on $\mathcal{D}_{\mathrm{env}}$. In line with common understanding in system 
identification \cite{Ljung1998}, more informative data yields a better model. Thus, employing effective exploration mechanisms to generate a meaningful $\mathcal{D}_{\mathrm{env}}$ will improve the model and thus expand $\mathcal{E}$, increasing the effectiveness of model-based rollouts.

Although we do not place a strong focus on different exploration mechanisms for Dyna-style MBRL in this work, we tested different approaches. As we discuss in Section \ref{sec:experiments}, we see MACURA perform particularly well with pink exploration noise \cite{eberhard2023pink}, which introduces a certain degree of temporal correlation between consecutive actions blending white noise and Brownian motion.

We combine the mechanisms discussed above in MACURA. Pseudocode is provided in Algorithm \ref{alg:macura} of Appendix \ref{appendix:algos}. 

\begin{figure*}[tb]
     \centering
    \includegraphics[width=0.95\textwidth]{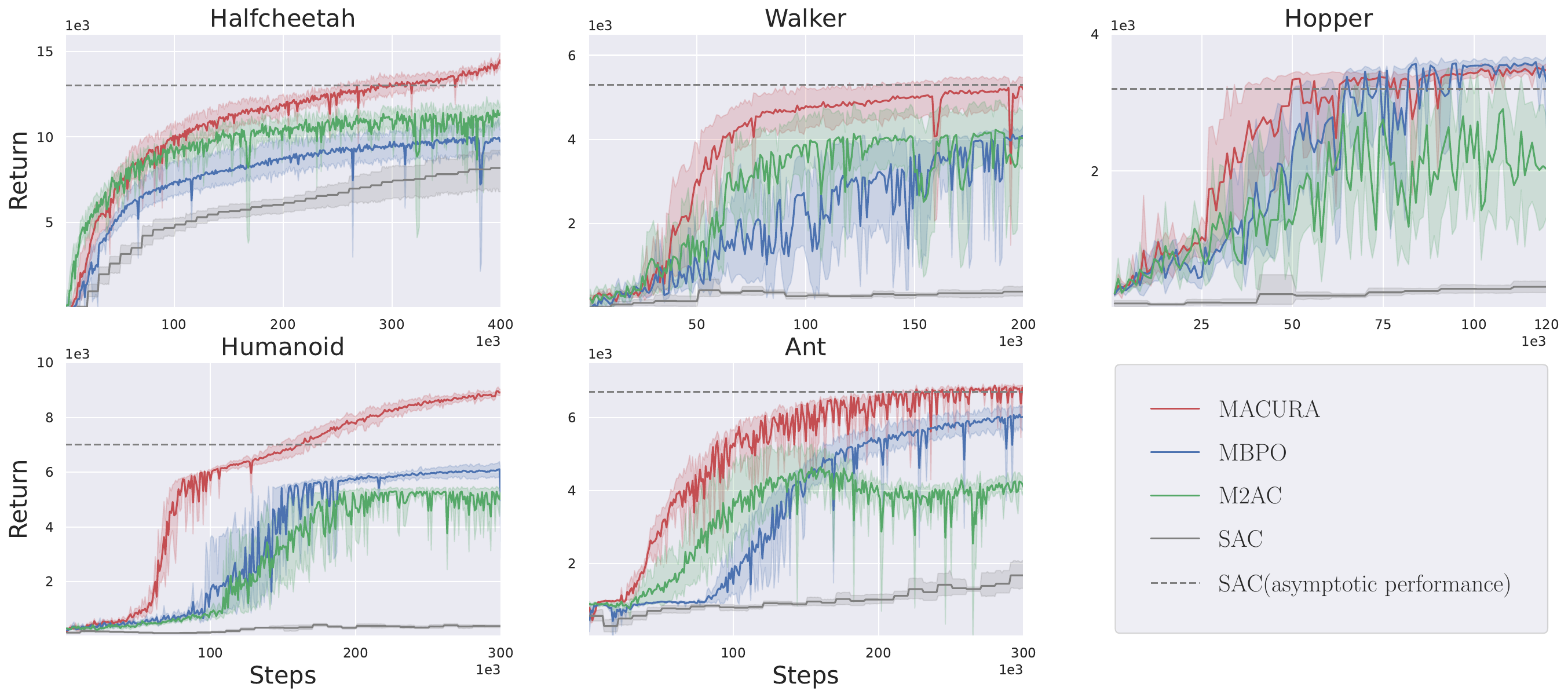}
\caption{Performance on the MuJoCo Benchmark. \textit{MACURA shows substantial improvements in data efficiency and asymptotic performance over state-of-the-art Dyna-style MBRL approaches (MBPO, M2AC) in most tasks. Most noticeably, MACURA is on par with or outperforms the asymptotic performance of the model-free SAC baseline.}}
 \label{fig:mujoco}
 \vspace{-5mm}
\end{figure*}

\section{Experiments and Discussion}
\label{sec:experiments}
Next, we evaluate MACURA on the MuJoCo \cite{Todorov} benchmark.
A direct comparison to state-of-the-art Dyna-style methods reveals a substantial improvement in data efficiency and asymptotic performance.
In an ablation study, we investigate the influence of different exploration schemes on Dyna-style approaches and show that MACURA with pink noise exploration yields the best overall results.
We further discuss how to tune the scaling parameter $\xi$ of MACURA and show its robustness to a wide range of values, substantially reducing the tuning effort as compared to prior work in Dyna-style MBRL. 

\subsection{Experimental Setup} 
We compare MACURA to model-based MBPO \cite{Janner2019Dec} and M2AC \cite{Pan2020Dec} approaches as well as the SAC \cite{Haarnoja2018Dec} algorithm, which represents the model-free learner in all of the methods above.
All implementations\footnote{The code is available online: \url{https://github.com/Data-Science-in-Mechanical-Engineering/macura}} are based on the recent mbrl-lib library \cite{Pineda2021Apr}.
A detailed description of the experimental setup is provided in Appendix \ref{sec:app_exp_experimental_setup}.

\subsection{Performance Evaluation}
The results on the MuJoCo benchmark are depicted in Figure \ref{fig:mujoco}.
We see that MACURA learns substantially faster than MBPO and M2AC, especially in high-dimensional environments such as Humanoid and Ant. MACURA further shows considerably stronger asymptotic performance than other model-based approaches
. Remarkably, MACURA frequently even outperforms SAC.
All model-based methods learn substantially faster than SAC. MBPO with fine-tuned rollout schedules can compete with M2AC.

\subsection{The Importance of Environment Exploration}
\begin{figure}[]
     \centering
         \includegraphics[width=\columnwidth]{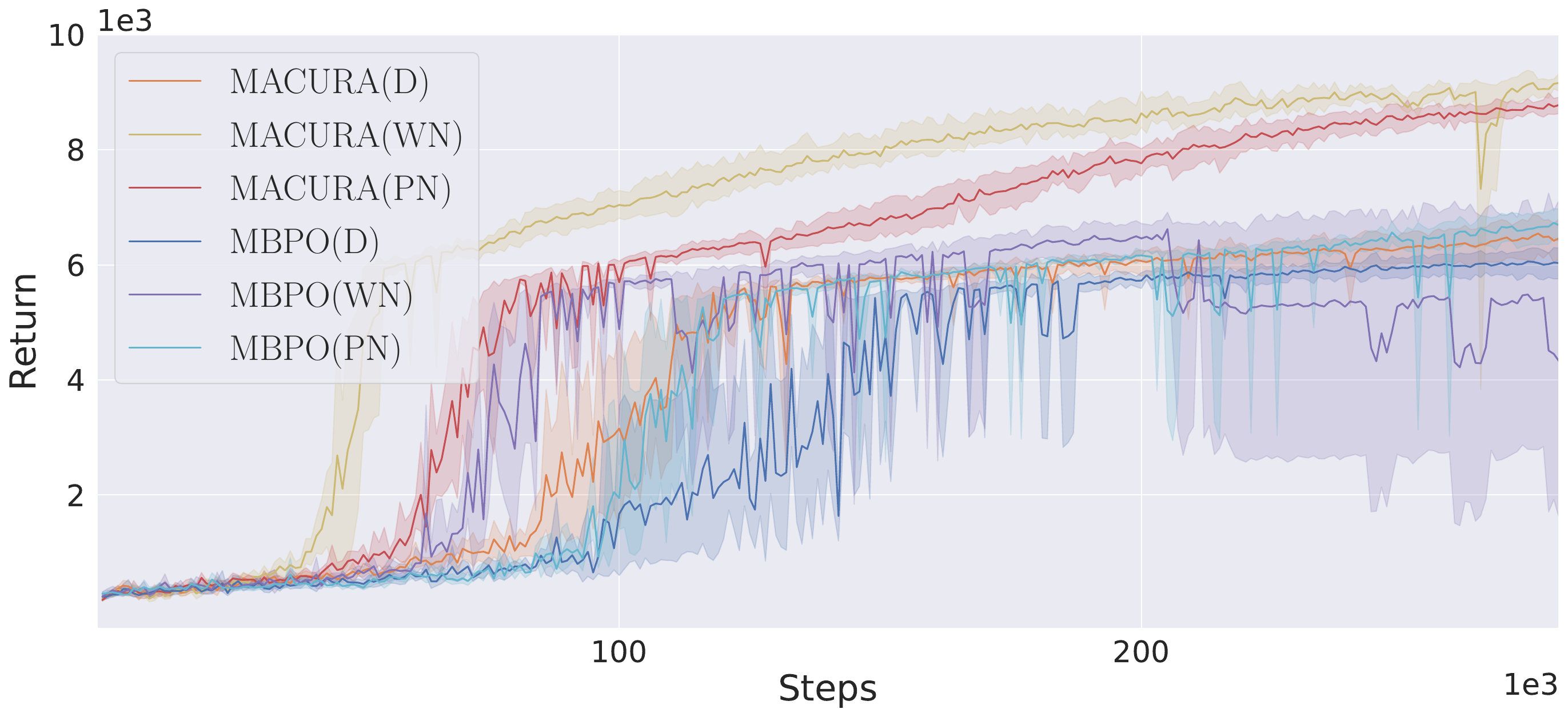}
         \caption{Exploration Schemes on MACURA and MBPO. \textit{Impact of deterministic (D), white noise (WN), and pink noise (PN) exploration on algorithmic performance.}}
         \label{fig:humanoid_pink} 
         \vspace{-5mm}
\end{figure}

Following the state-of-the-art implementation \cite{Pineda2021Apr}, the agent interacts deterministically with the environment in MBPO and M2AC, while we choose pink noise exploration for MACURA in Figure \ref{fig:mujoco}. The impact of exploration on MACURA and MBPO  is illustrated in Figure \ref{fig:humanoid_pink} for the Humanoid task. We observe a general trend that is discussed in more detail in Appendix \ref{sec:app_exp_exploration}. For both algorithms, deterministic interaction yields limited yet stable performance. Classic white noise exploration, where actions are sampled independently, has the potential of strong performance, as can be seen in MACURA, while introducing the risk of destabilizing learning, as in the case of MBPO. Pink noise exploration, instead, shows an intermediate, overall best behavior, combining high performance with stable learning. The strongest gains from environment interaction can be observed for MACURA, underscoring the effectiveness of the notion of $\mathcal{E}$ for model usage.

\subsection{Tuning the Uncertainty-Aware Adaption Scheme}
\label{sec:exp_xi}
\begin{figure}[tb]
     \centering
         \includegraphics[width=\columnwidth]{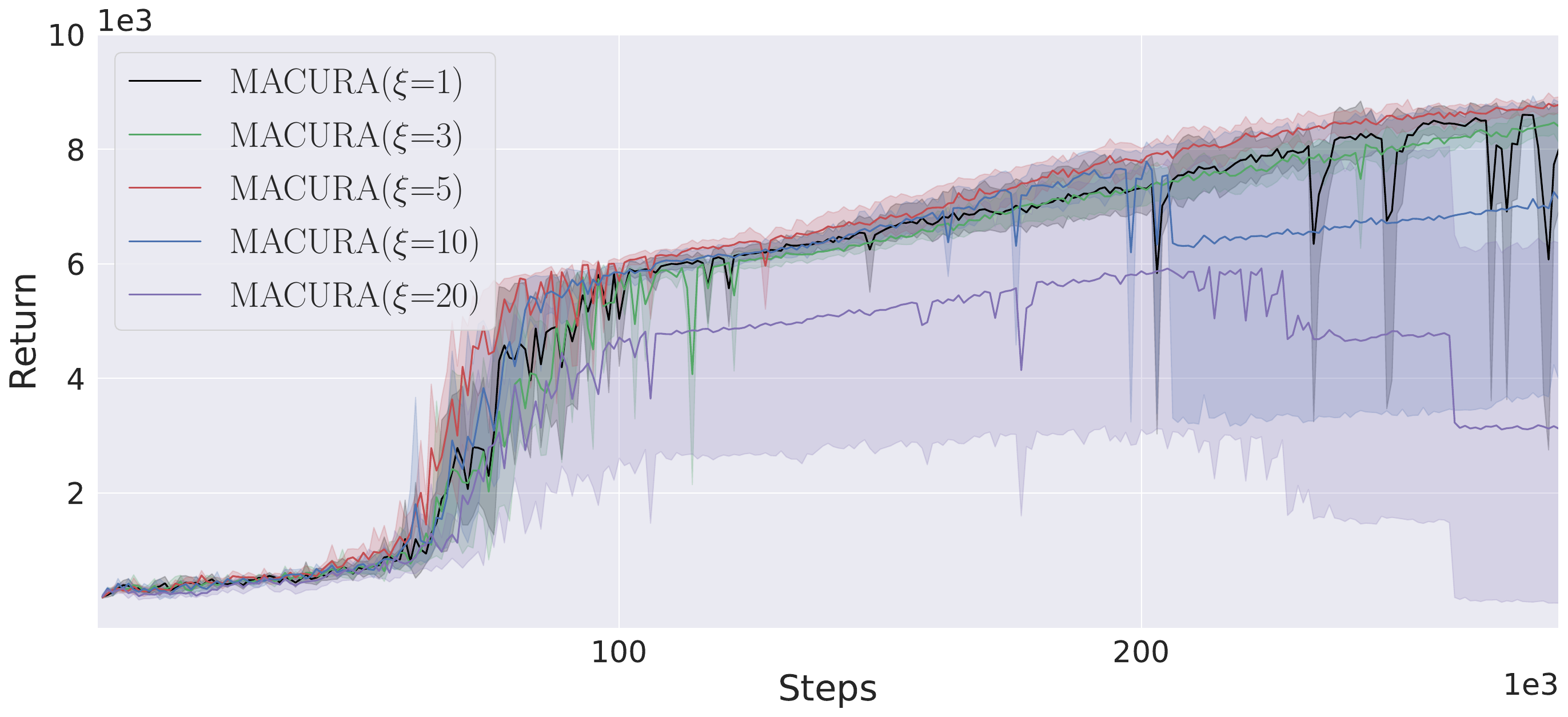}
         \caption{Tuning scaling parameter $\xi$. \textit{MACURA performs well for intermediate $\xi$. Too large values of $\xi$ lead to model exploitation, too small values lead to overfitting of the agent.}}
         \label{fig:humanoid_kappa}
         \vspace{-5mm}
\end{figure}
The uncertainty-aware rollout adaption scheme comprises three hyperparameters. We set $T_{\mathrm{max}} = 10$ and $\zeta = 95\%$ for all tasks and never tune them any further \footnote{While potential performance gains are possible tuning $T_{\mathrm{max}}$ and $\zeta$, our focus is to underscore the ease of tuning MACURA.}. However, the scaling factor $\xi$ requires application-specific tuning.

Figure \ref{fig:humanoid_kappa} shows the influence of different choices of $\xi$ on the performance on Humanoid. While we see that MACURA performs well for intermediate values for $\xi$, learning destabilizes for both high and low choices. This is expected for high values of $\xi $ as they enforce model exploitation.
However, that low values of $\xi$ also destabilize training is more surprising. We assume that the on-policy nature of Dyna-style MBRL, discussed in Section \ref{subsec: dyna-style_mbrl}, in combination with the effective rollout adaption mechanism of MACURA leads to narrow data distributions in $\mathcal{D}_{\mathrm{mod}}$ when restricting model uncertainty too harshly.
Consequently, the Q-functions of the model-free agent overfit, which destabilizes learning.
A practical guide on tuning $\xi$ is provided in Appendix \ref{sec:app_exp_xi}.

\subsection{Ablation and Further Results}
\begin{figure}[tb]
     \centering
    \includegraphics[width=\columnwidth]{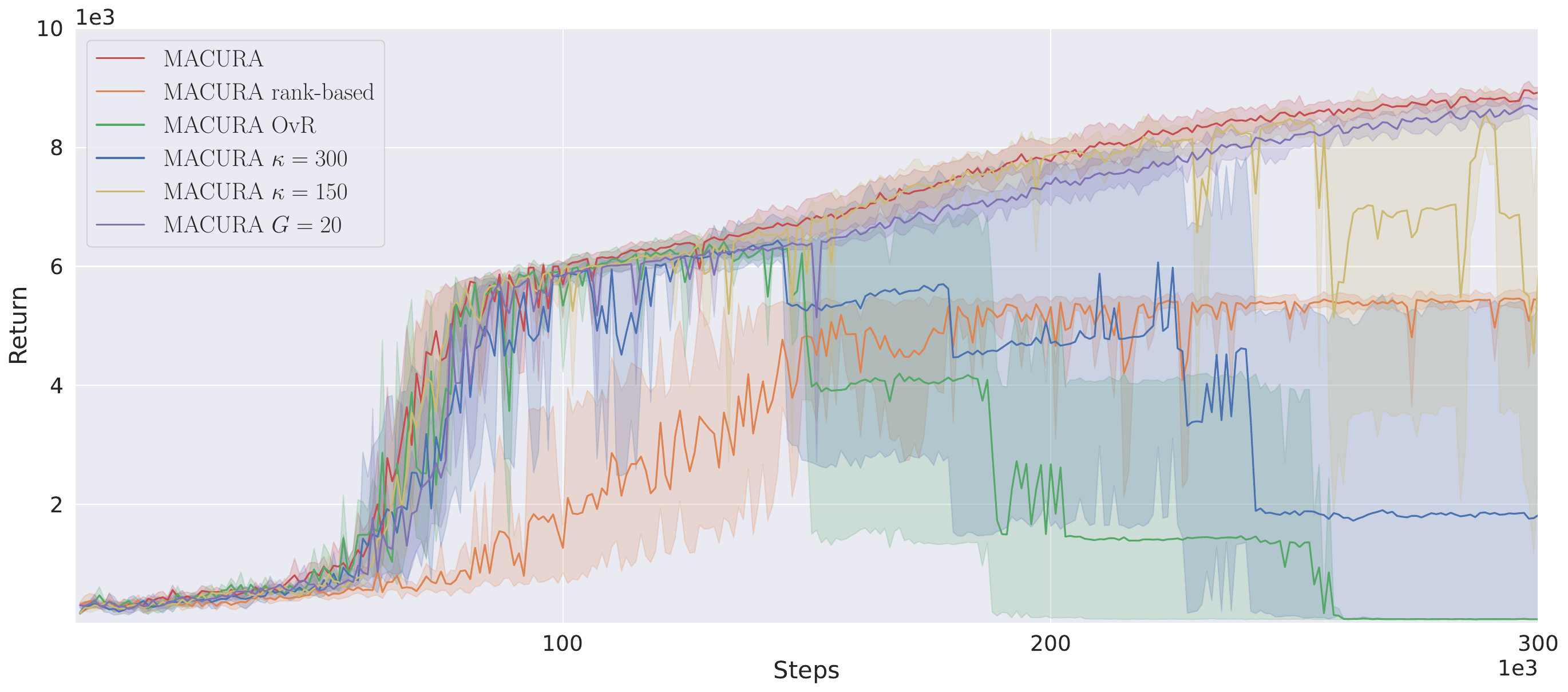}
\caption{Ablation on MACURA building blocks. \textit{The rank-based rollout mechanism with a stable uncertainty estimate \eqref{eq:u_gjs} and adaptive threshold \eqref{eq:kappa_update} yields strong performance. }}
 \label{fig:return_abl}
 \vspace{-3mm}
\end{figure}

Finally, we provide an ablation study on the building blocks of MACURA on the Humanoid task in Figure \ref{fig:return_abl}. The analysis reveals that the performance gain of MACURA mostly stems from the combination of a threshold-based rollout length adaption mechanism, the self-tuning threshold $\kappa$, and the reliable GJS uncertainty estimate.

Replacing the threshold-based rollout adaption of MACURA with the rank-based heuristic proposed in \cite{Pan2020Dec} substantially reduces performance.
The threshold-based mechanism, however, requires a reliable uncertainty estimate as in \eqref{eq:u_gjs} and an adaptive threshold $\kappa$ as presented in \eqref{eq:kappa_update} for stable learning. 

Replacing the GJS uncertainty estimate with One-vs-Rest (OvR) uncertainty \cite{Pan2020Dec} leads to divergence. We assume this is attributed to the brittleness of the OvR uncertainty estimate as illustrated in Appendix \ref{sec:toy_example_ugjs_test}. Thus, faulty predictions are frequently used for training.

\begin{figure}[tb]
     \centering
    \includegraphics[width=\columnwidth]{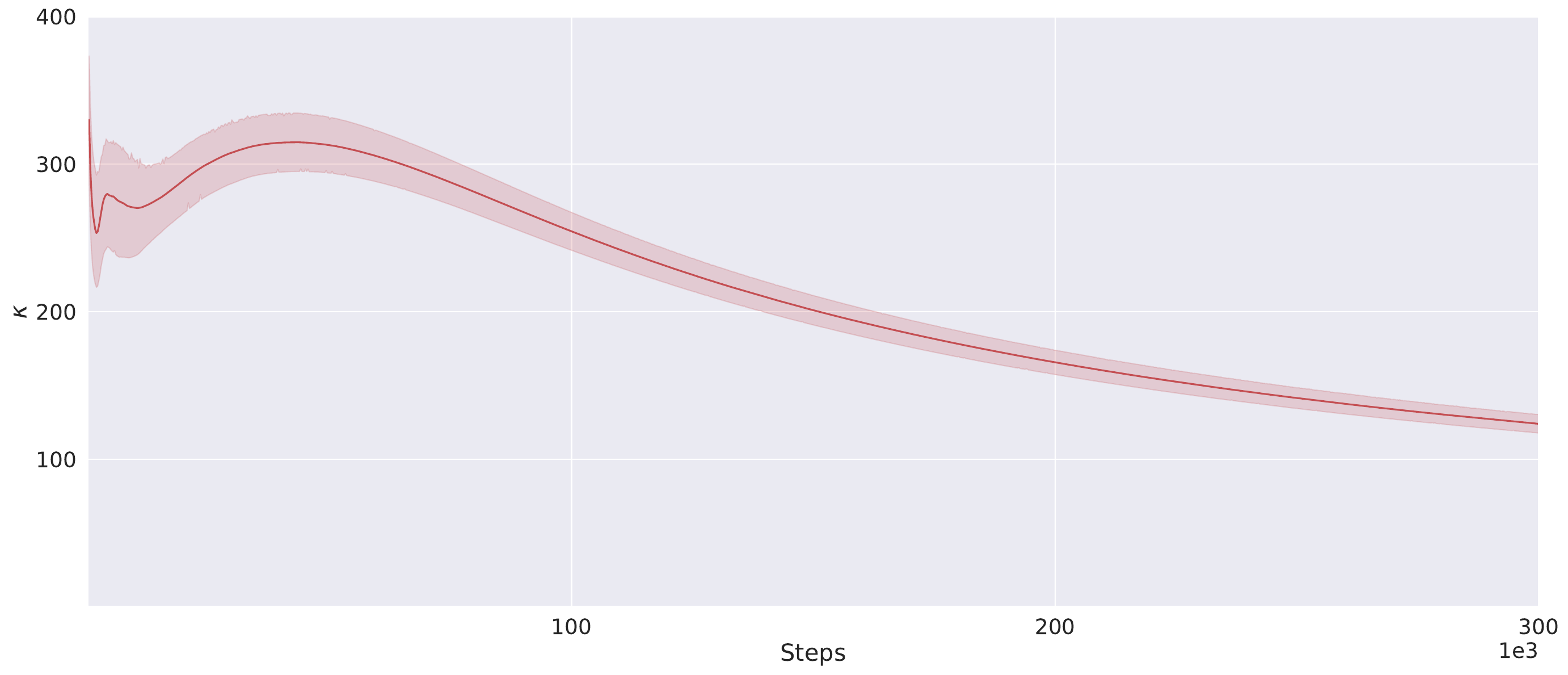}
\caption{Uncertainty threshold $\kappa$. \textit{MACURA initially keeps $\kappa$ comparatively high and reduces $\kappa$ over time as more precise information is required to refine the policy. }}
\vspace{-5mm}
 \label{fig:kappa_abl}
\end{figure}

Figure \ref{fig:kappa_abl} depicts the adaption of $\kappa$ according to \eqref{eq:kappa_update} for the standard MACURA runs in Figure \ref{fig:return_abl}. Following our intuition, $\kappa$ is comparatively high in the early stages of training, as uncertain data is sufficient to learn an initial policy. Throughout training, $\kappa$ decreases as more precise information is required to further improve the policy. Keeping $\kappa$ fixed at $300$ and $150$, respectively, destabilizes learning at different stages of training due to model exploitation.

Replacing the gradient step adaption in \eqref{eq:G_adaption} with a fixed number of gradient steps $G=20$ yields moderate performance reduction on Humanoid, while deterministic environment interaction hinders an effective expansion of $\mathcal{E}$ and thus substantially reduces performance.

We provide further experimental results in Appendix \ref{appendix:experiments}. Performance in tasks with process noise is presented in\ref{sec:app_exp_noisy}, \ref{sec:app_exp_rollout} provides detailed results for rollout length and update step adaption, \ref{sec:app_exp_uncertainty_est} discusses the impact of different uncertainty estimates, and \ref{sec:app_exp_long_exp} provides results for long experiments.

\section{Related Work}
\label{sec:related_work}

The selection of the model class for dynamics approximation is an important design decision in MBRL. Early works focus on local linear models \cite{Levine2013May, Gu2016Jun} and Gaussian processes \cite{Deisenroth2011Jun}. Transferring MBRL to more complex problems, however, requires models with higher capacity, such as neural networks (NN) \cite{Nagabandi2018, Williams2017May}. As deterministic NNs bear the risk of overfitting, probabilistic NN architectures \cite{gal2016improving, Buckman2018Dec, Kurutach2018Feb, Depeweg2016May} have proven a suitable model class for MBRL. The PE model \cite{Lakshminarayanan2017Dec} is a particularly successful architecture in this domain and has led to several breakthroughs in MBRL \cite{Chua2018, Janner2019Dec, Yu2020}. While world models \cite{Ha2018Mar, Hafner2019May} represent the current state-of-the-art in vision-based MBRL, the PE model is considered state-of-the-art when 
learning from the physics state of a system. As we are concerned with the latter, we focus on uncertainty quantification of the PE model.

Different approaches have been adopted to combine the RL objective \eqref{eq:RL_objective} with information from a model.
Analytic gradient methods \cite{Deisenroth2011Jun, Levine2013May, Hafner2020Apr} optimize a policy propagating through model-based experience, while 
model-based planning algorithms \cite{Williams2017May, Nagabandi2018, Chua2018} use the model for receding horizon control. Other methods combine model-free RL approaches with a model. In value expansion approaches \cite{Feinberg2018Feb, Buckman2018Dec} the model is used to stabilize update targets of the Q-function, whereas in Dyna-style approaches \cite{Sutton1991Jul, Kalweit2017Oct, Janner2019Dec} the model generates data to train the model-free agent. In vision-based MBRL, analytic gradient methods \cite{Hafner2020Apr, Hafner2021, Hafner2023Jan} represent the state-of-the-art, while Dyna-style approaches yield state-of-the-art results in terms of data efficiency and asymptotic performance for MBRL with compact state representations.
 
 A variety of improvements to the general idea of deep Dyna-style MBRL \cite{Janner2019Dec} have been proposed. These comprise online parameter tuning \cite{Lai2021Nov}, reduction of model error during model-based rollouts \cite{Frohlich2021, Lai2020Nov, Shen2020Dec}, improved exploration in the model \cite{Morgan2021May, Zhang2020Jun}, model learning \cite{Ji2022Nov, Wang2023Jul, Wu2022Nov}, and consideration of model uncertainty in the model-free Q-function \cite{Luis2023Feb, Luis2023Dec, Wang2021Dec}.
All the above are orthogonal to our method.

Using uncertainty in the model or the RL agent to control model usage is a common technique in MBRL. One approach is to use model data where the agent is uncertain \cite{Kalweit2017Oct, nguyen2018}. Model-based offline RL methods \cite{Yu2020, Zhai2024Jan, Jeong2022Oct} often construct a pessimistic MDP that penalizes model uncertainty in the value function. \citet{Zhang2021Aug} adapt rollout length in a multi-agent setting based on the error in the policy model of opponent agents. In value expansion methods, rollout steps are frequently reweighted based on model uncertainty \cite{Buckman2018Dec, Jeong2022Oct, Vuong2019Jun}. Further, \citet{Abbas2020Jul} address uncertainty due to model inadequacy.

Despite the importance of reliable model-based rollouts in Dyna-style MBRL, the idea of adapting rollouts based on model accuracy has received little attention. An exception is the M2AC algorithm \cite{Pan2020Dec}, which is closest to this work. M2AC schedules rollout lengths using a rank-based filtering heuristic depending on model uncertainty.
They introduce a reward penalty for model uncertainty similar to offline MBRL approaches, which can, however, hinder the expansion of the known subset of $\mathcal{S}$ in an online setting. Further, the uncertainty estimate of M2AC is comparatively brittle as presented in Appendix \ref{sec:toy_example_m2ac_comparison}. The proposed rollout scheme of MACURA, instead, takes a strictly spatial perspective on rollout length which is conceptually different from M2AC and generally new in Dyna-style MBRL.

\section{Conclusion}

Learning predictive dynamics models to tame data requirements of RL is an established concept, leading to state-of-the-art MBRL approaches like MBPO that achieve high data efficiency and asymptotic performance.
This work builds on the successful MBPO architecture, addressing the critical question of adapting the length of model-based rollouts.
While it is common knowledge that models are only helpful \emph{where} they are accurate, and they are accurate only \emph{where} they have seen data, only few works in MBRL address model accuracy in general, and its spatial nature in particular.
We make the consideration of model accuracy as a local property a fundamental building block of our theoretical analysis of Dyna-style MBRL, which provides us with an effective mechanism for model usage. Combining this mechanism with an easy-to-compute and expressive estimate for model accuracy, we propose the Model-Based Actor-Critic with Uncertainty-Aware Rollout Adpation (MACURA) algorithm. Benchmarking on MuJoCo, we show that MACURA outperforms the current state-of-the-art substantially concerning data efficiency and asymptotic performance. Finally, the rollout mechanism of MACURA solely introduces one essential hyperparameter, making it considerably easier to tune than competitor approaches.

\section*{Acknowledgements}

ZF Friedrichshafen AG partially funded this research.
Furthermore, the research was in part supported by the German Federal
Ministry for Economic Affairs and Climate Action (BMWK) through the project EEMotion.
Computations were performed with computing resources granted by RWTH Aachen University under projects rwth1428, rwth1472, and rwth1552.

\section*{Impact Statement}
We present the MACURA algorithm that substantially improves data efficiency and asymptotic performance of Dyna-style MBRL, while considerably reducing tuning effort. Thus, it addresses two major issues of RL applications in the real world; data inefficiency and laborious hyperparameter tuning. Therefore, MACURA contributes to the general applicability of RL approaches.

 There are many potential societal consequences of our work, none which we feel must be specifically highlighted here.

\nocite{langley00}

\bibliography{macura}
\bibliographystyle{icml2024}

\newpage
\appendix
\onecolumn
\section{Pseudocode Algorithms}
\label{appendix:algos}

\begin{algorithm}
\caption{``Vanilla'' Dyna-style Deep Model-based Reinforcement Learning}\label{alg:standard_deep_dyna}
\begin{algorithmic}
\STATE Initialize: dynamics model $\tilde{p}_{\theta}$, RL policy $\pi$,  environment replay buffer $\mathcal{D}_{\mathrm{env}} \gets \emptyset$, model replay buffer $\mathcal{D}_{\mathrm{mod}} \gets \emptyset$ , rollout length schedule for $T_{\mathrm{max}}$, steps before retraining $R$, number of model-based rollouts $M$, RL update steps $G$.
\FOR {each iteration}
\STATE $s_0 \sim \rho_0(s)$
\FOR{each environment step}
\STATE $a_t \sim \pi ( \cdot \mid s_t)$
\STATE $s_{t+1} \sim p( \cdot \mid s_t, a_t)$
\STATE $r_{t+1} = r(s_t, a_t)$
\STATE $\mathcal{D}_{\mathrm{env}} \gets \mathcal{D}_{\mathrm{env}} \cup \{ (s_t, a_t, r_{t+1}, s_{t+1} )\}$
\IF{$\mod(\text{environment step}, R)=0$}
    \FOR{each epoch}
    \STATE Train $\tilde{p}_{\theta}$ on $\mathcal{D}_{\mathrm{env}}$
    \ENDFOR
    \STATE Evict old data from $\mathcal{D}_{\mathrm{mod}}$
    \FOR{$m \in M$ model rollouts}
    \STATE $s_0^m \sim \mathcal{U}(\mathcal{D}_{\mathrm{env}})$
    \FOR {$t =0, \dots, T_{\mathrm{max}}-1$}
    \STATE $e_t^m \sim \mathcal{U}(1, \dots, E)$
    \STATE $a_t^m \sim \pi ( \cdot \mid s_t^m)$
    \STATE $s_{t+1}^m  \sim \tilde{p}_{\theta_{e_t^m}}( \cdot \mid s_t^m, a_t^m), $
    \STATE $r_{t+1}^m = r(s_t^m, a_t^m)$
    \STATE $\mathcal{D}_{\mathrm{mod}} \gets \mathcal{D}_{\mathrm{mod}} \cup \{ (s_t^m, a_t^m, r_{t+1}^m, s_{t+1}^m )\}$
    \ENDFOR
    \ENDFOR
\ENDIF
\FOR{$G$ gradient steps}
\STATE Train $\pi$ on $\mathcal{D}_{\mathrm{mod}} \cup \mathcal{D}_{\mathrm{env}}$
\ENDFOR
\ENDFOR
\ENDFOR
\end{algorithmic}
\end{algorithm}
\clearpage

\begin{algorithm}
\caption{Model-based Actor-Critic with Uncertainty-aware Rollout Adaption (MACURA)}\label{alg:macura}
\begin{algorithmic}
\STATE Initialize: dynamics model $\tilde{p}_{\theta}$, RL policy $\pi$,  environment replay buffer $\mathcal{D}_{\mathrm{env}} \gets \emptyset$, model replay buffer $\mathcal{D}_{\mathrm{mod}} \gets \emptyset$ , steps before retraining $R$, number of model-based rollouts $M$, maximum RL update steps $G_{\mathrm{max}}$.
\STATE Initialize $\tilde{p}_{\theta}, \pi,  \mathcal{D}_{\mathrm{env}} \gets \emptyset$, $\mathcal{D}_{\mathrm{mod}} \gets \emptyset$, \textcolor{blue}{ fixed $T_{\mathrm{max}}$, $\zeta$ }
\FOR {each iteration}
\STATE $s_0 \sim \rho_0(s)$
\FOR {each environment step}
\STATE $a_t \sim \pi (\cdot \mid s_t)$ \textcolor{blue}{with correlated exploration noise \cite{eberhard2023pink}}
\STATE $s_{t+1} \sim p( \cdot \mid s_t, a_t)$
\STATE $r_{t+1} = r(s_t, a_t)$
\STATE $\mathcal{D}_{\mathrm{env}} \gets \mathcal{D}_{\mathrm{env}} \cup \{ (s_t, a_t, r_{t+1}, s_{t+1} )\}$
\IF{$\mod(\text{environment step}, R )=0$}
    \STATE \textcolor{blue}{$K \gets K+1$}
    \STATE \textcolor{blue}{$k \gets K$}
    \FOR{each epoch}
    \STATE Train $\tilde{p}_{\theta}$ on $\mathcal{D}_{\mathrm{env}}$
    \ENDFOR
    \STATE Evict old data from $\mathcal{D}_{\mathrm{mod}}$
    
    \FOR{$m \in M$ model rollouts}
    \STATE $s_0^m \sim \mathcal{U}(\mathcal{D}_{\mathrm{env}})$
    \STATE $e_0^m \sim \mathcal{U}(1, \dots, E)$
    \STATE $a_0^m \sim \pi (\cdot \mid s_0^m)$
    \STATE $s_{1}^m  \sim \tilde{p}_{\theta_{e_0^m}}(\cdot \mid s_0^m, a_0^m) $
    \STATE $r_{1}^m = r(s_0^m, a_0^m)$
    \STATE \textcolor{blue}{$u_\mathrm{GJS}(s_0^m, a_0^m)$ according to \eqref{eq:u_gjs}}
    \IF{\textcolor{blue}{$u_\mathrm{GJS}(s_0^m, a_0^m) < \kappa$}}
    \STATE $\mathcal{D}_{\mathrm{mod}} \gets \mathcal{D}_{\mathrm{mod}} \cup \{ (s_0^m, a_0^m, r_{1}^m, s_{1}^m )\}$
    \ELSE
    \STATE{\textcolor{blue}{stop rollout $m$ and discard data}}
    \ENDIF
    \ENDFOR
    \STATE \textcolor{blue}{$\hat{u}_{\mathrm{GJS}, k} = \inf \left\{ u_\mathrm{GJS}(s_0, a_0) \in \left\{ u_\mathrm{GJS}(s_0^1, a_0^1), \dots, u_\mathrm{GJS}(s_0^M, a_0^M) \right\} : \zeta \leq \mathrm{CDF}_k(u_\mathrm{GJS}(s_0, a_0))\right\} $}\footnotemark
    \STATE \textcolor{blue}{$\kappa \gets \frac{\xi}{K} \sum_{k=1}^K \hat{u}_{\mathrm{GJS}, k}$}
    \FOR {$t =1, \dots, T_{\mathrm{max}}-1$}
    \FOR{$m \in M$ model rollouts}
    \STATE $e_t^m \sim \mathcal{U}(1, \dots, E)$
    \STATE $a_t^m \sim \pi (\cdot \mid s_t^m)$
    \STATE $s_{t+1}^m  \sim \tilde{p}_{\theta_{e_t^m}}(\cdot \mid s_t^m, a_t^m), $
    \STATE $r_{t+1}^m = r(s_t^m, a_t^m)$
    \STATE \textcolor{blue}{$u_\mathrm{GJS}(s_t^m, a_t^m)$ according to \eqref{eq:u_gjs}}
    \IF{\textcolor{blue}{$u_\mathrm{GJS}(s_t^m, a_t^m) < \kappa$}}
    \STATE $\mathcal{D}_{\mathrm{mod}} \gets \mathcal{D}_{\mathrm{mod}} \cup \{ (s_t^m, a_t^m, r_{t+1}^m, s_{t+1}^m )\}$
    \ELSE
    \STATE{\textcolor{blue}{stop rollout $m$ and discard data}}
    \ENDIF
    \ENDFOR
    \ENDFOR
\ENDIF
\FOR{\textcolor{blue}{$ G = \Bigl\lfloor G_{\mathrm{max}} \frac{|\mathcal{D}_{\mathrm{mod}}|}{|\mathcal{D}_{\mathrm{mod}}|_{\mathrm{max}} }\Bigr\rceil $ gradient steps}}
\STATE Update $\pi$ on $\mathcal{D}_{\mathrm{mod}} \cup \mathcal{D}_{\mathrm{env}} $ 
\ENDFOR
\ENDFOR
\ENDFOR
\end{algorithmic}
\end{algorithm}
\clearpage
\footnotetext{$\mathrm{CDF}_k(u_\mathrm{GJS}(s_0, a_0))$ denotes the cumulative distribution function of GJS uncertainty estimates at time step $0$ at the $k$\textsuperscript{th} round of model-based rollouts.}

\section{Proofs}
\label{appendix_proofs}

\begin{lemma}[Return mismatch with respect to state distribution shift]
\label{lemma:return_bound}
Be the expected return following policy $\pi$ in $\hat{\mathcal{M}}$ 
$$
\mathbb{E}_{s \sim \hat{\mathcal{M}}, a \sim \pi}\left[\sum_{t=0}^{T(\omega)} \gamma^t r^{\hat{\mathcal{M}}}_{t+1}\right]:=\eta[\pi] 
$$
and the expected return following the same policy in $\widetilde{\mathcal{M}}$
$$
 \mathbb{E}_{s \sim \widetilde{\mathcal{M}}, a \sim \pi}\left[\sum_{t=0}^{T(\omega)} \gamma^t r^{\widetilde{\mathcal{M}}}_{t+1}\right]:=\tilde{\eta}[\pi],
$$
then

$$|\eta[\pi]-\tilde{\eta}[\pi]| \leq  2 r_{\mathrm{max}} \footnote{Dynamics mismatch is especially an issue in high-rewarding areas of the state-action space. In this work, however, we will neglect the dynamics-reward coupling and instead, only focus on the dynamics mismatch. Thus, we consider the conservative upper bound of $r_{\max }$.}\sum_{t=0}^{T(\omega)} \gamma^t D_{\mathrm{TV}} \left( p^t(s) \| \tilde{p}^t (s) \right) \footnote{We follow the common abuse of notation introduced in \cite{Janner2019Dec}, formulating the TV distance with respect to the probability densities rather than the stochastic process as would be formally correct.}.$$ 

\begin{proof}

\begin{equation*}
\begin{aligned}
& |\eta[\pi]-\tilde{\eta}[\pi]| \\
& =  \left| \mathbb{E}_{s \sim \hat{\mathcal{M}}, a \sim \pi } \left[ \sum_{t=0}^{T(\omega)} \gamma^t r^{\hat{\mathcal{M}}}_{t+1}\right] - \mathbb{E}_{s \sim \widetilde{\mathcal{M}}, a \sim \pi } \left[ \sum_{t=0}^{T(\omega)} \gamma^t r^{\widetilde{\mathcal{M}}}_{t+1}\right] \right| \\
&=  \left| \int_{\mathcal{E}} \int_{\mathcal{A}} \left( \sum_{t=0}^{T(\omega)} \gamma^t \left( p^t(s,a) - \tilde{p}^t (s,a) \right) \right) r(s,a)  \,da\ \,ds\ \right|\\
&= \left| \sum_{t=0}^{T(\omega)} \int_{\mathcal{E}} \int_{\mathcal{A}} \gamma^t \left( p^t(s,a) - \tilde{p}^t (s,a) \right) r(s,a)  \,da\ \,ds\ \right|\\
&\leq   \sum_{t=0}^{T(\omega)} \int_{\mathcal{E}} \int_{\mathcal{A}} \gamma^t \left| p^t(s,a) - \tilde{p}^t (s,a) \right| r(s,a)  \,da\ \,ds\ \\
&\leq  r_{\mathrm{max}} \sum_{t=0}^{T(\omega)} \int_{\mathcal{E}} \int_{\mathcal{A}} \gamma^t \left| p^t(s,a) - \tilde{p}^t (s,a) \right| \,da\ \,ds\ \\
&=  r_{\mathrm{max}} \sum_{t=0}^{T(\omega)} \int_{\mathcal{E}} \int_{\mathcal{A}} \gamma^t \left| \left( p^t(s) - \tilde{p}^t (s) \right) \pi(a \mid s) \right| \,da\ \,ds\ \\
&=  r_{\mathrm{max}} \sum_{t=0}^{T(\omega)} \gamma^t  \int_{\mathcal{E}} \left| p^t(s) - \tilde{p}^t (s) \right| \,ds\ \\
&=  2 r_{\mathrm{max}} \sum_{t=0}^{T(\omega)} \gamma^t D_{\mathrm{TV}} \left( p^t(s) \| \tilde{p}^t (s) \right) \\
\end{aligned}
\end{equation*}
\end{proof}
\end{lemma}
\clearpage

\begin{lemma}[Recursive Formulation]
\label{lemma:recursive}
Be $D_{T V}\left(p^t(\iota) \| \tilde{p}^t(\iota)\right):=\epsilon_t$ for an arbitrary $\iota$ and by construction $\epsilon_0=0$. Further be $\mathbb{E}_{s \sim \tilde{p}^{t-1}}\left[D_{T V}\left(p\left(s^{\prime} \mid s\right) \| \tilde{p}\left(s^{\prime} \mid s\right)\right)\right]:=\delta_t$ and $\delta_0 = \epsilon_0 = 0$.  Then we can bound $\epsilon_t$ by
$$\epsilon_t \le \sum_{\tau=0}^t \delta_\tau$$
\begin{proof}
\begin{equation*}
    \begin{aligned}
& \epsilon_t = D_{\mathrm{TV}}\left(p^t(s^{\prime}) \| \tilde{p}^t(s^{\prime})\right) \\
& =\frac{1}{2} \int_{\mathcal{E}}\left|p^t(s^{\prime})-\tilde{p}^t(s^{\prime})\right| \, ds^{\prime}\ \\
& =\frac{1}{2} \int_{\mathcal{E}}\left|\int_{\mathcal{E}}  p(s^{\prime} \mid s) p^{t-1}(s)- \tilde{p}(s^{\prime} \mid s) \tilde{p}^{t-1}(s) \, ds \ \right| \, ds^{\prime}\ \\
& \leq \frac{1}{2} \int_{\mathcal{E}}\int_{\mathcal{E}} \left|  p(s^{\prime} \mid s) p^{t-1}(s)- \tilde{p}(s^{\prime} \mid s) \tilde{p}^{t-1}(s) \right| \, ds \  \, ds^{\prime}\ \\
& = \frac{1}{2} \int_{\mathcal{E}}\int_{\mathcal{E}} \left|  p(s^{\prime} \mid s) p^{t-1}(s) - p(s^{\prime} \mid s) \tilde{p}^{t-1}(s) + p(s^{\prime} \mid s) \tilde{p}^{t-1}(s) - \tilde{p}(s^{\prime} \mid s) \tilde{p}^{t-1}(s) \right| \, ds \  \, ds^{\prime}\ \\
& \leq \frac{1}{2} \int_{\mathcal{E}} \int_{\mathcal{E}} \tilde{p}^{t-1}(s) \left| p(s^{\prime} \mid s) - \tilde{p}(s^{\prime} \mid s)  \right| \, ds \  \, ds^{\prime}\ +
\frac{1}{2} \int_{\mathcal{S}}\int_{\mathcal{E}} p(s^{\prime} \mid s) \left|  p^{t-1}(s) -\tilde{p}^{t-1}(s) \right| \, ds \  \, ds^{\prime}\ \\
& = \mathbb{E}_{s \sim \tilde{p}^{t-1}} \left[ \frac{1}{2} \int_{\mathcal{E}} \left| p(s^{\prime} \mid s) - \tilde{p}(s^{\prime} \mid s) \right| \, ds^{\prime}\ \right] + \frac{1}{2} \int_{\mathcal{E}} \left| p^{t-1}(s) - \tilde{p}^{t-1}(s) \right| \, ds \ \\
& = \mathbb{E}_{s \sim \tilde{p}^{t-1}} \left[ D_{\mathrm{TV}} \left( p(s^{\prime} \mid s) \| \tilde{p}(s^{\prime} \mid s) \right) \right] + D_{\mathrm{TV}} \left( p^{t-1}(s) \| \tilde{p}^{t-1}(s) \right) \\
& =\delta_t+\epsilon_{t-1}=\delta_t+\delta_{t-1}+\epsilon_{t-2}=\ldots= \\
& =\epsilon_0+\sum_{\tau=1}^t \delta_\tau =\delta_0+\sum_{\tau=1}^t \delta_\tau=\sum_{\tau=0}^t \delta_\tau \\
&
\end{aligned}
\end{equation*}
\end{proof}

\end{lemma}

\begin{lemma}[Dependency on dynamics mismatch]

\label{lemma:dyn_form}
    Be 
    $$
    \Delta p_{\mathcal{E}} [\pi] := \sup_{s \in \mathcal{E}, a \sim \pi} \{ D_{T V} \left( p\left(s^{\prime} \mid s, a \right) \| \tilde{p}\left(s^{\prime} \mid s, a\right)\right)  \}
    $$
    then 
    $$  \mathbb{E}_{s \sim \tilde{p}^{t-1}}\left[D_{T V} \left( p\left(s^{\prime} \mid s\right) \| \tilde{p}\left(s^{\prime} \mid s\right)\right)\right] \leq \Delta p_{\mathcal{E}} [\pi] $$
\begin{proof}
\begin{equation*}
    \begin{aligned}
    & \mathbb{E}_{s \sim \tilde{p}^{t-1}}\left[D_{T V} \left( p\left(s^{\prime} \mid s \right) \| \tilde{p}\left(s^{\prime} \mid s\right)\right)\right] \\
    &=  \frac{1}{2} \int_{\mathcal{E}} \int_{\mathcal{E}} \tilde{p}^{t-1}\left(s\right)\left|p\left(s^{\prime} \mid s\right)-\tilde{p}\left(s^{\prime} \mid s \right)\right| \, ds\ \, ds^{\prime}\ \\
    &=  \frac{1}{2}  \int_{\mathcal{E}} \int_{\mathcal{E}} \tilde{p}^{t-1}\left(s\right)\left| \int_{\mathcal{A}} \left(p\left(s^{\prime} \mid s, a\right)-\tilde{p}\left(s^{\prime} \mid s, a\right)\right) \pi(a \mid s) \, da\  \right| \, ds\ \, ds^{\prime}\ \\
    & \leq  \frac{1}{2} \int_{\mathcal{E}} \int_{\mathcal{E}} \int_{\mathcal{A}} \tilde{p}^{t-1}\left(s \right) \pi\left(a \mid s \right)\left|p\left(s^{\prime} \mid s, a\right)-\tilde{p}\left(s^{\prime} \mid s, a\right)\right| \, da\ \, ds\ \, ds^{\prime}\ \\
    &  = \mathbb{E}_{s \sim \tilde{p}^{t-1}, a \sim \pi}\left[\frac{1}{2} \int_{\mathcal{E}} \left|p\left(s^{\prime} \mid s, a \right)-\tilde{p} \left(s^{\prime} \mid s, a\right)\right| \, ds^{\prime}\ \right] \\
    & = \mathbb{E}_{s \sim  \tilde{p}^{t-1}, a \sim \pi}\left[D_{\mathrm{TV}}\left(p\left(s^{\prime} \mid s, a\right) \| p\left(s^{\prime} \mid s, a \right)\right)\right]\\
    & \leq \Delta p_{\mathcal{E}} [\pi]
    \end{aligned}
\end{equation*}
\end{proof}

\end{lemma}

\clearpage
\begin{theorem}[Monotonic Improvement under Dynamics Misalignment on $\mathcal{E} \subseteq \mathcal{S}$ ]
\label{theo: return_bound_appendix}

We define two MDPs $\hat{\mathcal{M}}$ and $\widetilde{\mathcal{M}}$ with a common state space $\mathcal{S}$, action space $\mathcal{A}$ and reward function $r: \mathcal{S} \times \mathcal{A} \rightarrow \mathbb{R}^{+}$~\footnote{We assume rewards to be strictly positive. An equivalent reward function can be trivially constructed from any bounded reward function $r(s, a) \in [r_{\mathrm{min}}, r_{\mathrm{max}}] \forall s \in \mathcal{S}, a \in \mathcal{A}$}. $\hat{\mathcal{M}}$ has dynamics $p(s^{\prime} \mid s, a): \mathcal{S} \times \mathcal{A} \rightarrow \mathcal{S}$, while $\widetilde{\mathcal{M}}$ has dynamics $\tilde{p}(s^{\prime} \mid s, a): \mathcal{S} \times \mathcal{A} \rightarrow \mathcal{S}$.
For both MDPs, we define probability densities
$$
\mathbb{P}[s^{\hat{\mathcal{M}}}_{t} \in \mathcal{B}] = \int_{\mathcal{B}} p^t(s) \, ds\
$$
$$
\mathbb{P}[s^{\hat{\mathcal{M}}}_{t} \in \mathcal{B}, a^{\hat{\mathcal{M}}}_{t} \in \mathcal{C} ] = \int_{\mathcal{B}} \int_{\mathcal{C}} p^t(s, a) \, da\ \, ds\ 
$$
as well as a dynamics function
$$
p(s^{\prime} \mid s ) = \int_{\mathcal{A}} p(s^{\prime} \mid s , a) \pi(a \mid s) \, da\
$$

and equivalently,
$$
\mathbb{P}[s^{\widetilde{\mathcal{M}}}_{t} \in \mathcal{B}] = \int_{\mathcal{B}} \tilde{p}^t(s) \, ds\
$$
$$
\mathbb{P}[s^{\mathcal{\tilde{M}}}_{t} \in \mathcal{B}, a^{\mathcal{\tilde{M}}}_{t} \in \mathcal{C} ] = \int_{\mathcal{B}} \int_{\mathcal{C}} \tilde{p}^t(s, a) \, da\ \, ds\ 
$$
$$
\tilde{p}(s^{\prime} \mid s ) = \int_{\mathcal{A}} \tilde{p}(s^{\prime} \mid s , a) \pi(a \mid s) \, da\
$$
for all Borel-measuarble sets $\mathcal{B} \subseteq \mathcal{S}, \mathcal{C} \subseteq \mathcal{A}$ and conditional probability densities $\pi(a \mid s): \mathcal{S} \rightarrow \mathcal{A}$.

Further, we define a coupling between $\hat{\mathcal{M}}$ and $\widetilde{\mathcal{M}}$ via a random stopping time
$$
T^{\hat{\mathcal{M}}}(\omega) := \min \{ t \in \mathbb{N} \mid s^{\hat{\mathcal{M}}}_{t} (\omega) \in \mathcal{E}^C  \}, 
T^\mathcal{\tilde{M}}(\omega) := \min \{ t \in \mathbb{N} \mid 
s^{\mathcal{\tilde M}}_{t} (\omega) \in \mathcal{E}^C\}, T(\omega) := \min\{T^{\hat{\mathcal{M}}}, T^\mathcal{\tilde M}\} - 1,
$$
where $s_t$ is a trajectory with respect to the MDPs $\hat{\mathcal{M}}$ or $\widetilde{\mathcal{M}}$ respectively and regarded as a random variable,  $\mathcal{E} \subseteq \mathcal{S}$, and $\mathcal{E}^C$ the compliment of $\mathcal{E}$.

as well as identical start states $s^{\widetilde{\mathcal{M}}}_{0} = s^{\hat{\mathcal{M}}}_{0} \in \mathcal{E}$.

Suppose the expected return following policy $\pi$ in $\hat{\mathcal{M}}$ is denoted by  
$$
\mathbb{E}_{s \sim \hat{\mathcal{M}}, a \sim \pi}\left[\sum_{t=0}^{T(\omega)} \gamma^t r^{\hat{\mathcal{M}}}_{t+1}\right]:=\eta[\pi] 
$$
and $\tilde{\eta}[\pi]$ describes the expected return following the same policy in $\widetilde{\mathcal{M}}$
$$
 \mathbb{E}_{s \sim \widetilde{\mathcal{M}}, a \sim \pi}\left[\sum_{t=0}^{T(\omega)} \gamma^t r^{\widetilde{\mathcal{M}}}_{t+1}\right]:=\tilde{\eta}[\pi],
$$
then we can define a lower bound for $\eta[\pi]$ of the form 
$$
\eta[\pi] \ge \tilde{\eta}[\pi]-2 r_{\max } \sum_{t=0}^{T(\omega)} \gamma^t \sum_{\tau=0}^t \Delta p_{\mathcal{E}} [\pi].
$$
\begin{proof}

$$
\eta[\pi] \ge \tilde{\eta}[\pi]-|\eta[\pi]-\tilde{\eta}[\pi]|
$$
Using Lemma \ref{lemma:return_bound}
$$
\eta[\pi] \ge \tilde{\eta}[\pi]-2 r_{\max } \sum_{t=0}^{T(\omega)} \gamma^t D_{T V}\left(p^t(s) \| \tilde{p}^t(s)\right)
$$
Using Lemma \ref{lemma:recursive}
$$
\eta[\pi] \ge \tilde{\eta}[\pi]-2 r_{\max } \sum_{t=0}^{T(\omega)} \gamma^t \sum_{\tau=0}^t \mathbb{E}_{s \sim \tilde{p}^{ \tau-1}}\left[D_{ T V } \left(p\left(s^{\prime} \mid s \right) \| \tilde{p}\left(s^{\prime} \mid s \right)\right)\right]
$$
Using Lemma \ref{lemma:dyn_form}
$$
\eta[\pi] \ge \tilde{\eta}[\pi]-2 r_{\max } \sum_{t=0}^{T(\omega)} \gamma^t \sum_{\tau=0}^t \Delta p_{\mathcal{E}} [\pi]
$$
\end{proof}

\end{theorem}

\clearpage

\section{Toy Example}
\label{appendix:toy_example}
In order to test the proposed uncertainty measure $u_{\mathrm{GJS}}$ \eqref{eq:u_gjs}, we use a two-dimensional toy example of a pendulum.
We discuss the dynamics of the pendulum in Section \ref{sec:toy_example_pendulum}, the feedback controller used as policy in Section \ref{sec:toy_example_controller}, and the experimental setup leading to Figure \ref{fig:uncertainty_measures} in Section \ref{sec:toy_example_data_generation}. We further compare our uncertainty estimate to the one proposed in \cite{Pan2020Dec} and show a substantially more reliable behavior of the  $u_{\mathrm{GJS}}$ measure in Section \ref{sec:toy_example_m2ac_comparison}.

\subsection{Pendulum Dynamics}
\label{sec:toy_example_pendulum}
\begin{figure}[tb]
    \centering
    \includegraphics[width=0.5\linewidth]{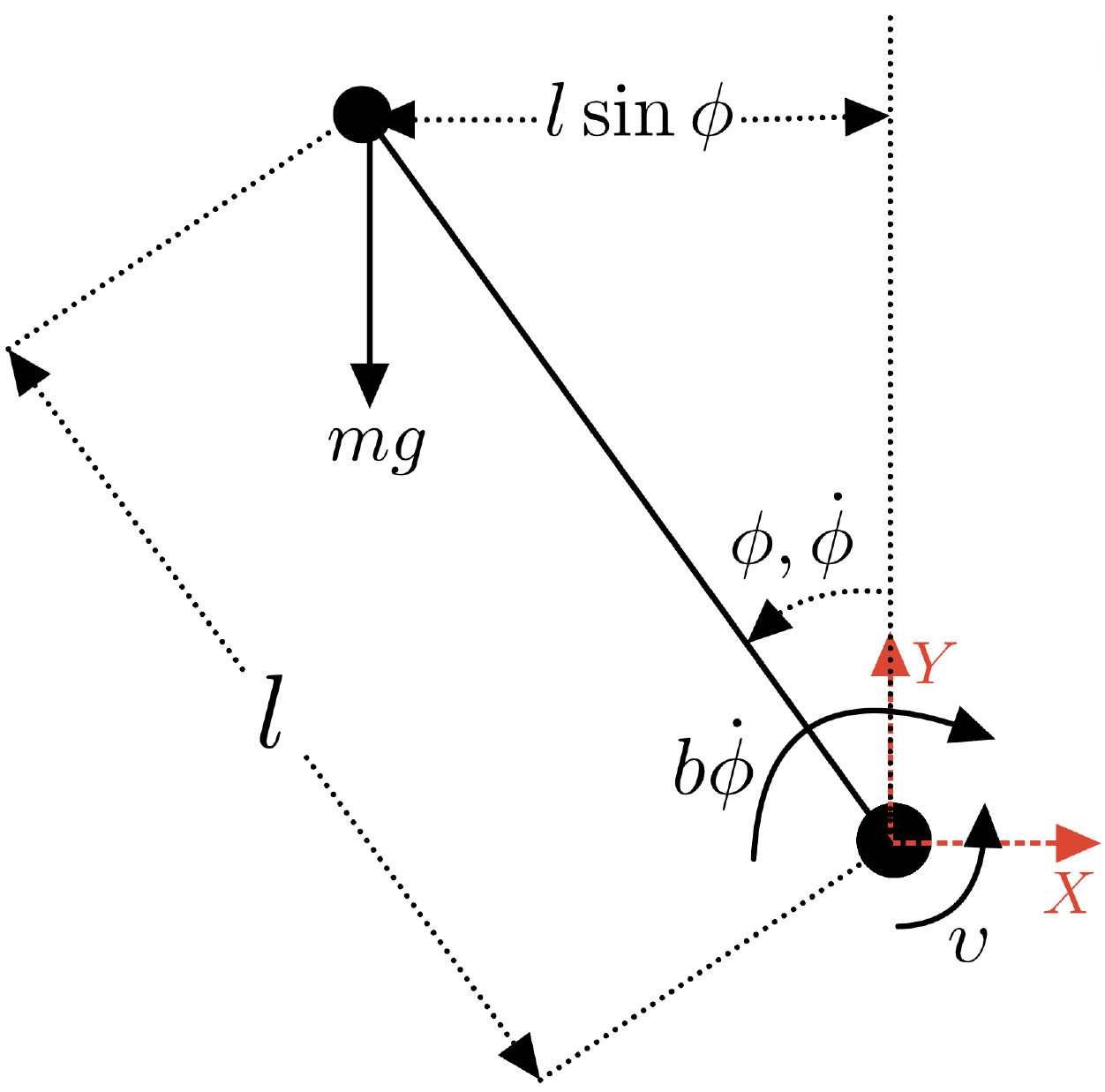}
    \caption{Free body diagram of the pendulum.}
    \label{fig:simple_pendulum}
\end{figure}

Figure \ref{fig:simple_pendulum} shows a free-body diagram of the pendulum near the upper fixed point. 
We can write the equation of motion for such a pendulum as 
\begin{equation}
    m l^2 \ddot \phi - mgl \sin \phi + b \phi = \upsilon,
    \label{pend_dynamics}
\end{equation}
where $m$ is the mass of the pendulum, $l$ is its length, $g$ is the acceleration due to gravity, and $b$ is the coefficient of viscous friction. Further, $\upsilon$ is the torque applied at the base of the pendulum that is used to control the pendulum. We define the continuous-time state as $x=\begin{bmatrix}
    \phi & \dot \phi
\end{bmatrix}^T$ with $\phi \in [ -3 , 3 ][\mathrm{rad}]$ the pendulum angle and $\dot{\phi} \in [-16, 16][\frac{\mathrm{rad}}{\mathrm{s}}]$ the pendulum's angular velocity. The nonlinear state-space equation for this system in continuous time are
\begin{equation}
    \dot x = \begin{bmatrix}
        \dot \phi \\
        \frac{1}{ml^2}\left(\upsilon + mgl \sin \phi - b \dot \phi \right)
    \end{bmatrix} = f(x,\upsilon).
    \label{pend_state_dynamics}
\end{equation}

When considering the system in discrete time, we sample observations and apply actions $a = \upsilon$ at discrete time points separated by a fixed time interval $\Delta t$. We simulate the pendulum by integrating the state equation \eqref{pend_state_dynamics} between the sampling time steps, keeping the applied action fixed throughout the integration. We additionally consider homoscedastic Gaussian process noise with covariance matrix $\Sigma$. Hence we obtain the discrete-time MDP dynamics as
\begin{equation}
p(\cdot|s,a) = \mathcal N\left(\mu(s,a),\Sigma \right),
\label{eq:toy_example_transition_kernel}
\end{equation}
with,
\begin{equation}
    \mu(s,a) = \int_{0}^{\Delta t}f(x(t),a)dt,
\end{equation}
where $x(0)=s$. The values for the different parameters can be seen in Table \ref{pendulum_params}.

\begin{table}[h!]
    \centering
    \caption[Pendulum parameters.]{Parameters of the Pendulum.}
    \begin{tabular}{ |c|c| } 
         \hline
         \textbf{Parameter} &\textbf{Value} \\ 
         \hline
         mass $m$ & 0.1\\ 
         \hline
         length $l$ & 1\\ 
         \hline
          acceleration due to gravity $g$ & 9.81\\ 
         \hline
          coefficient of viscous friction $b$ & 0.1\\ 
         \hline
          sampling time interval $\Delta t$ & 0.01 \\
         \hline
          &  \\
         process noise covariance matrix $\Sigma$ & $\begin{bmatrix}
             10^{-6} & 0 \\
             0 & 10^{-3}
         \end{bmatrix}$\\ 
           &  \\
         \hline
    \end{tabular}

\label{pendulum_params}
\end{table}

\subsection{Controller}
\label{sec:toy_example_controller}
The pendulum is controlled via the applied torque $\upsilon$.
We can use feedback linearization \cite{Adamy2022} to obtain a controller of the form
\begin{equation}
    \pi_{\mathrm{FL}}(s) =  \upsilon_{\mathrm{FL}}(x) = ml^2\left(\ddot \phi^d + K_{D} (\dot \phi^d - \dot \phi) + K_{P} (\phi^d - \phi)\right) - mgl \sin \phi + b \dot \phi,
    \label{feedback_lin}
\end{equation}
where $\phi^d$, $\dot \phi^d$, and $\ddot \phi^d$ denote the desired pendulum angle, angular velocity, and angular acceleration, respectively. The positive constants $K_P$ and $K_D$ denote the proportional and derivative gains respectively. We choose them to give under-damped feedback dynamics. The values can be seen in Table \ref{pendulum_controller}. For our purpose, we use the upper fixed point as the desired position, that is, $\phi^d = 0$, $\dot \phi^d =0$ and $\ddot \phi^d =0$.
\begin{table}[h!]
    \centering
    \caption[Pendulum controller parameters.]{Parameters of the Controller.}
    \begin{tabular}{ |c|c| } 
         \hline
         \textbf{Parameter} &\textbf{Value} \\ 
         \hline
         proportional gain $K_P$ & 25\\ 
         \hline
         derivative gain $K_D$ & 1\\ 
         \hline
    \end{tabular}
\label{pendulum_controller}
\end{table}

\subsection{Data Generation}
\label{sec:toy_example_data_generation}

We create interaction data that is stored $\mathcal{D}_{\mathrm{env}}$ to train a PE model. For this, we let the feedback linearization controller \eqref{feedback_lin} interact with the pendulum \eqref{pend_state_dynamics}. Each trajectory starts from a fixed start state $\begin{bmatrix}
    \phi_0 & \dot \phi_0
\end{bmatrix}^T$. The trajectories are terminated upon reaching $T_{\mathrm{max}}$ steps. Ten such trajectories are generated. This results in a data distribution with a characteristic spiral shape as depicted in Figure \ref{fig:unc_data} and \ref{fig:unc_data_appendix}. The parameters used for generating the trajectories can be found in Table \ref{pendulum_traj_params}.
\begin{table}[h!]
    \centering
    \caption[Simple pendulum trajectory parameters.]{Trajectory parameters for data generation.}
    \begin{tabular}{ |c|c| } 
         \hline
         \textbf{Parameter} &\textbf{Value} \\ 
         \hline
         initial angle $\phi_0$ & 3\\ 
         \hline
         initial angular velocity $\dot \phi_0$ & 0\\ 
         \hline
         number of steps $T_{\mathrm{max}}$ & 170\\ 
         \hline
    \end{tabular}
\label{pendulum_traj_params}
\end{table}

\subsection{Illustrating the $u_{\mathrm{GJS}}$ Uncertainty Measure}
\label{sec:toy_example_ugjs_test}
To illustrate the effectiveness of the proposed uncertainty measure $u_{\mathrm{GJS}}$, we investigate the connection between dynamics misalignment $D_{ \mathrm{TV} } \left(p\left(s^{\prime} \mid s, a\right) \| \tilde{p}\left(s^{\prime} \mid s, a\right)\right)$ and model uncertainty on the toy example.

We train a PE dynamics model on data created according to Section \ref{sec:toy_example_data_generation} and evaluate both quantities over $\mathcal{S}$.
To do this, we discretize the state space into a uniform grid $\{s_{ij}\}$ where $i$ is used to index over $\phi$ and $j$ is used to index over $\dot \phi$. For each $s_{ij}$ the corresponding action is obtained as
\begin{equation}
    a_{ij} = \pi_{\mathrm{FL}}(s_{ij}).
\end{equation}
First, we evaluate dynamics misalignment $D_{ \mathrm{TV} } \left(p\left(s^{\prime} \mid s, a\right) \| \tilde{p}\left(s^{\prime} \mid s, a\right)\right)$ over the predefined grid.
We obtain the true and predicted next state distribution as
\begin{equation}
p_{ij} = p(\cdot | s_{ij}, a_{ij}),
\end{equation}
querying the dynamics of the toy example \eqref{eq:toy_example_transition_kernel}, and
\begin{equation}
\tilde{p}_{ij, e} = \tilde{p}_{\theta_e}(\cdot|s_{ij}, a_{ij})
\end{equation}
for each PNN prediction with $e \in (1 \dots E)$.

As there is no closed-form solution for computing $D_{ \mathrm{TV} } \left(p_{ij} \| \tilde{p}_{ij}\right)$ we use the common upper bound of the total variation distance with respect to the Hellinger distance $D_{ \mathrm{H} }$:
\begin{equation}
    D_{ \mathrm{TV} } \left(p_{ij} \| \tilde{p}_{ij}\right) \leq \sqrt{2} D_{ \mathrm{H} } \left(p_{ij} \| \tilde{p}_{ij}\right).
    \label{eq:upper_bound_TV}
\end{equation}

\begin{figure}
     \centering
     \begin{subfigure}[b]{0.245\columnwidth}
         \centering
         \includegraphics[width=\textwidth]{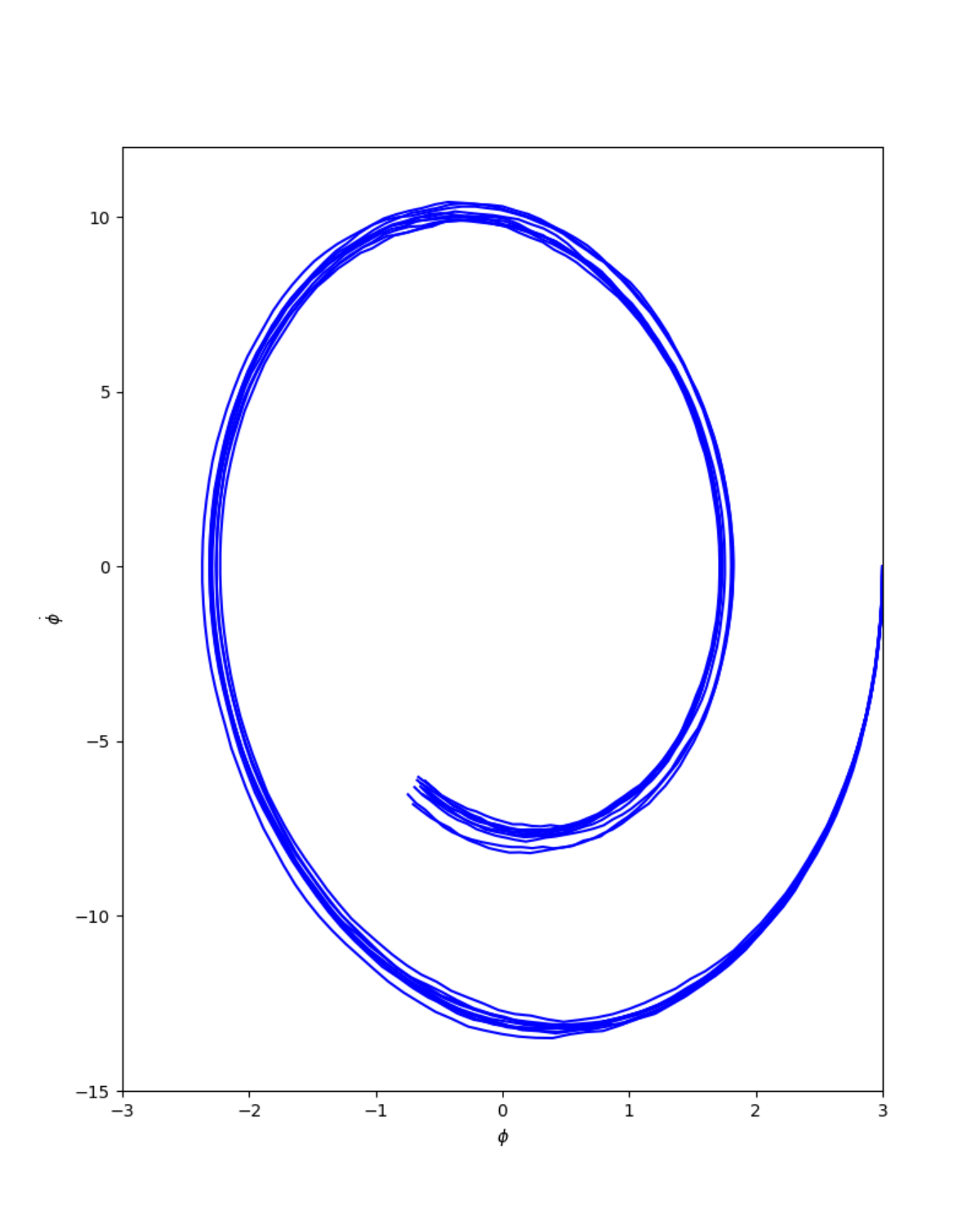}
         \vspace{-2\baselineskip}
         \caption{$\mathcal{D}_{\mathrm{env}}$}
         \label{fig:unc_data_appendix}
     \end{subfigure}
     \hfill
     \begin{subfigure}[b]{0.245\columnwidth}
         \centering
         \includegraphics[width=\textwidth]{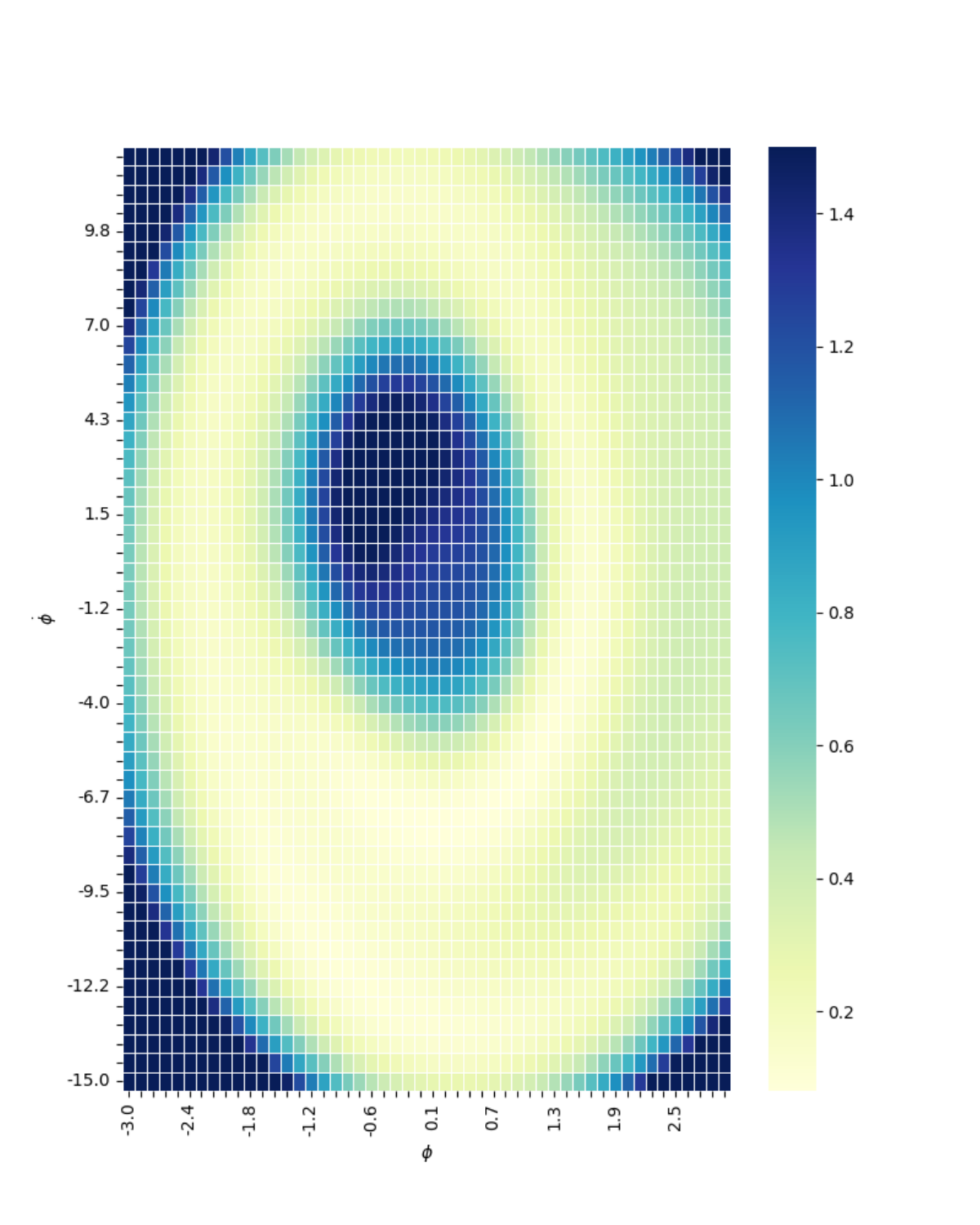}
         \vspace{-2\baselineskip}
         \caption{$u_{\mathrm{GJS}}$ (Ours)}
         \label{fig:unc_ours_appendix}
     \end{subfigure}
     \hfill
     \begin{subfigure}[b]{0.245\columnwidth}
         \centering
         \includegraphics[width=\textwidth]{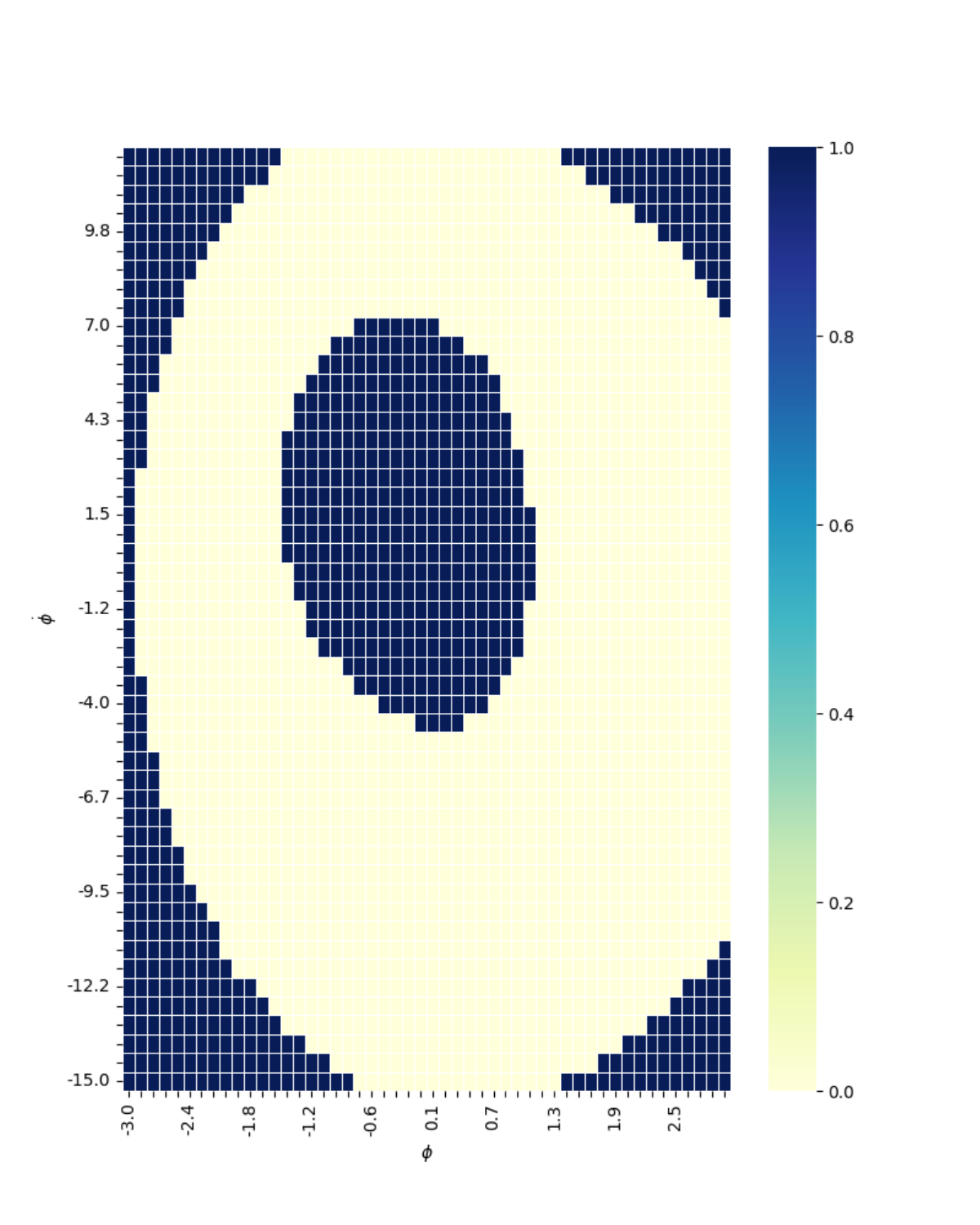}
         \vspace{-2\baselineskip}
         \caption{$\mathcal{E}$ with $u_{\mathrm{GJS}} < \kappa$}
         \label{fig:E_ours_appendix}
     \end{subfigure}
     \hfill
     \begin{subfigure}[b]{0.245\columnwidth}
         \centering
         \includegraphics[width=\textwidth]{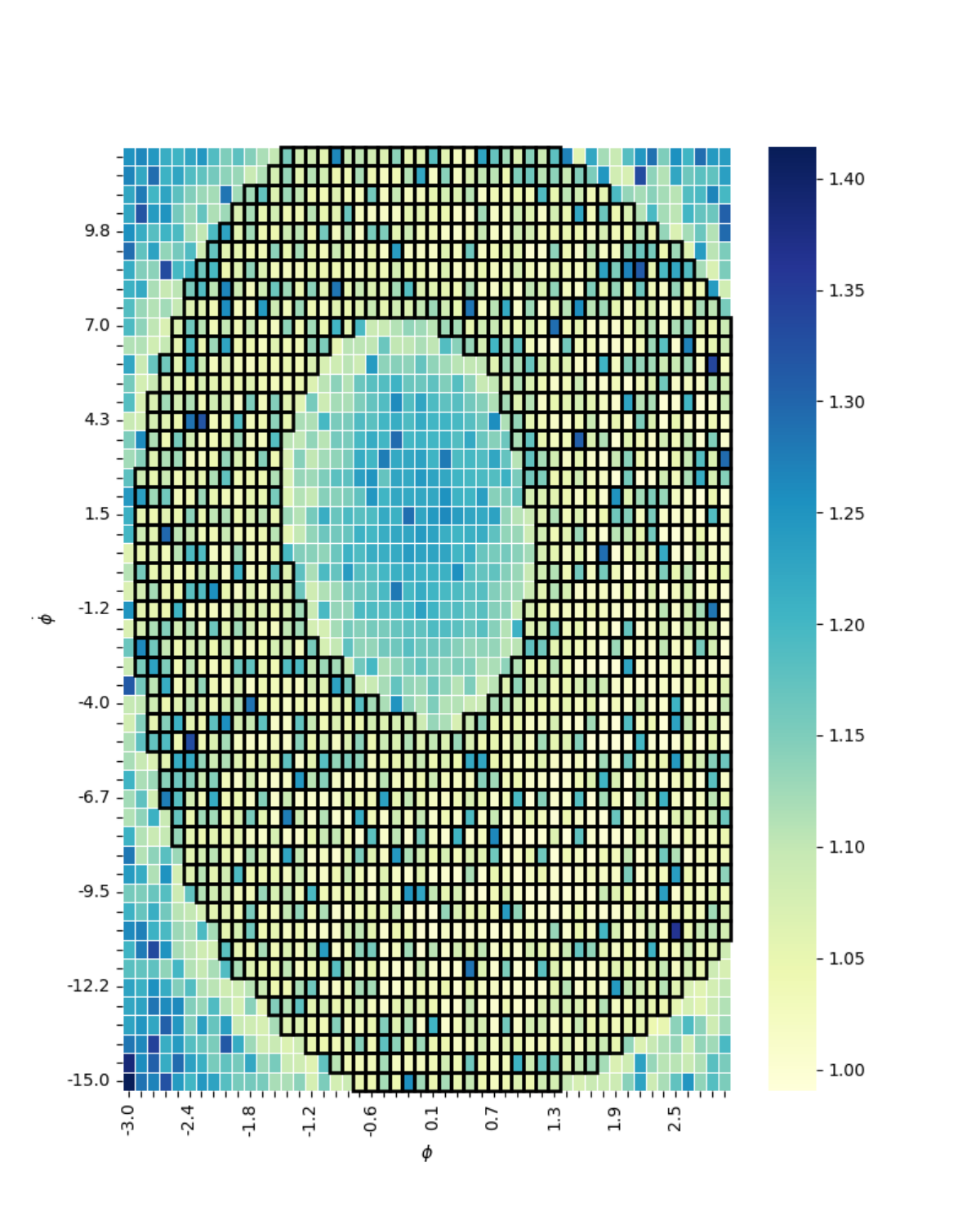}
         \vspace{-2\baselineskip}
         \caption{$\mathcal{E}$ with $u_{\mathrm{GJS}} $ over $D_{\mathrm{TV}}(p \| \tilde{p})$}
         \label{fig:E_TV_ours_appendix}
     \end{subfigure}
     \hfill
     \begin{subfigure}[b]{0.245\columnwidth}
         \centering
         \includegraphics[width=\textwidth]{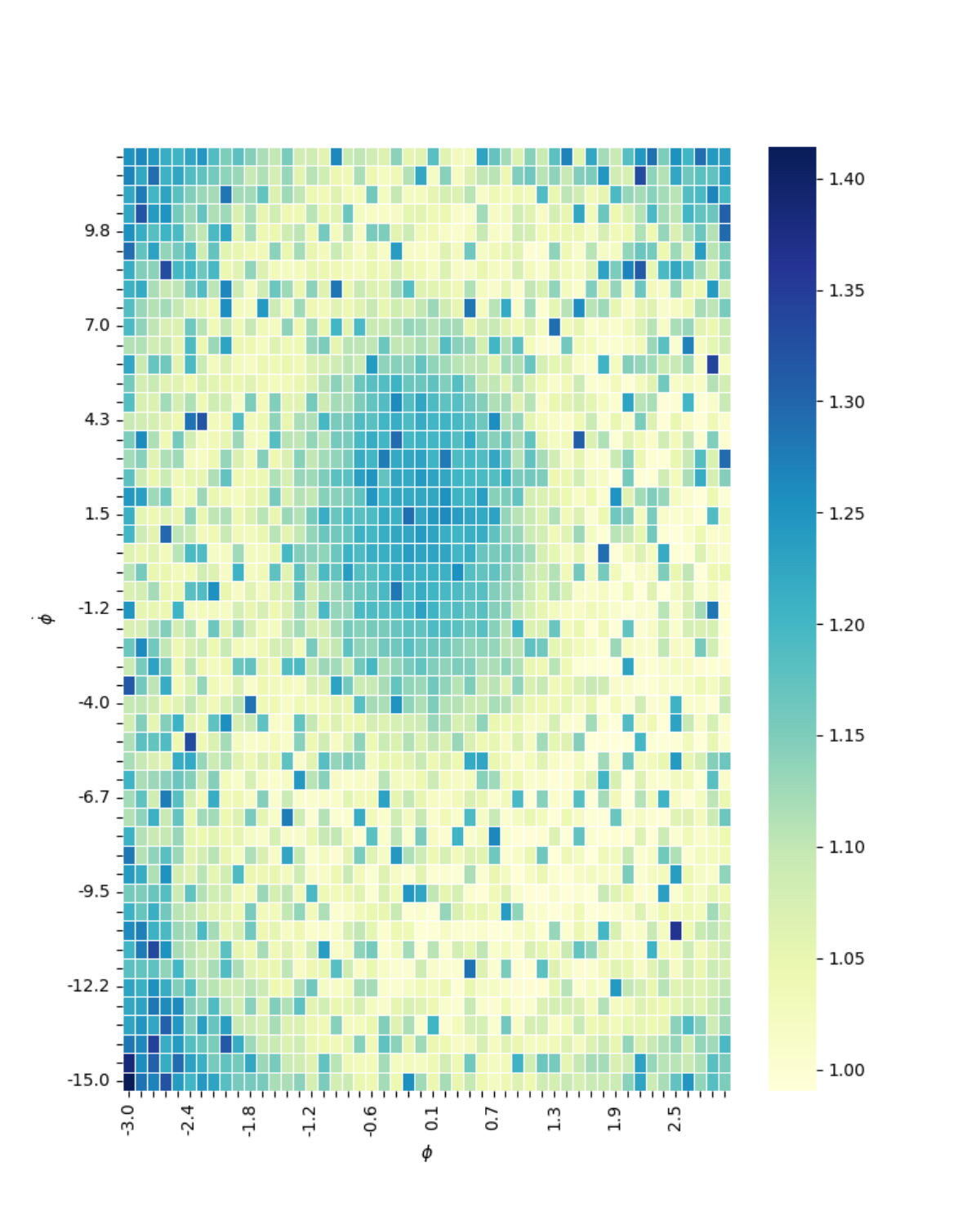}
         \vspace{-2\baselineskip}
         \caption{$D_{\mathrm{TV}}(p(s^{\prime}|s, a) \| \tilde{p}(s^{\prime}|s, a))$}
         \label{fig:unc_tv_appendix}
     \end{subfigure}
     \hfill
     \begin{subfigure}[b]{0.245\columnwidth}
         \centering
         \includegraphics[width=\textwidth]{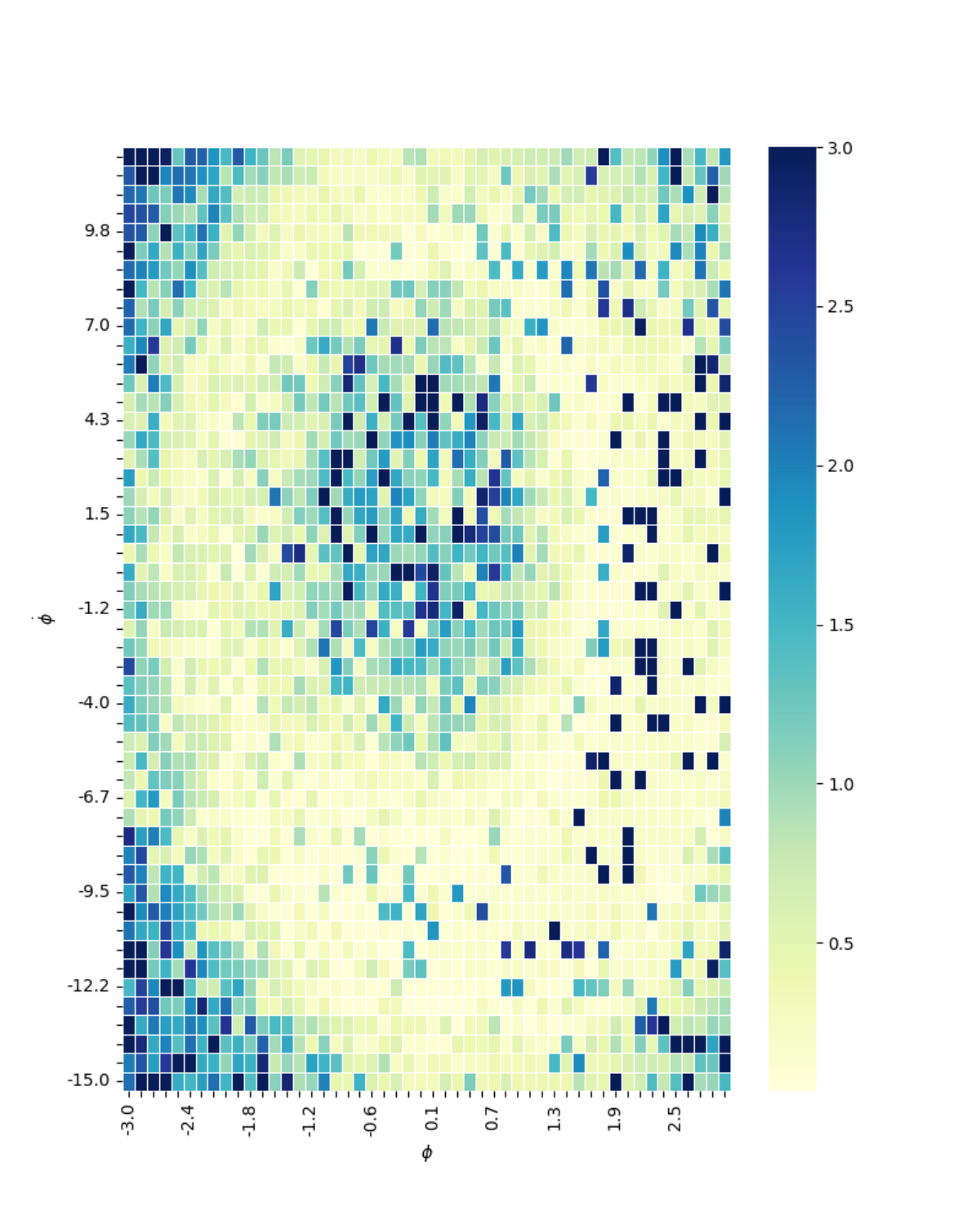}
         \vspace{-2\baselineskip}
         \caption{$u_{\mathrm{OvR}}$ (\cite{Pan2020Dec})}
         \label{fig:unc_m2ac_appendix}
     \end{subfigure}
     \hfill
     \begin{subfigure}[b]{0.245\columnwidth}
         \centering
         \includegraphics[width=\textwidth]{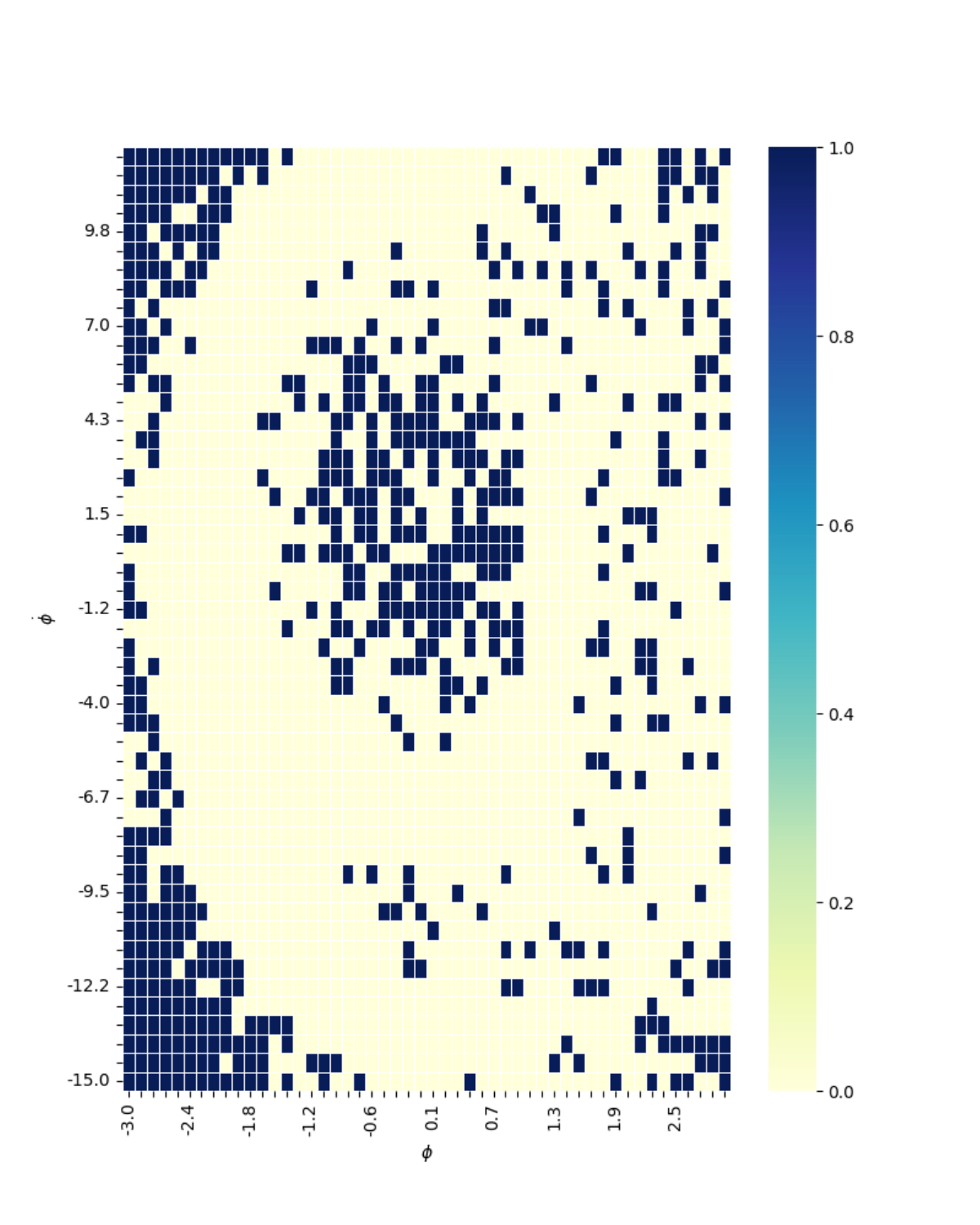}
         \vspace{-2\baselineskip}
         \caption{$\mathcal{E}$ with $u_{\mathrm{OvR}} < \kappa$}
         \label{fig:E_m2ac_appendix}
     \end{subfigure}
     \hfill
    \begin{subfigure}[b]{0.245\columnwidth}
         \centering
         \includegraphics[width=\textwidth]{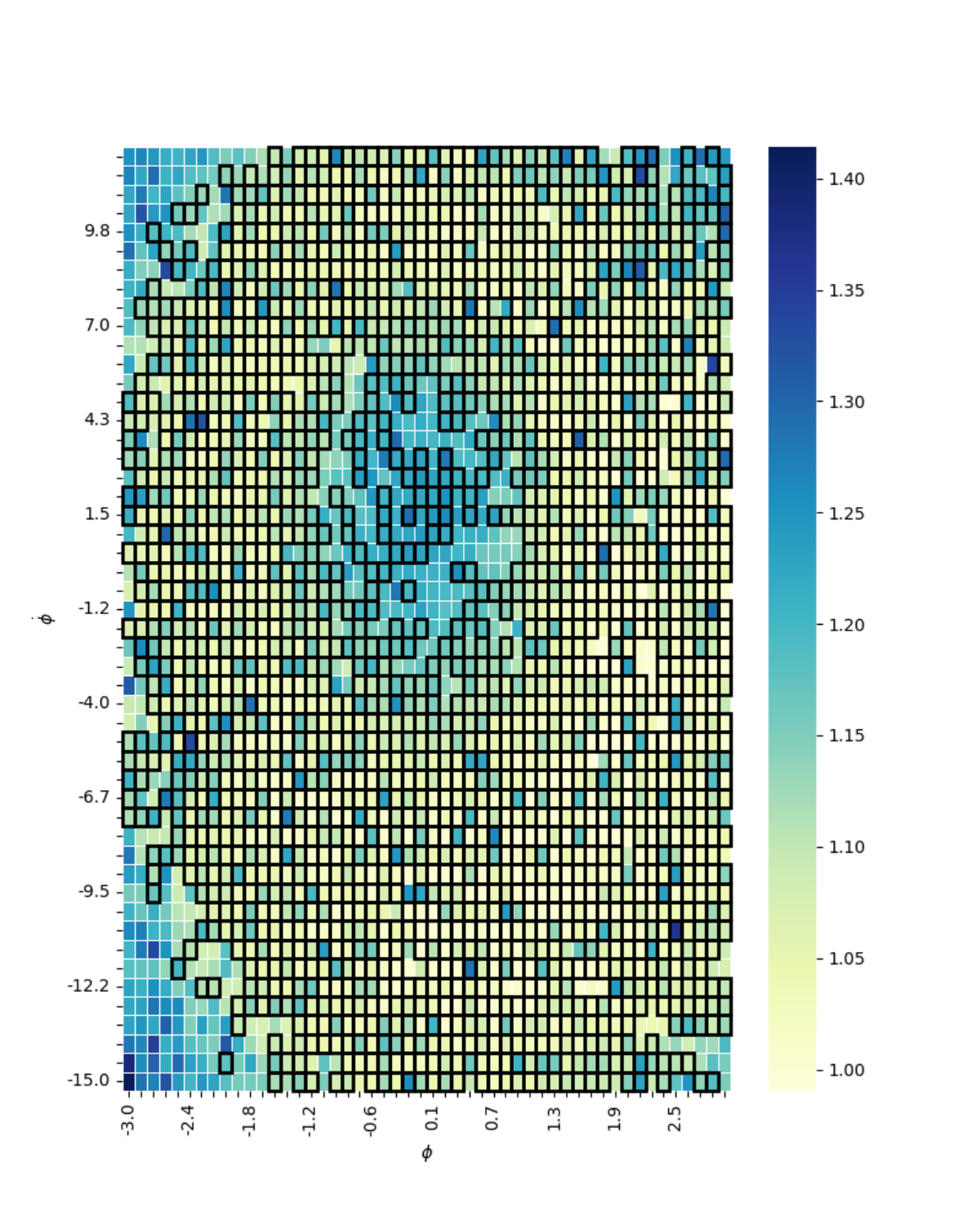}
         \vspace{-2\baselineskip}
         \caption{$\mathcal{E}$ with $u_{\mathrm{OvR}}$ over $D_{\mathrm{TV}}(p \| \tilde{p})$}
         \label{fig:E_TV_m2ac_appendix}
     \end{subfigure}
     
        \caption{Effectiveness of the GJS uncertainty measure}
        \label{fig:uncertainty_measures_appendix}
\end{figure}

Figures \ref{fig:unc_tv} and \ref{fig:unc_tv_appendix} show a heatmap over $\mathcal{S}$ of the discretized dynamics misalignment measurements 
\begin{equation}
    d_{ij} = \frac{1}{E} \sum_{e=1}^{E} \sqrt{2} D_{\mathrm{H}}(p_{ij} || \tilde{p}_{ij, e}).
\end{equation}

Similarly, we evaluate model uncertainty in Figures \ref{fig:unc_ours} and \ref{fig:unc_ours_appendix} over $\mathcal{S}$ plotting
\begin{equation}
u_{\mathrm{GJS}, ij} = u_{\mathrm{GJS}}(s_{ij},a_{ij}).
\end{equation}
As discussed in Section \ref{sec:u_gjs_illustration}, both align well. Choosing a suitable $\kappa$ allows to construct a meaningful set $\mathcal{E}$ as depicted in Figures \ref{fig:E_ours} and \ref{fig:E_ours_appendix} that aligns with areas of low dynamics misalignment as illustrated in Figure \ref{fig:E_TV_ours_appendix}.

\subsection{Illustrating the $u_{\mathrm{OvR}}$ Uncertainty Measure of the M2AC Algorithm \cite{Pan2020Dec}}
\label{sec:toy_example_m2ac_comparison}
Similar to MACURA, \citet{Pan2020Dec} propose to adapt the length of branched model-based rollouts in Dyna-style MBRL using model uncertainty.
Therefore, they present the One-versus-Rest $u_{\mathrm{OvR}}$ uncertainty estimate for PE models:
\begin{equation}
        u_{\mathrm{OvR}}(s,a) = D_{\mathrm{KL}} \left( \mathcal{N}(\mu_{\theta_e}(s,a), \Sigma_{\theta_e}(s,a)) \| \mathcal{N}(\mu_{\theta_{-e}}(s,a), \Sigma_{\theta_{-e}}(s,a)) \right).
\end{equation}
This uncertainty estimate is defined as the Kullback-Leibler divergence between a randomly chosen PNN prediction $e \sim \mathcal{U}(1, \dots, E)$ and a Gaussian distribution defined by merging the remaining PNNs of the PE model, such that
\begin{equation}
        \mu_{\theta_{-e}}(s,a) = \frac{1}{E-1}\sum_{f \neq e}^E \mu_{\theta_f}(s,a)
\end{equation}
and
\begin{equation}
        \Sigma_{\theta_{-e}}(s,a) = \frac{1}{E-1}\sum_{f \neq e}^E\left( \Sigma_{\theta_f}(s,a) + \mu_{\theta_f}(s,a) \mu_{\theta_f}(s,a)^\top \right) - \mu_{\theta_{-e}}(s,a) \mu_{\theta_{-e}}(s,a)^\top .
\end{equation}

We evaluate the connection between dynamics misalignment and the uncertainty estimate $u_{\mathrm{OvR}}$ on the pendulum toy example. Here we use the exact same setup as the one discussed in Section \ref{sec:toy_example_ugjs_test}, including the identical PE model. The only difference is that we compute the model uncertainty according to 
\begin{equation}
u_{\mathrm{OvR}, ij} = u_{\mathrm{OvR}}(s_{ij},a_{ij}).
\end{equation}
The evaluation over $\mathcal{S}$ is depicted in Figure \ref{fig:unc_m2ac_appendix}. We see a substantially more noisy uncertainty estimate of $u_{\mathrm{OvR}}$ as compared to the $u_{\mathrm{GJS}}$ uncertainty estimate proposed in MACURA and depicted in Figure \ref{fig:unc_ours_appendix}. Most importantly, we see $u_{\mathrm{OvR}}$ to be overconfident in areas around $\phi = 0$ and $\dot{\phi} = 0$ where the $u_{\mathrm{OvR}}$ model uncertainty is low, while dynamics misalignment is high as can be seen from Figure \ref{fig:unc_tv_appendix}.

Trying to construct a subset
\begin{equation}
\mathcal{E}_{\mathrm{OvR}} := \{ s \in \mathcal{S} \mid u_{\mathrm{OvR}}(s, a) < \kappa_{\mathrm{OvR}} , a \sim \pi(\cdot \mid s) \}.
\label{eq: subset_gjs}
\end{equation}
 choosing a suitable $\kappa_{\mathrm{OvR}}$ does not yield a reasonable result as shown in Figure \ref{fig:E_m2ac_appendix}. The set has no clear boundary, due to the noisiness of the $u_{\mathrm{OvR}}$ uncertainty estimate. More importantly, the set does not align with areas of low dynamics misalignment, as depicted in Figure \ref{fig:E_TV_m2ac_appendix}, due to the overconfidence of $u_{\mathrm{OvR}}$.

\clearpage

\section{Experiments}
\label{appendix:experiments}

In the following, we discuss the details of the experimental setup in Section \ref{sec:app_exp_experimental_setup} and different approaches to exploring the environment in Dyna-style MBRL in Section \ref{sec:app_exp_exploration}.

\subsection{Experimental Setup}
\label{sec:app_exp_experimental_setup}

Instead of the original implementation of MBPO\footnote{\url{https://github.com/jannerm/mbpo}}, all implementations of this work are based on the more recent mbrl-lib library \cite{Pineda2021Apr}. For MBPO and SAC, we use the implementations provided by the library, while M2AC is reimplemented as an extension to MBPO, as no open-source version of the M2AC code is available. We further implement the MACURA algorithm based on the mbrl-lib version of MBPO.   The code is available online\footnote{\url{https://github.com/Data-Science-in-Mechanical-Engineering/macura} }.

We run five random seeds for each experiment. Plots show the mean over the corresponding runs as a solid line and the 95\% confidence interval as a shaded region.

In all experiments, the training data for the SAC agent comprises 95\% $\mathcal{D}_{\mathrm{mod}}$ and 5\% $\mathcal{D}_{\mathrm{env}}$ for MBPO, M2AC, and MACURA.

To achieve results comparable to the ones published in \cite{Janner2019Dec}, we tune the MBPO hyperparameters according to Table \ref{tab:mbpo_params}.

\begin{table}[h!]
    \centering
    \caption{Hyperparameters MBPO}
    \begin{tabular}{ |c|c|c|c|c|c| } 
         \hline
         \textbf{Environment} & Humanoid & Ant & Halfcheetah  & Walker & Hopper\\ 
         \hline
         \textbf{Epochs} & 300 & 300 & 400 & 200 & 125\\
         \hline
         \textbf{Steps per Epoch} & \multicolumn{5}{c|}{1000}\\
         \hline
         \textbf{PNNs per PE} & \multicolumn{5}{c|}{7}\\
         \hline
         \textbf{PNN Layers} & \multicolumn{5}{c|}{4}\\
         \hline
         \textbf{PNN Nodes per Layer} & 400 & \multicolumn{4}{c|}{200}\\
         \hline
         \textbf{Critic Layers} & \multicolumn{5}{c|}{3}\\
         \hline
         \textbf{Critic Nodes per Layer} & 2048 & \multicolumn{4}{c|}{1024}\\
         \hline
         \textbf{Actor Layers} & \multicolumn{5}{c|}{3}\\
         \hline
         \textbf{Actor Nodes per Layer} & 2048 & \multicolumn{4}{c|}{1024}\\
         \hline
         \textbf{SAC Target Ent.} & -10 & -1 & -4 & -1 & 0\\
         \hline
         \textbf{SAC Updates $G$} & \multicolumn{2}{c|}{20} & 10 & \multicolumn{2}{c|}{30}  \\
         \hline
         \textbf{Rollouts per Round $M$} & \multicolumn{2}{c|}{406} & 203 & \multicolumn{2}{c|}{406}  \\
         \hline
         \textbf{Rollouts length $T_\mathrm{max}$} & 1 $\rightarrow$ 25 & 1 $\rightarrow$ 25 & \multicolumn{2}{c|}{1} & 1 $\rightarrow$ 15 \\
         \hline
         \textbf{Episodes Schedule $T_\mathrm{max}$} & 20 $\rightarrow$ 300 & 20 $\rightarrow$ 100 & \multicolumn{2}{c|}{ } & 20 $\rightarrow$ 100 \\
         \hline
    \end{tabular}
    \label{tab:mbpo_params}
\end{table}

For MACURA, we choose the hyperparameters very close to MBPO for a fair comparison. In the MBPO implementation, the number of model-based rollouts per iteration $M$ needs to be a multiple of the number of ensemble members. We remove this requirement for the MACURA rollout scheme, allowing us to choose $M$ to plain hundreds. Also, we observe that the actual number of update steps $G$ according to \eqref{eq:G_adaption} in MACURA is roughly one-half of $G_{\mathrm{max}}$. Thus we choose $G_{\mathrm{max}}$ for MACURA to be twice as much as $G$ in MBPO. We further introduce the hyperparameters $T_{\mathrm{max}}$, $\zeta$, and $\xi$ of the uncertainty-aware rollout adaption scheme of MACURA. Here, we keep $T_{\mathrm{max}}$ and $\zeta$ constant among all environments and solely tune $\xi$ for the specific task. A comprehensive overview of the MACURA hyperparameters is provided in Table \ref{tab:macura_params}.
\begin{table}[h!]
    \centering
    \caption{Hyperparameters MACURA}
        \begin{tabular}{ |c|c|c|c|c|c| } 
         \hline
         \textbf{Environment} & Humanoid & Ant & Halfcheetah  & Walker & Hopper\\ 
         \hline
         \textbf{Epochs} & 300 & 300 & 400 & 200 & 125\\
         \hline
         \textbf{Steps per Epoch} & \multicolumn{5}{c|}{1000}\\
         \hline
         \textbf{PNNs per PE} & \multicolumn{5}{c|}{7}\\
         \hline
         \textbf{PNN Layers} & \multicolumn{5}{c|}{4}\\
         \hline
         \textbf{PNN Nodes per Layer} & 400 & \multicolumn{4}{c|}{200}\\
         \hline
         \textbf{Critic Layers} & \multicolumn{5}{c|}{3}\\
         \hline
         \textbf{Critic Nodes per Layer} & 2048 & \multicolumn{4}{c|}{1024}\\
         \hline
         \textbf{Actor Layers} & \multicolumn{5}{c|}{3}\\
         \hline
         \textbf{Actor Nodes per Layer} & 2048 & \multicolumn{4}{c|}{1024}\\
         \hline
         \textbf{SAC Target Ent.} & -10 & -1 & -4 & -1 & 0\\
         \hline
         \textbf{SAC Updates $G_{\mathrm{max}}$} & \multicolumn{2}{c|}{40} & 20 & \multicolumn{2}{c|}{60}  \\
         \hline
         \textbf{Rollouts per Round $M$} & \multicolumn{2}{c|}{400} & 200 & \multicolumn{2}{c|}{400}  \\
         \hline
         \textbf{Rollouts length $T_\mathrm{max}$} & \multicolumn{5}{c|}{10}\\
         \hline
         \textbf{Quantile Factor $\zeta$}  & \multicolumn{5}{c|}{0.95} \\
         \hline
         \textbf{Scaling Factor $\xi$}  &  \multicolumn{2}{c|}{5} & 2 & 0.3 & 30 \\
         \hline
    \end{tabular}
    \label{tab:macura_params}
\end{table}

In the case of M2AC, we have issues producing stable results. This forces us to deviate from the hyperparameter setup of MBPO. Similar to MACURA, we remove the requirement for $M$ to be a multiple of the number of ensemble members, thus choosing $M$ to plain hundreds. In some environments, we observe M2AC yield better results with a fixed temperature parameter of the SAC agent, thus we disable automatic entropy tuning in these cases\footnote{If the SAC Target Ent. hyperparameter is shown as N/A in the hyperparameter tables, this means that automatic entropy tuning \cite{Haarnoja2018Dec} is not used for this case, instead a fixed temperature parameter \cite{Haarnoja2018Jul} of 0.2 is used.}. For the study of different exploration schemes in Section \ref{sec:app_exp_exploration}, we find that hyperparameter settings of M2AC that are stable for deterministic environment interaction, destabilize when exploring and vice versa. Therefore, we provide two sets of hyperparameters. Table \ref{tab:m2ac_det_params} represents the hyperparameter setting for deterministic interaction, and Table \ref{tab:m2ac_stoch_params} shows the hyperparameters used when exploring with white or pink noise.

\begin{table}[]
    \centering
    \caption{Hyperparameters M2AC (Deterministic Environment Policy)}
    \begin{tabular}{ |c|c|c|c|c|c| } 
         \hline
         \textbf{Environment} & Humanoid & Ant & Halfcheetah  & Walker & Hopper\\ 
         \hline
         \textbf{Epochs} & 300 & 300 & 400 & 200 & 125\\
         \hline
         \textbf{Steps per Epoch} & \multicolumn{5}{c|}{1000}\\
         \hline
         \textbf{PNNs per PE} & \multicolumn{5}{c|}{7}\\
         \hline
         \textbf{PNN Layers} & \multicolumn{5}{c|}{4}\\
         \hline
         \textbf{PNN Nodes per Layer} & \multicolumn{5}{c|}{200}\\
         \hline
         \textbf{Critic Layers} & \multicolumn{5}{c|}{3}\\
         \hline
         \textbf{Critic Nodes per Layer} & \multicolumn{2}{c|}{1024} & 512 & \multicolumn{2}{c|}{1024}\\
         \hline
         \textbf{Actor Layers} & \multicolumn{5}{c|}{3}\\
         \hline
         \textbf{Actor Nodes per Layer} & \multicolumn{2}{c|}{1024} & 512 & \multicolumn{2}{c|}{1024}\\
         \hline
         \textbf{SAC Target Ent.} & -17 & N/A & -6 & -1 & 0\\
         \hline
         \textbf{SAC Updates $G$}& \multicolumn{2}{c|}{20} & 10 & \multicolumn{2}{c|}{30}  \\
         \hline
         \textbf{Rollouts per Round $M$} & \multicolumn{2}{c|}{400} & 200 & \multicolumn{2}{c|}{400}  \\
         \hline
         \textbf{Rollouts length $T_\mathrm{max}$} & \multicolumn{5}{c|}{10}\\
         \hline
    \end{tabular}
    \label{tab:m2ac_det_params}
\end{table}

\begin{table}[]
    \centering
    \caption{Hyperparameters M2AC (Stochastic Environment Policy)}
    \begin{tabular}{ |c|c|c|c|c|c| } 
         \hline
         \textbf{Environment} & Humanoid & Ant & Halfcheetah  & Walker & Hopper\\ 
         \hline
         \textbf{Epochs} & 300 & 300 & 400 & 200 & 125\\
         \hline
         \textbf{Steps per Epoch} & \multicolumn{5}{c|}{1000}\\
         \hline
         \textbf{PNNs per PE} & \multicolumn{5}{c|}{7}\\
         \hline
         \textbf{PNN Layers} & \multicolumn{5}{c|}{4}\\
         \hline
         \textbf{PNN Nodes per Layer} & \multicolumn{5}{c|}{200}\\
         \hline
         \textbf{Critic Layers} & \multicolumn{5}{c|}{3}\\
         \hline
         \textbf{Critic Nodes per Layer} & \multicolumn{5}{c|}{1024}\\
         \hline
         \textbf{Actor Layers} & \multicolumn{5}{c|}{3}\\
         \hline
         \textbf{Actor Nodes per Layer} & \multicolumn{5}{c|}{1024}\\
         \hline
         \textbf{SAC Target Ent.} & -17 & -1 & -4 & \multicolumn{2}{c|}{N/A}\\
         \hline
         \textbf{SAC Updates $G$}& \multicolumn{2}{c|}{20} & \multicolumn{2}{c|}{10} &  30 \\
         \hline
         \textbf{Rollouts per Round $M$} & \multicolumn{2}{c|}{400} & 200 & \multicolumn{2}{c|}{400}  \\
         \hline
         \textbf{Rollouts length $T_\mathrm{max}$} & \multicolumn{5}{c|}{10}\\
         \hline
    \end{tabular}
    \label{tab:m2ac_stoch_params}
\end{table}

Finally, we choose the hyperparameters of the SAC baseline, according to Table \ref{tab:sac_params}. Similar to M2AC, SAC yields better results with a fixed temperature hyperparameter\footnote{If the SAC Target Ent. hyperparameter is shown as N/A in the hyperparameter table, this means that automatic entropy tuning \cite{Haarnoja2018Dec} is not used for this case, instead a fixed temperature parameter \cite{Haarnoja2018Jul} of 0.2 is used.}.
\begin{table}[]
    \centering
    \caption{Hyperparameters SAC}
    \begin{tabular}{ |c|c|c|c|c|c| } 
    \hline
         \textbf{Environment} & Humanoid & Ant & Halfcheetah  & Walker & Hopper\\ 
         \hline
         \textbf{Critic Layers} & \multicolumn{5}{c|}{3}\\
         \hline
         \textbf{Critic Nodes per Layer} & \multicolumn{5}{c|}{256}\\
         \hline
         \textbf{Actor Layers} & \multicolumn{5}{c|}{3}\\
         \hline
         \textbf{Actor Nodes per Layer} & \multicolumn{5}{c|}{256}\\
         \hline
         \textbf{SAC Target Ent.} & -17 & N/A & -6 & -6 & -3\\
         \hline
         \textbf{SAC Updates $G$}& \multicolumn{5}{c|}{1} \\
         \hline
    \end{tabular}
    \label{tab:sac_params}
\end{table}

\subsection{Discussion $\xi$}
\label{sec:app_exp_xi}

The parameter $\xi$ allows us to tune what \emph{sufficiently} certain is. The general idea is that in the first rollout step, the model is evaluated in states that are in $\mathcal{D}_{\mathrm{env}}$ and have been seen before. Thus, the uncertainty values we get for this first step are values of the GJS that correspond to being \emph{certain}. So we choose the base uncertainty $\hat{u}_{\mathrm{GJS}}$ to be the $\zeta = 0.95$ quantile of uncertainties after this first rollout step. Taking this base uncertainty as the uncertainty threshold, which corresponds to $\xi = 1$, yields reasonable results in all environments. In this case, whenever a prediction step overshoots the $0.95$ quantile of uncertainty within the data support, we consider it to have left $\mathcal{E}$ and terminate the model-based rollout.
    
    However, the model is often very certain within the data support towards the end of the training, leading to a low uncertainty threshold $\kappa$ resulting from $\xi=1$. This leads to a narrow data distribution in $\mathcal{D}_{\mathrm{mod}}$. Repeatedly training the critics on this narrow distribution causes overfitting. This introduces instability towards the end of training as discussed in Section \ref{sec:exp_xi} and shown in Figure \ref{fig:humanoid_kappa}. Thus, we typically choose $\xi > 1$. This is true for all MuJoCo environments but Walker. The Walker dynamics seem hard to learn for the PE model. This can e.g. be seen from the learning behavior of MBPO in Figure \ref{fig:mujoco}. Even though MBPO solely performs model-based rollouts of length $1$ throughout training, learning appears rather brittle. Thus, in the Walker task, the $0.95$ quantile of uncertainties after the first rollout step is too uncertain for stable learning. Therefore, we chose $\xi < 1$ to stabilize learning in this case.
    
    From our experience, $\xi$ can be tuned in the following way: 
    \begin{compactitem}
        \item Perform a run with $\xi=1$. This should result in stable learning as long as the agent is in the early stages of training and improves sufficiently fast. As soon as the algorithm is close to its asymptotic performance, the uncertainty threshold might get too low such that instabilities in learning occur.
        \item  If this is the case, increase $\xi$. As indicated in Figure \ref{fig:humanoid_kappa}, there is a relatively broad range of $\xi$ values that yields stable performance. If $\xi$ is chosen too large, however, instabilities due to model exploitation occur.
        \item  If instabilities occur early in training, when running MACURA with $\xi = 1$ the model probably approximates the true dynamics poorly. This can be checked e.g. by reproducing trajectories in $\mathcal{D}_{\mathrm{env}}$ through model-based rollouts and considering the deviation. In this case, $\xi$ can be reduced until the learning behavior is stable.
    \end{compactitem} 
    
    The fact that the entire rollout mechanism can be tuned by choosing $\xi$, from our experience, makes tuning considerably easier than determining all the required hyperparameters of the M2AC mechanism or designing a suitable rollout schedule for MBPO.

\subsection{Exploration in the Environment}
\label{sec:app_exp_exploration}

The implementation used in this work, based on the mbrl-lib library \cite{Pineda2021Apr}, uses deterministic environment exploration for MBPO yielding similar results to the original implementation \cite{Janner2019Dec}. As far as we can reconstruct, the original implementation \cite{Janner2019Dec} uses white noise exploration in the agent environment interaction, however, when implementing white noise exploration in the mbrl-lib code, we observe substantially different results from those reported in \cite{Janner2019Dec}. These range from better performance to divergence. For our reimplementation of M2AC based on the mbrl-lib code, deterministic environment interaction also yields the most stable results. Thus, we present the performance of deterministic environment interaction for MBPO and M2AC in Figure \ref{fig:mujoco}.

\begin{figure}
     \centering
         \includegraphics[width=\textwidth]{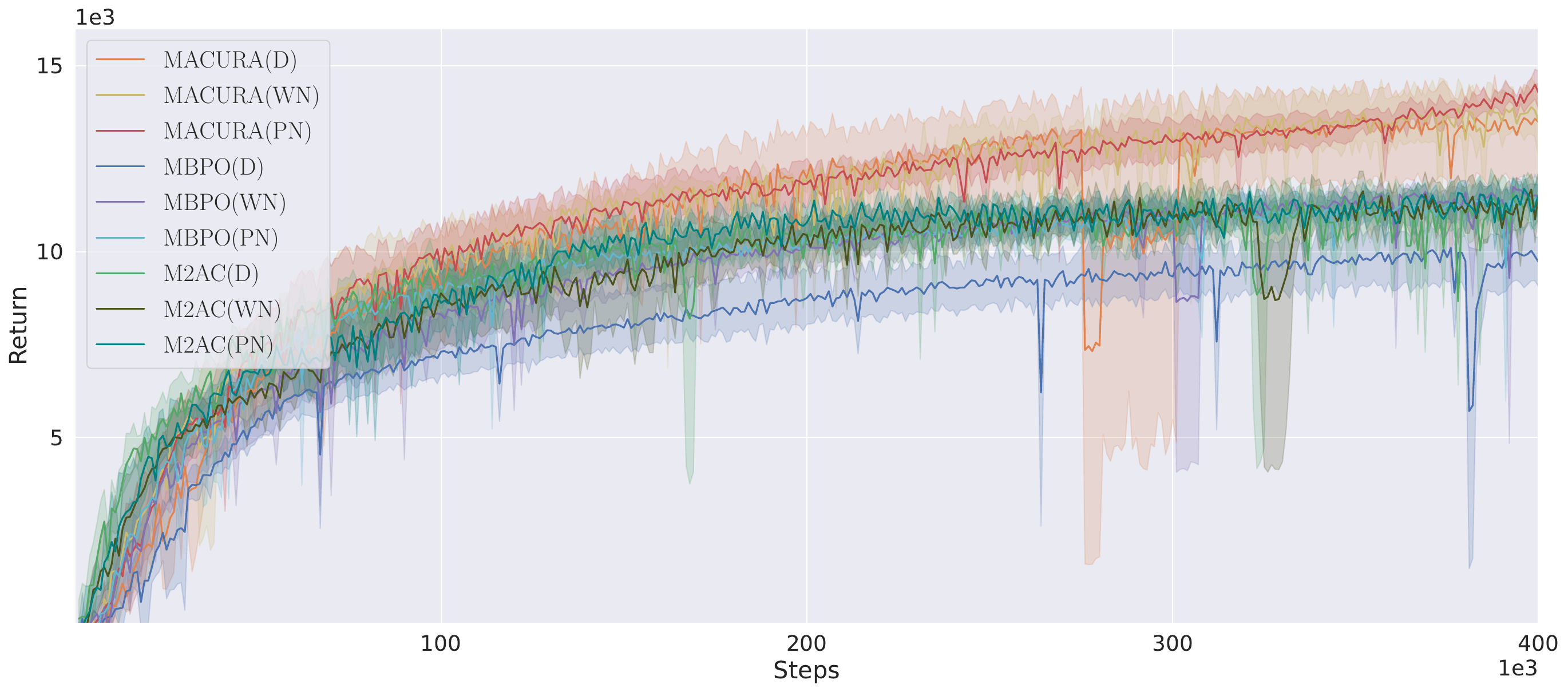}
         \caption{Exploration Schemes on MACURA, MBPO, and M2AC on Halfcheetah. \textit{Impact of deterministic (D), white noise (WN), and pink noise (PN) exploration on algorithmic performance.}}
         \label{fig:halfcheetah_exp} 
\end{figure}

\begin{figure}
     \centering
         \includegraphics[width=\textwidth]{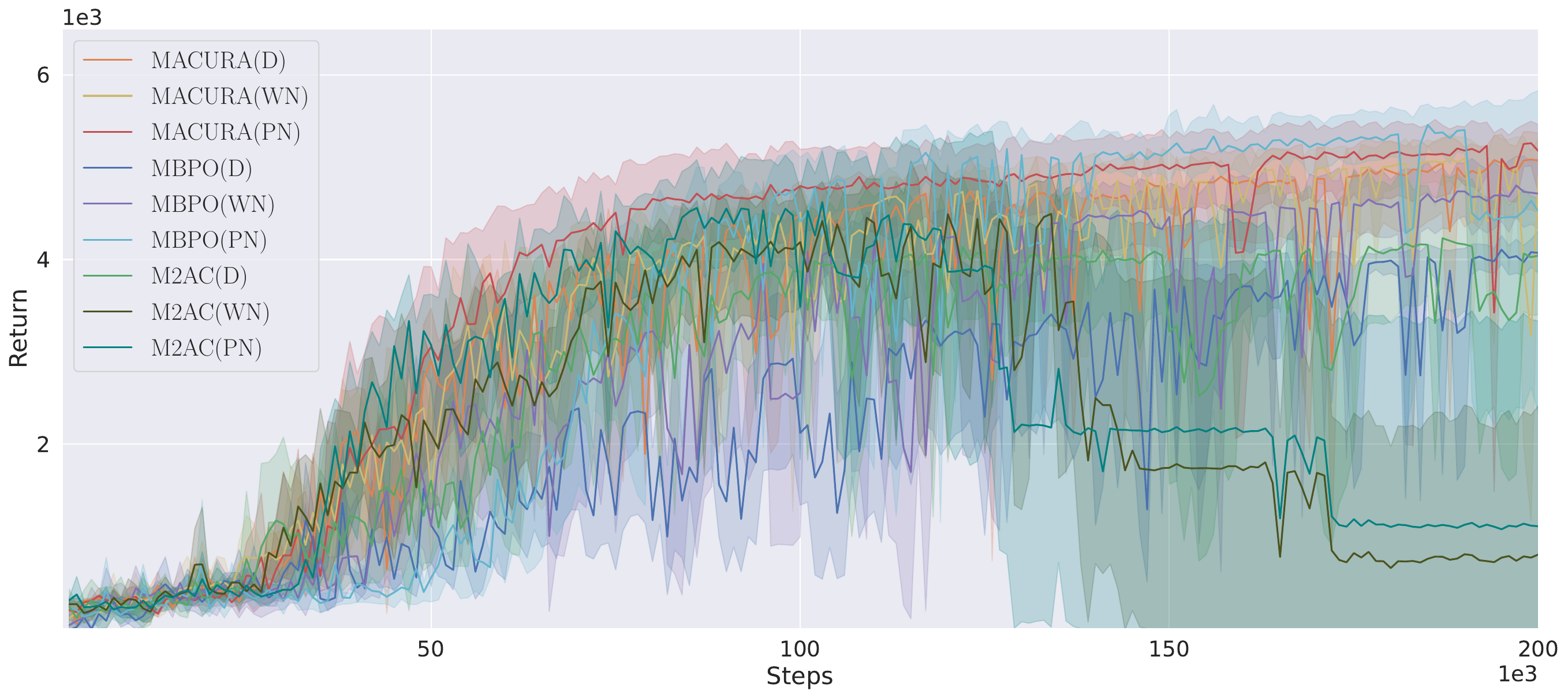}
         \caption{Exploration Schemes on MACURA, MBPO, and M2AC on Walker. \textit{Impact of deterministic (D), white noise (WN), and pink noise (PN) exploration on algorithmic performance.}}
         \label{fig:walker_exp} 
\end{figure}

\begin{figure}
     \centering
         \includegraphics[width=\textwidth]{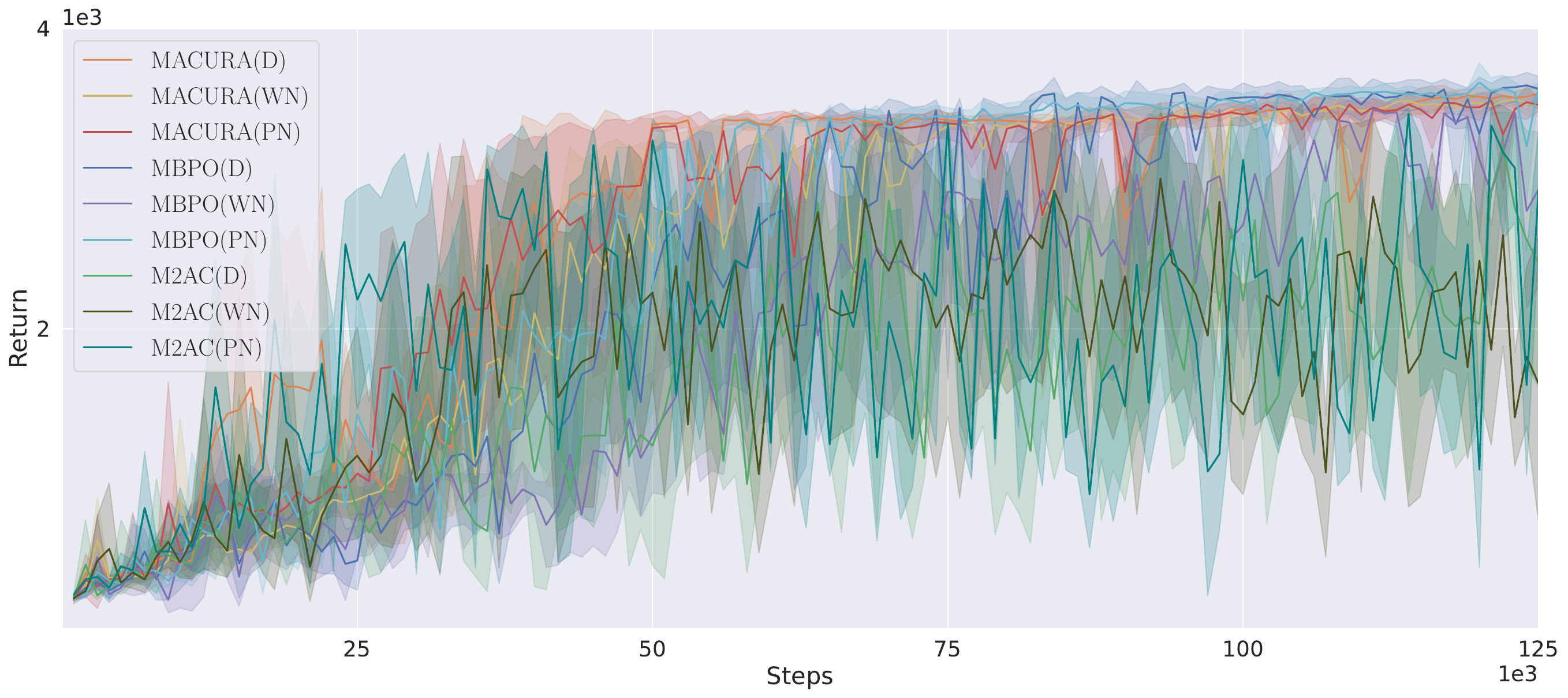}
         \caption{Exploration Schemes on MACURA, MBPO, and M2AC on Hopper. \textit{Impact of deterministic (D), white noise (WN), and pink noise (PN) exploration on algorithmic performance.}}
         \label{fig:hopper_exp} 
\end{figure}

\begin{figure}
     \centering
         \includegraphics[width=\textwidth]{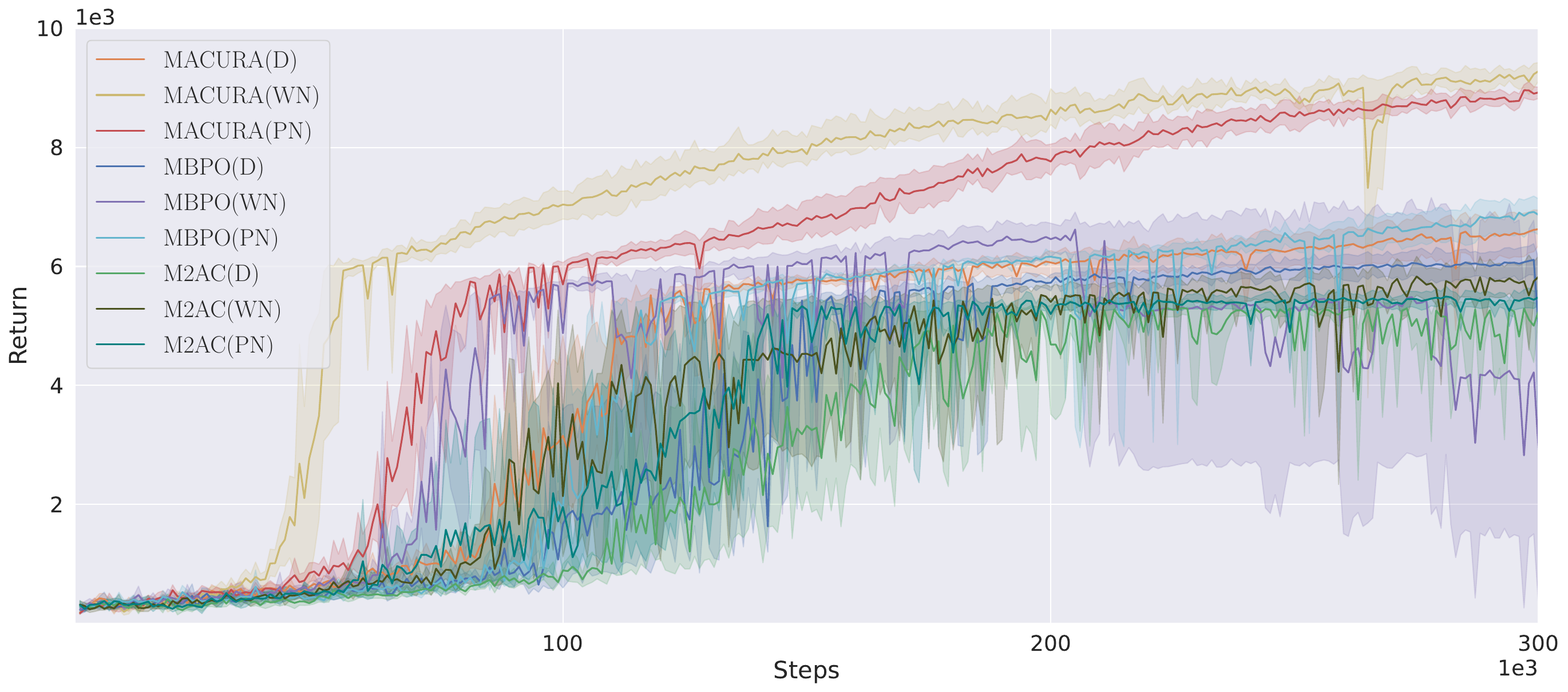}
         \caption{Exploration Schemes on MACURA, MBPO, and M2AC on Humanoid. \textit{Impact of deterministic (D), white noise (WN), and pink noise (PN) exploration on algorithmic performance.}}
         \label{fig:humanoid_exp} 
\end{figure}

\begin{figure}
     \centering
         \includegraphics[width=\textwidth]{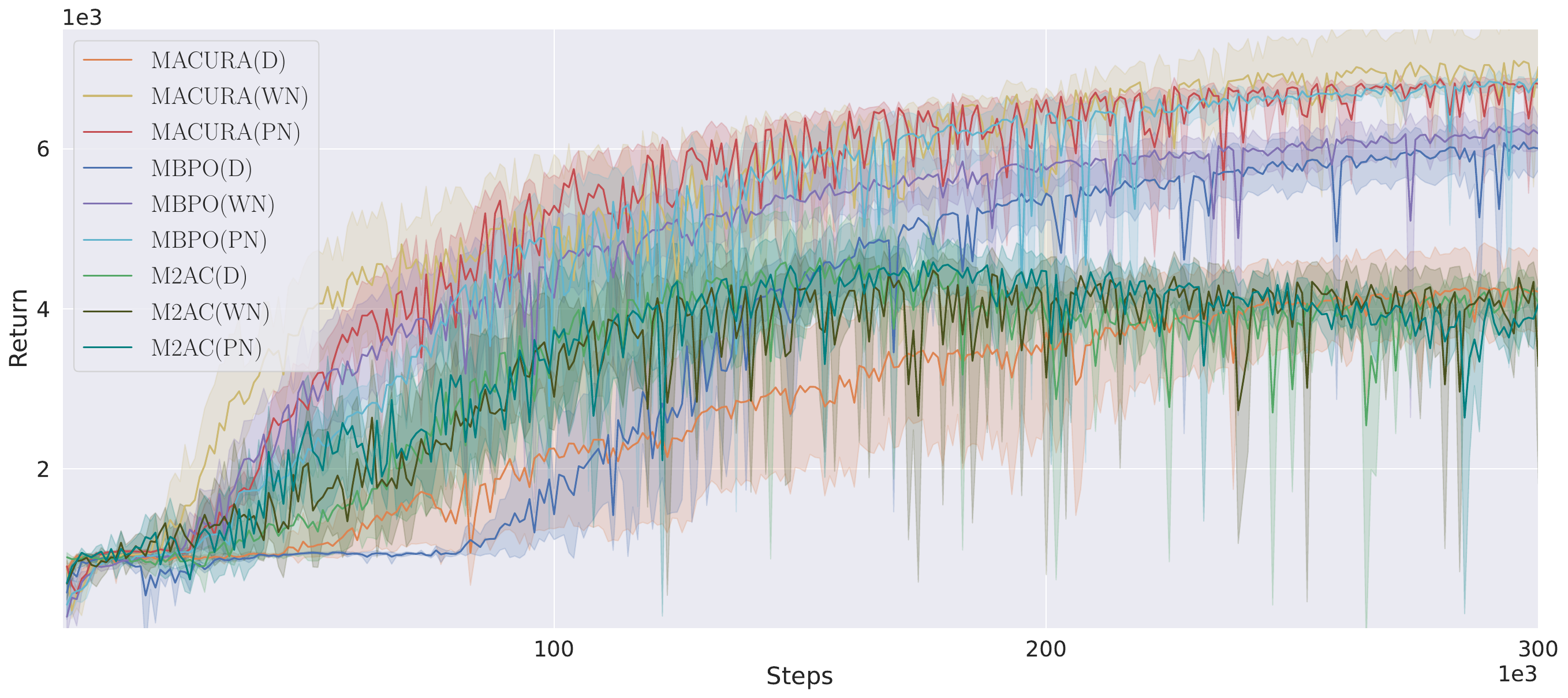}
         \caption{Exploration Schemes on MACURA, MBPO, and M2AC on Ant. \textit{Impact of deterministic (D), white noise (WN), and pink noise (PN) exploration on algorithmic performance.}}
         \label{fig:ant_exp} 
\end{figure}

In Figures \ref{fig:halfcheetah_exp} - \ref{fig:ant_exp} we present the performance of MACURA, MBPO, and M2AC with the considered exploration schemes: deterministic interaction, white exploration noise and pink noise exploration \cite{eberhard2023pink}, on the MuJoCo benchmark.

We have issues producing stable results for M2AC, despite putting the most tuning effort among the compared approaches. We find that hyperparameter settings that are stable for deterministic interaction destabilize in the case of exploration noise and vice versa. Thus we provide a set of hyperparameters for deterministic environment interaction and one that we use for white and pink noise exploration. For MACURA and MBPO we run the same sets of hyperparameters for all exploration schemes. Generally, we see the performance of M2AC fall short as compared to MACURA and MBPO.

For MBPO, we use the fine-tuned rollout schemes reported in \cite{Janner2019Dec}. White noise exploration in most cases positively impacts asymptotic performance but increases the variance among runs. In some instances, we observe that white noise exploration destabilizes learning in MBPO, leading to divergence. For pink noise exploration, we can report a generally positive impact on the performance of MBPO, increasing data efficiency and asymptotic performance without destabilizing learning. In some environments, MBPO with a fine-tuned rollout schedule and pink noise exploration can even compete with MACURA.

MACURA with deterministic environment interaction shows the strongest performance among deterministic interaction methods in most environments. White noise exploration in some cases yields better results than pink noise exploration, while considerably increasing variance among trained agents. Different from MBPO, however, white noise exploration does not destabilize MACURA up to the point of divergence.
Overall, we observe that MACURA with pink noise exploration yields the best data efficiency, asymptotic performance, and stability among all approaches on the benchmark. Combined with a considerably easier-to-tune rollout length scheduling mechanism than MBPO and M2AC we consider this the most promising approach for Dyna-style MBRL.

\subsection{Noisy Environment}
\label{sec:app_exp_noisy}
\begin{figure}[h]
     \centering
    \includegraphics[width=\textwidth]{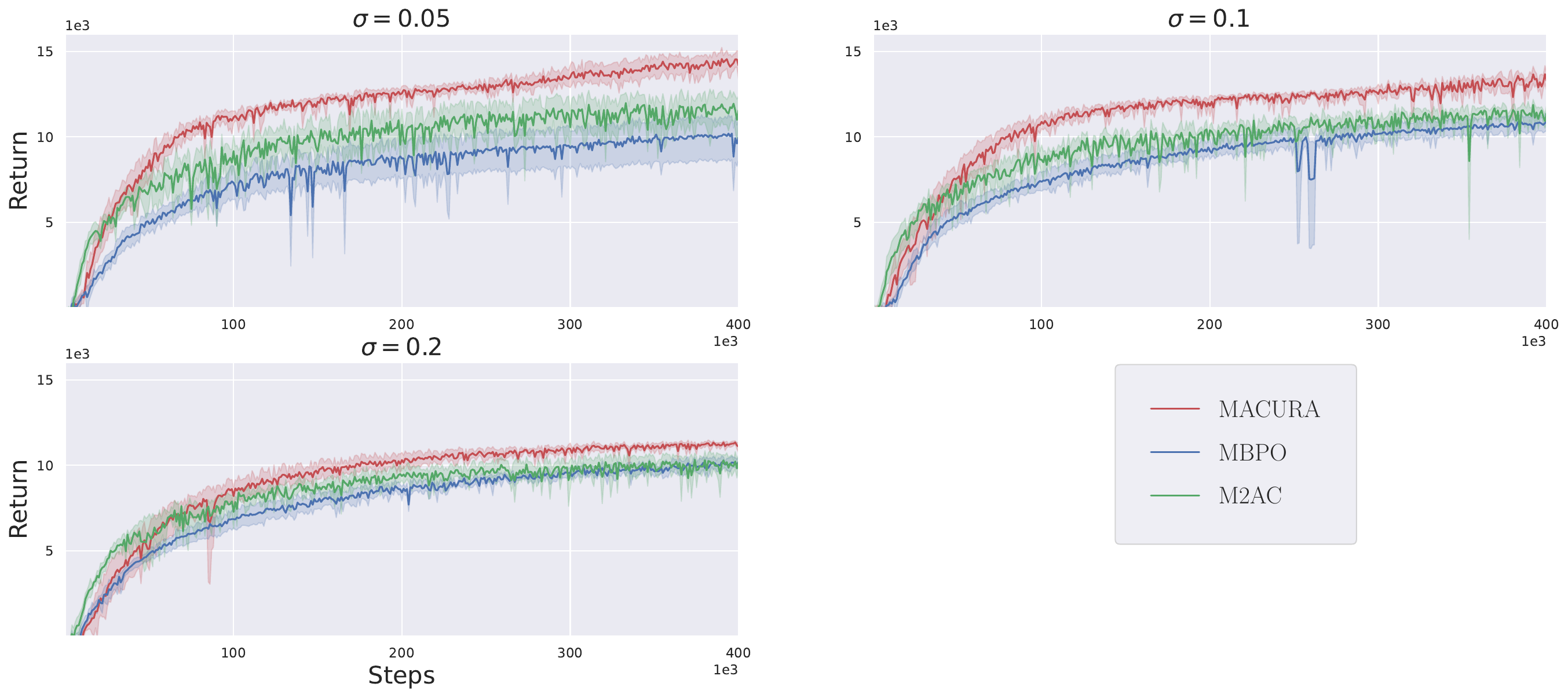}
\caption{Returns obtained by MACURA, MBPO, and M2AC on noisy Halfcheetah.}
 \label{fig:noise}
\end{figure}

We evaluate MBPO, M2AC, and MACURA on a noisy version of the Halfcheetah environment. To introduce noise, we take the approach proposed in \cite{Pan2020Dec} and add noise to the actions applied to the environment. Therefore, the action applied to the environment is $\tilde{a}_t = a_t + \epsilon$ with $\epsilon \sim \mathcal{N}(0, \Sigma)$. The covariance matrix $\Sigma$ is a diagonal matrix with diagonal elements $\sigma^2$. The additive noise is not observable to the agent and therefore introduces process noise. Following the experimental setup in \cite{Pan2020Dec}, we conduct three experiments with $\sigma = 0.05$, $\sigma = 0.1$ , and $\sigma = 0.2$ respectively. The corresponding results are depicted in Figure \ref{fig:noise}. MACURA consistently outperforms MBPO and M2AC.

\subsection{Rollout Horizon, Data Uncertainty, Gradient Steps}
\label{sec:app_exp_rollout}
We evaluate the rollout - and gradient step adaption mechanism of MACURA on the Walker environment. We compare MACURA to M2AC and MBPO.

\begin{figure}[h]
     \centering
    \includegraphics[width=0.95\textwidth]{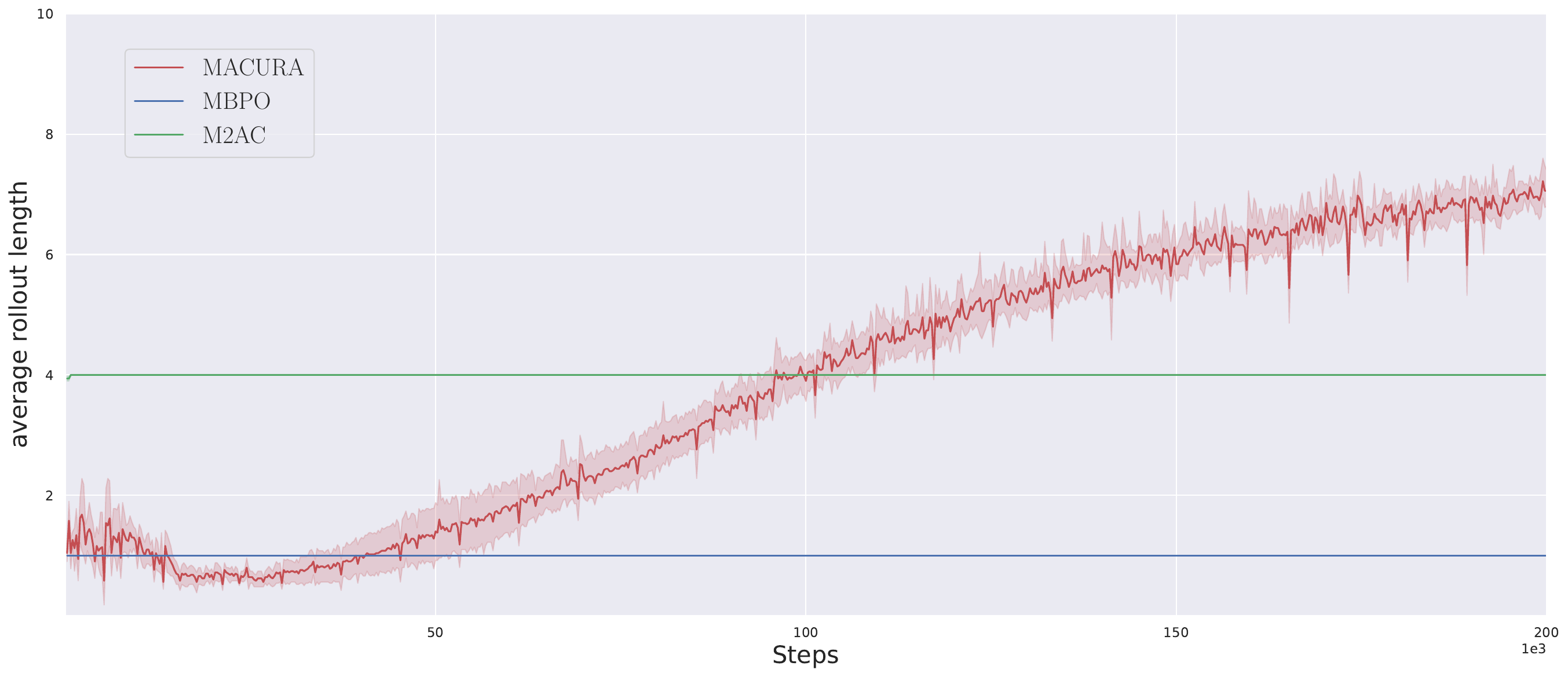}
\caption{Average rollout lengths for MACURA, MBPO, and M2AC on Walker.}
 \label{fig:rollout}
\end{figure}

First, we consider the average rollout length of the respective approaches throughout training as depicted in Figure \ref{fig:rollout}. Therefore, we record the rollout lengths of all $M$ model-based rollouts performed in parallel during one round of model-based rollouts and compute their average.
As proposed in \cite{Janner2019Dec}, MBPO performs one-step rollouts throughout training leading to an average rollout length of one. The rank-based filtering heuristic of M2AC terminates a fixed quantile of model-based rollouts after each rollout step.
Effectively, this yields a predefined distribution over rollout lengths, which does not change throughout training.
As a side effect, this leads to a constant average rollout length as depicted in Figure \ref{fig:rollout}.
MACURA, instead, has a threshold-based rollout length adaption mechanism, thus the average rollout length varies throughout training. MACURA performs short rollouts in the early stages of training, where the model has limited capabilities and increases the rollout length as the model improves. Different from MBPO and M2AC, MACURA can even discard the first rollout step enabling an average rollout length lower than one.

Next, we investigate the uncertainty of the corresponding data created in model-based rollouts. We measure uncertainty with the variance of the mean predictions of the individual PNNs within the PE. To recover a scalar uncertainty, we take the Frobenius norm of the resulting matrix: 
\begin{equation}
    u_{\mathrm{MV}}(s,a) = \norm{ \frac{1}{E} \sum_{e=1}^{E} (\mu_{\theta_e}(s,a) - \mu_{\mathrm{PE}}(s,a))(\mu_{\theta_e}(s,a) - \mu_{\mathrm{PE}}(s,a))^{\top} }_{\mathrm{F}} 
    \label{eq:mean_variance}
\end{equation}
with $\mu_{\mathrm{PE}}(s,a) = \frac{1}{E} \sum_{e=1}^{E} \mu_{\theta_e}(s,a)$. By doing this, we aim to avoid artifacts introduced by comparing uncertainty in the metrics used in M2AC or MACURA and instead provide an impartial analysis of how uncertainty evolves in model-based rollouts.

\begin{figure}[h]
     \centering
    \includegraphics[width=0.95\textwidth]{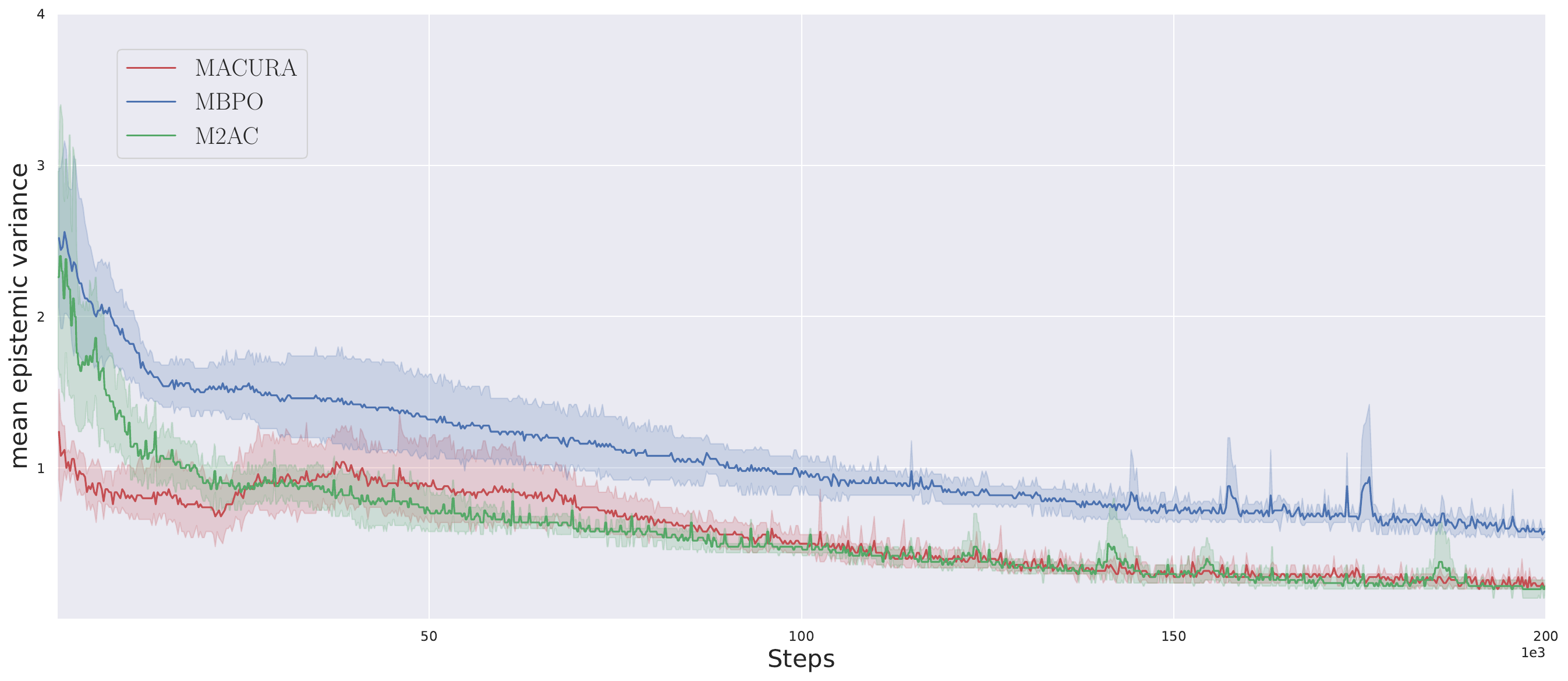}
\caption{Mean epistemic uncertainty of rollouts for MACURA, MBPO, and M2AC on Walker.}
 \label{fig:uncertainty}
\end{figure}

The average uncertainty corresponding to the respective Dyna-style approaches is depicted in Figure \ref{fig:uncertainty}. MBPO has the highest average uncertainty that reduces throughout training. M2AC shows lower average uncertainty than MBPO. We assume this is the case, as low-uncertainty rollouts are propagated for several rollout steps reducing the mean uncertainty. MACURA shows an average uncertainty comparable to M2AC. However, the initial average uncertainty in MACURA is substantially lower than in M2AC and MBPO due to the threshold-based rollout adaption mechanism. Different from the rank-based heuristic of M2AC that always takes a particular most certain percentage into account (which can include very uncertain data when initially most data is bad), MACURA can discard all the data that is above the uncertainty threshold. 

It should be noted that the average uncertainty gives a rough impression of data quality but is not a suitable metric for detecting low-quality outliers that occur with low probability. These outliers, however, have a strong influence on the learning process from our experience.

\begin{figure}[h]
     \centering
    \includegraphics[width=0.95\textwidth]{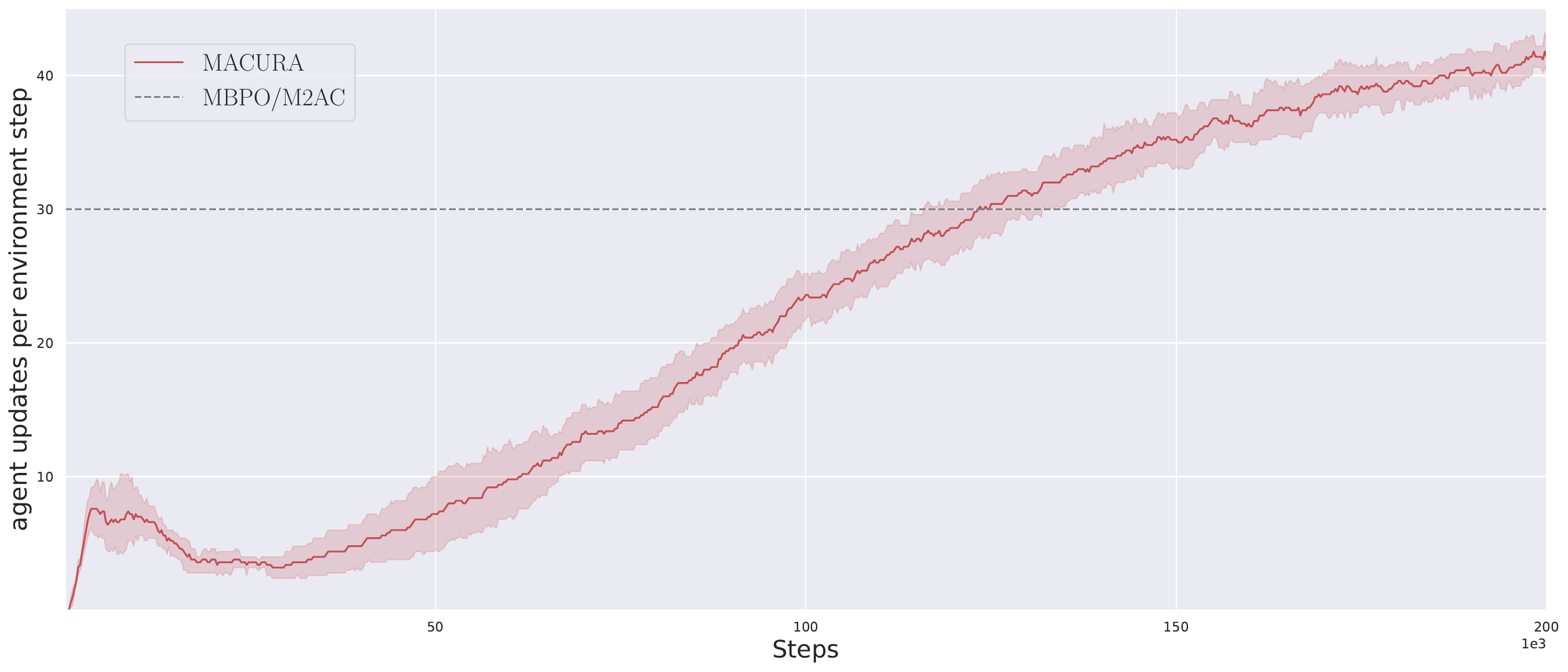}
\caption{SAC Update Steps}
 \label{fig:sac_updates}
\end{figure}

\begin{figure}[h]
     \centering
    \includegraphics[width=0.95\textwidth]{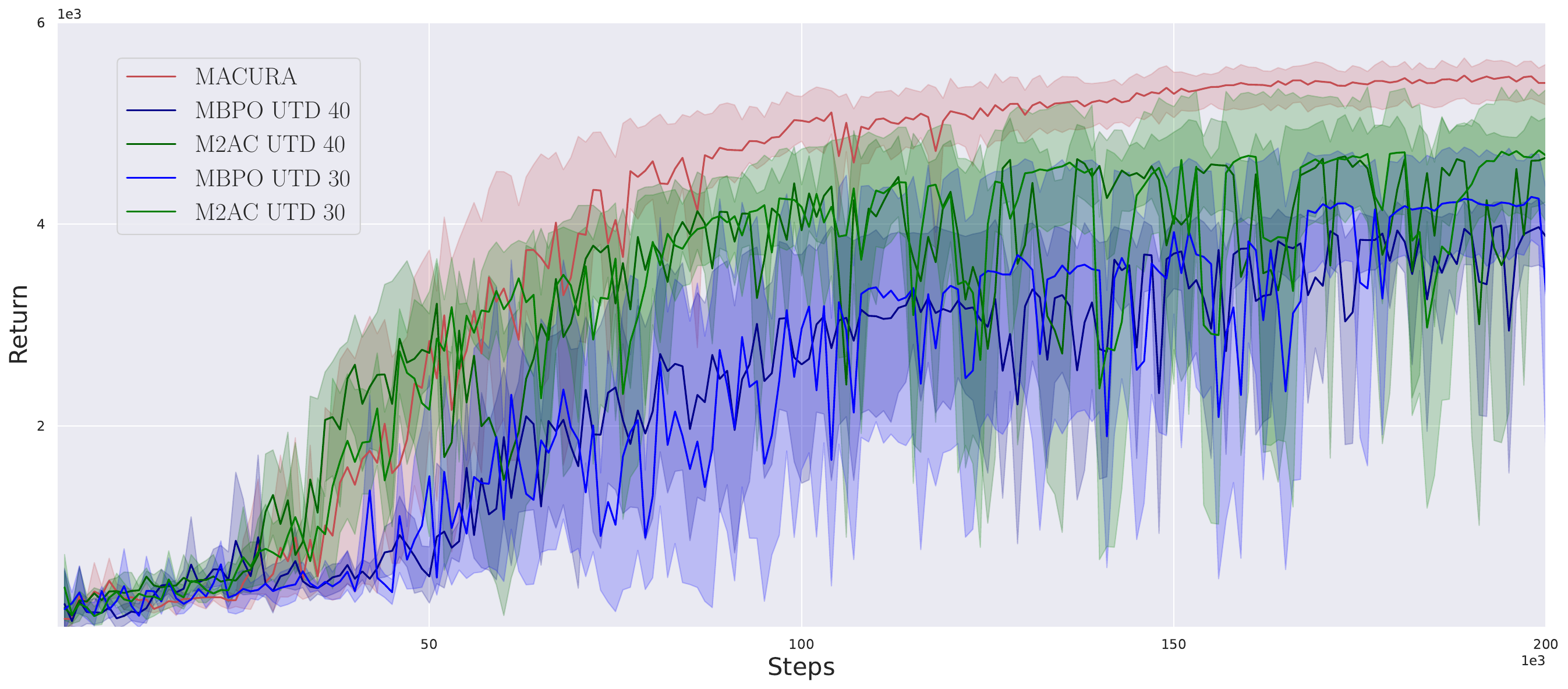}
\caption{Comparison of returns obtained on Walker.}
 \label{fig:return_utd40}
\end{figure}

The varying amounts of data created in model-based rollouts require adapting the amount of SAC updates in MACURA. Figure \ref{fig:sac_updates} shows the average gradient steps $G$ of the respective approaches.
Both MBPO and M2AC have a $G=30$ fixed throughout training, while MACURA adapts $G$ depending on the occupancy of $\mathcal{D}_{\mathrm{mod}}$ \eqref{eq:G_adaption} that depends on the average rollout length depicted in Figure \ref{fig:rollout}. Therefore, MACURA performs fewer SAC updates in the early stages of training but catches up as the model improves. Towards the end of training, MACURA performs a bit over 40 SAC updates per timestep. Simply increasing $G$ to 40 in MBPO and M2AC, however, does not yield stable results, as depicted in Figure \ref{fig:return_utd40} (in the figure $G$ is referred to as update-to-data (UTD) ratio).

\subsection{Uncertainty Estimates}
\label{sec:app_exp_uncertainty_est}
\begin{figure}[h]
     \centering
    \includegraphics[width=0.95\textwidth]{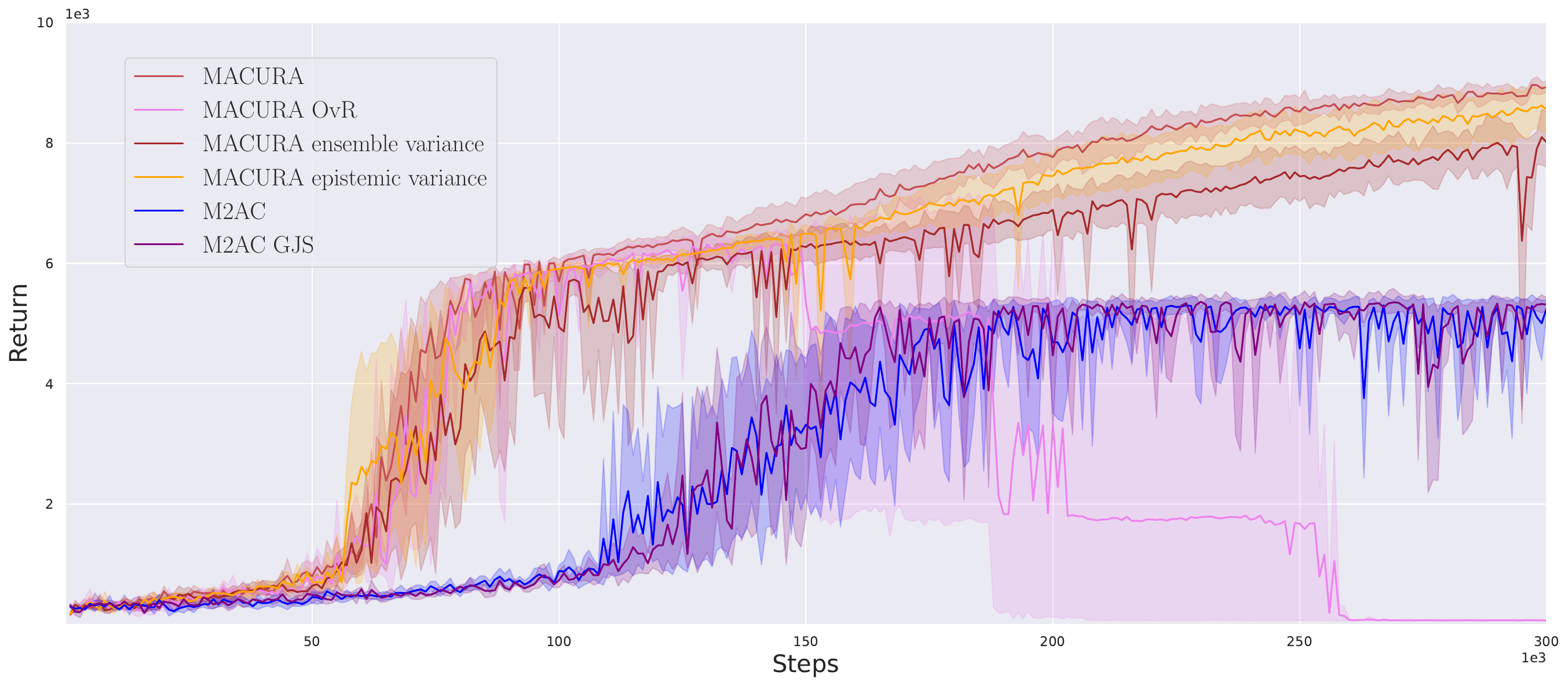}
\caption{MACURA and M2AC with different uncertainty metrics on Humanoid.}
 \label{fig:return_unc}
\end{figure}

In the following, we investigate the performance of MACURA and M2AC with different uncertainty estimates. Results on the Humanoid environment are depicted in Figure \ref{fig:return_unc}.

M2AC with the GJS uncertainty estimate learns slightly faster and shows slightly stronger performance with less variance. We assume this to be the case, as the GJS estimate is more reliable in detecting low-quality data, thus stabilizing learning.

MACURA with OvR uncertainty estimate \cite{Pan2020Dec} performs well up to a return of about 7000 and subsequently diverges presumably due to the brittleness of the OvR uncertainty estimate.

Additionally, we conduct experiments with MACURA using the ensemble variance estimate proposed in \cite{Lu2021Oct} and mean variance (\ref{eq:mean_variance}).

While MACURA works with both these uncertainty metrics, they yield weaker results than the GJS estimate. This aligns with our intuition, as ensemble variance is a combined estimate for epistemic and aleatoric uncertainty, while we are solely interested in epistemic uncertainty. The variance of ensemble means solely addresses epistemic variance but has no information about the aleatoric uncertainty at all, which can be problematic. Epistemic uncertainty, however, is defined by the disagreement of individual PNN predictions, which rather corresponds to the overlap in the probability mass of the individual predictions. The GJS uncertainty estimate measures this overlap in probability mass, thus providing a better metric for epistemic uncertainty and works better in practice.

\subsection{Prolonged Experiments}
\label{sec:app_exp_long_exp}

\begin{figure}[t]
     \centering
    \includegraphics[width=0.95\textwidth]{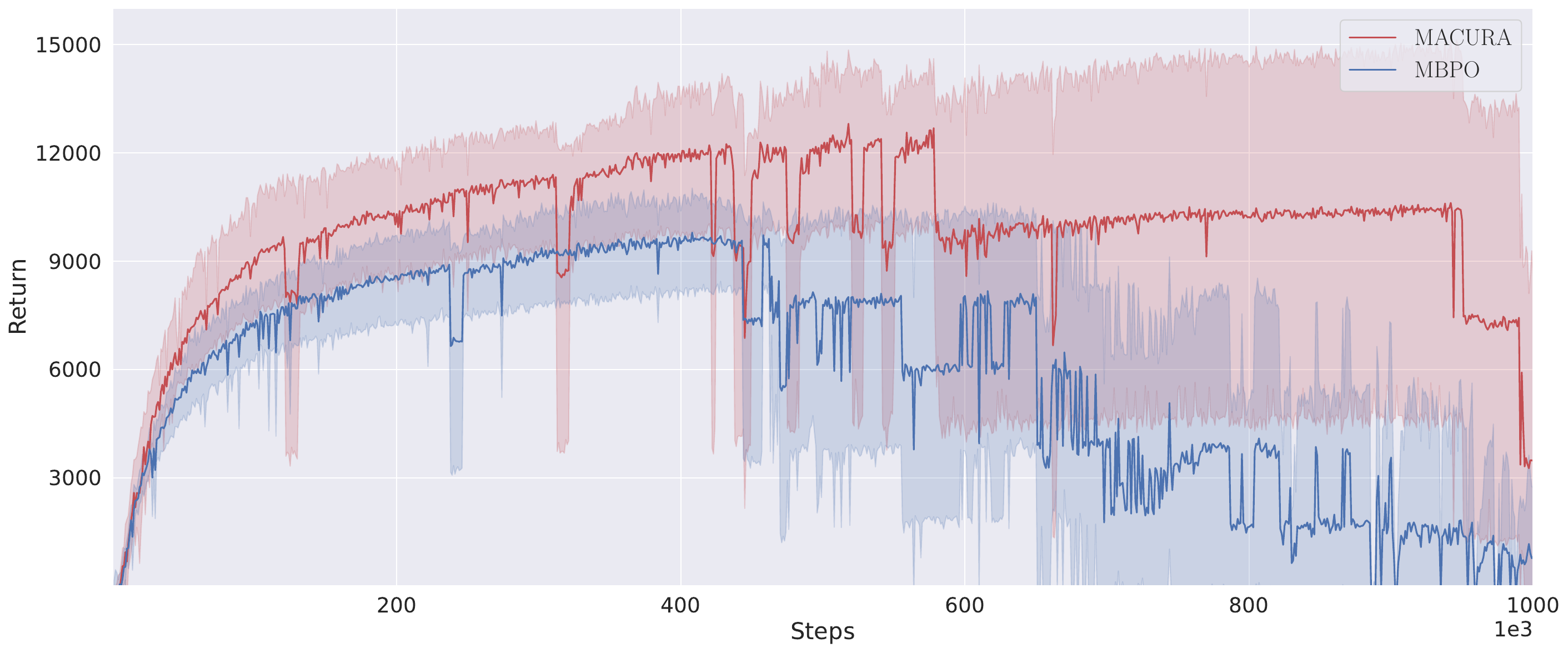}
\caption{MACURA and MBPO with prolonged experiment length on Halfcheetah.}
 \label{fig:longruns}
\end{figure}

In the following, we discuss the behavior of MBPO and MACURA, when trained beyond the typically reported length of experiments \cite{Janner2019Dec, Pan2020Dec}. Figure \ref{fig:longruns} depicts results on Halfcheetah for 1,000,000 environment interactions as opposed to the 400,000 environment interactions presented in Figure \ref{fig:mujoco}.  We do not consider M2AC in these experiments as stabilizing M2AC even for the first 400,000 environment interactions is a substantial challenge.
Both MBPO and MACURA destabilize after an extended amount of training. From our experience, this is due to overfitting of either the model or the critic after training on similar data distributions over and over again. The training heuristics for model and critic of MBPO that are largely inherited by MACURA are not well suited for continuing training. Extending these algorithms to a continuing training setting requires further research addressing \emph{when to train your model and agent} that is beyond the scope of this work.

\end{document}